\documentclass[10pt,letterpaper]{paper}

\usepackage[all]{hypcap}

\usepackage[authoryear, round]{natbib}

\usepackage[utf8]{inputenc} 
\usepackage[T1]{fontenc}    
\usepackage{url}            
\usepackage{booktabs}       
\usepackage{amsfonts}       
\usepackage{nicefrac}       
\usepackage{microtype}      
\usepackage{multirow} 
\usepackage[textsize=tiny]{todonotes}
\usepackage{algorithm}
\usepackage{amssymb}
\usepackage{cleveref}
\usepackage{dsfont}
\usepackage{nicefrac}
\usepackage{inconsolata}
\usepackage{xcolor}
\usepackage{graphicx}
\usepackage{amsmath}
\usepackage{amssymb}
\usepackage{booktabs}
\usepackage{multirow}
\usepackage{makecell}
\usepackage{caption}
\usepackage{subcaption}
\usepackage{bbm}
\usepackage{wrapfig}
\usepackage{mathtools,xparse}
\usepackage{pifont}
\usepackage{enumitem}
\usepackage{array}
\usepackage{scalerel,xparse}
\usepackage{colortbl}
\usepackage{tablefootnote}
\usepackage{tocloft}  
\usepackage{hyperref}  
\usepackage{dblfloatfix} 
\usepackage{adjustbox}
\usepackage{algpseudocode}
\usepackage{setspace}
\usepackage[most,skins,theorems]{tcolorbox}

\newboolean{showsection}
\setboolean{showsection}{true}
\makeatletter
\@namedef{ver@everyshi.sty}{}
\makeatother

\definecolor{custom_green}{rgb}{0.0, 0.5, 0.0}
\definecolor{custom_red}{rgb}{1.0, 0.01, 0.24}
\definecolor{custom_blue}{HTML}{C9DAF7}
\definecolor{custom_purple}{HTML}{D9D1E9}
\definecolor{title_blue}{HTML}{204899} 
\definecolor{cite_blue}{HTML}{044dc1}  
\definecolor{cite_purple}{HTML}{7406a7}  
\newcommand{\cmark}{\color{custom_green}\ding{51}}%
\newcommand{\xmark}{\color{custom_red}\ding{55}}%

\hypersetup{
    colorlinks = true,
    citecolor = {cite_blue},
    linkcolor = {cite_purple},
    urlcolor = {cite_purple},
}

\definecolor{blanchedalmond}{rgb}{1.0, 0.92, 0.8}
\definecolor{carmine}{rgb}{0.59, 0.0, 0.09}
\definecolor{lightblue}{rgb}{0.22,0.45,0.70}%

\renewcommand{\mathbf}{\boldsymbol}

\makeatletter
\def\Ddots{\mathinner{\mkern1mu\raise\p@
\vbox{\kern7\p@\hbox{.}}\mkern2mu
\raise4\p@\hbox{.}\mkern2mu\raise7\p@\hbox{.}\mkern1mu}}
\makeatother

\numberwithin{equation}{section}

\definecolor{amaranth}{rgb}{0.9, 0.17, 0.31}
\definecolor{antiquebrass}{rgb}{0.8, 0.58, 0.46}
\definecolor{antiquefuchsia}{rgb}{0.57, 0.36, 0.51}
\definecolor{chromeyellow}{rgb}{0.31, 0.47, 0.26}

\newcommand{\1}{\mathds 1}

\usepackage{amsmath,amsfonts,bm}

\def\eqref#1{equation~\ref{#1}}

\def\1{\bm{1}}

\DeclareMathAlphabet{\mathsfit}{\encodingdefault}{\sfdefault}{m}{sl}
\SetMathAlphabet{\mathsfit}{bold}{\encodingdefault}{\sfdefault}{bx}{n}

\tcbset{
    positive/.style={
        colback=custom_blue!5!white, 
        colframe=custom_blue!60!black, 
        coltitle=custom_blue!50!black, 
        colbacktitle=custom_blue!20!white, 
        fonttitle=\bfseries\centering, 
        title=Positive Steering,
        rounded corners,
        boxrule=0.5mm,
        enhanced,
        before skip=5mm, 
        after skip=5mm,  
    },
    neutral/.style={
        colback=gray!5!white, 
        colframe=black!50!white, 
        coltitle=black!50, 
        colbacktitle=black!10!white, 
        fonttitle=\bfseries\centering, 
        title=Neutral Response,
        rounded corners,
        boxrule=0.5mm,
        enhanced,
        before skip=5mm,
        after skip=5mm,
    },
    negative/.style={
        colback=custom_orange!5!white, 
        colframe=custom_orange!60!black, 
        coltitle=custom_orange!50!black, 
        colbacktitle=custom_orange!20!white, 
        fonttitle=\bfseries\centering, 
        title=Negative Steering,
        rounded corners,
        boxrule=0.5mm,
        enhanced,
        before skip=5mm,
        after skip=5mm,
    },
    dataexamplebox/.style={
        colframe=ccbgc_2,
        colback=ccbgc,
        coltitle=black,
        fonttitle=\small\bfseries,
        colbacktitle=ccbgc_2,
        title=Example: Myopic Alice/Bob,
        rounded corners,
        boxrule=0.5mm,
        arc=1mm,
        width=\columnwidth,
        top=1mm,
        bottom=1mm,
        left=1mm,
        right=1mm,
        fontupper=\footnotesize
    }
}

\newcommand\pythonstyle{\lstset{
basicstyle=\ttfamily\footnotesize,
language=Python,
morekeywords={self, clip, exp, mse_loss, uniform_sample, concatenate, logsumexp},              %
keywordstyle=\color{deepblue},
emph={MyClass,__init__},          %
emphstyle=\color{deepred},    %
stringstyle=\color{deepgreen},
frame=single,                         %
showstringspaces=false
}}

\lstnewenvironment{python}[1][]
{
\pythonstyle
\lstset{#1}
}
{}

\newcommand\pythoninline[1]{{\pythonstyle\lstinline!#1!}}

\makeatletter
\def\mathcolor#1#{\@mathcolor{#1}}
\def\@mathcolor#1#2#3{%
  \protect\leavevmode
  \begingroup
    \color#1{#2}#3%
  \endgroup
}
\makeatother

\tcbset{
  aibox/.style={
    width=474.18663pt,
    top=7pt,
    bottom=5pt,
    colback=blue!6!white,
    colframe=black,
    colbacktitle=black,
    enhanced,
    center,
    attach boxed title to top left={yshift=-0.1in,xshift=0.15in},
    boxed title style={boxrule=0pt,colframe=white,},
  }
}
\newtcolorbox{AIbox}[2][]{aibox,title=#2,#1}

\Crefformat{equation}{#2Eq.\;(#1)#3}

\Crefformat{figure}{#2Figure #1#3}
\Crefformat{assumption}{#2Assumption #1#3}
\Crefname{assumption}{Assumption}{Assumptions}

\usepackage{crossreftools}
\makeatletter
\renewcommand\footnoterule{%
  \kern 15\p@
  \hrule \@width 2in \kern 2.6\p@ 
  \vspace{4pt}
}
\makeatother

\title{Mixture-of-Recursions: Learning Dynamic Recursive Depths for Adaptive Token-Level Computation}

\reportnumber{} %

\author[1,\footnotesize{\textbf{*}}]{Sangmin Bae}
\author[1,\footnotesize{\textbf{*}}]{Yujin Kim}
\author[2,\footnotesize{\textbf{*}}]{Reza Bayat}
\author[1]{\\Sungnyun Kim}
\author[3]{Jiyoun Ha}
\author[4]{Tal Schuster}
\author[4]{Adam Fisch}
\author[5]{Hrayr Harutyunyan}
\author[4]{Ziwei Ji}
\author[2,6,\footnotesize{$\dagger$}]{\\Aaron Courville}
\author[1,\footnotesize{$\dagger$}]{Se-Young Yun}

\affil[1]{KAIST AI}
\affil[2]{Mila}
\affil[3]{Google Cloud}
\affil[4]{Google DeepMind}
\affil[5]{Google Research}
\affil[6]{Université de Montréal}

\correspondingauthor{Correspondence to: \email{\{bsmn0223,yujin399,yunseyoung\}@kaist.ac.kr}, \email{\{reza.bayat,courvila\}@mila.quebec}.}

\begin{abstract}
\textbf{Abstract:} Scaling language models unlocks impressive capabilities, but the accompanying computational and memory demands make both training and deployment expensive. Existing efficiency efforts typically target either parameter sharing or adaptive computation, leaving open the question of how to attain both simultaneously. We introduce \textit{Mixture-of-Recursions} (MoR), a unified framework that combines the two axes of efficiency inside a single Recursive Transformer. MoR reuses a shared stack of layers across recursion steps to achieve parameter efficiency, while lightweight routers enable adaptive token-level thinking by dynamically assigning different recursion depths to individual tokens. This allows MoR to focus quadratic attention computation only among tokens still active at a given recursion depth, further improving memory access efficiency by selectively caching only their key-value pairs. Beyond these core mechanisms, we also propose a KV sharing variant that reuses KV pairs from the first recursion, specifically designed to further decrease memory footprint. Across model scales ranging from 135M to 1.7B parameters, MoR forms a new Pareto frontier: at equal training FLOPs and smaller model sizes, it significantly lowers validation perplexity and improves few-shot accuracy, while delivering higher throughput compared with vanilla and existing recursive baselines. These gains demonstrate that MoR is an effective path towards large-model quality without incurring large-model cost.\looseness=-1
\end{abstract}

\begin{document}

\maketitle

\section{Introduction}
\label{sec:intro}

\begin{figure}[h]
    \vspace{-7pt}
    \centering
    \includegraphics[width=0.23\columnwidth]{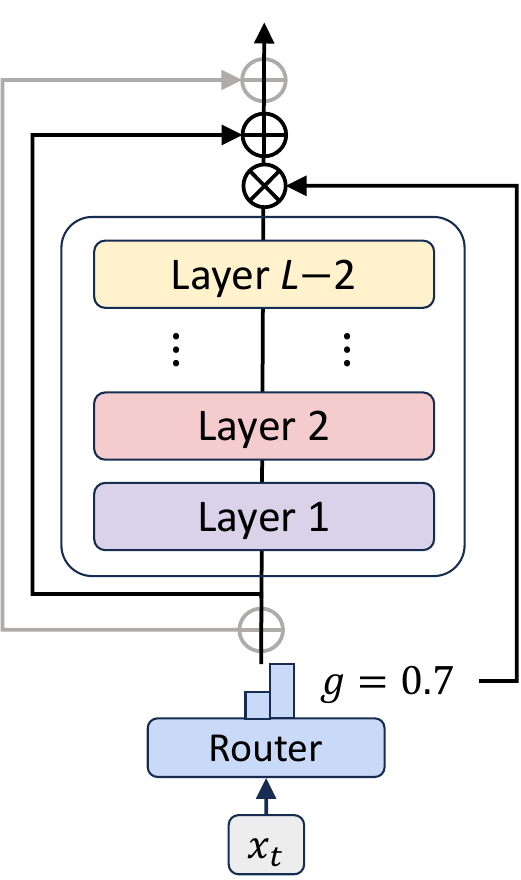}
    \hspace{23pt}
    \includegraphics[width=0.238\columnwidth]{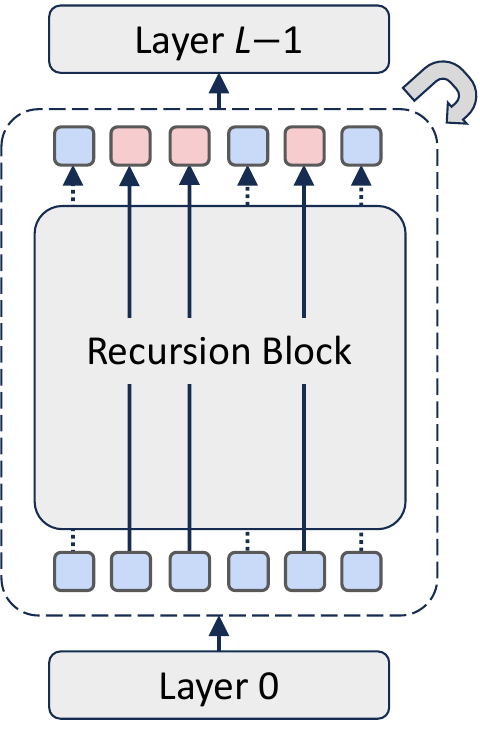}
    \hspace{7pt}
    \includegraphics[width=0.43\columnwidth]{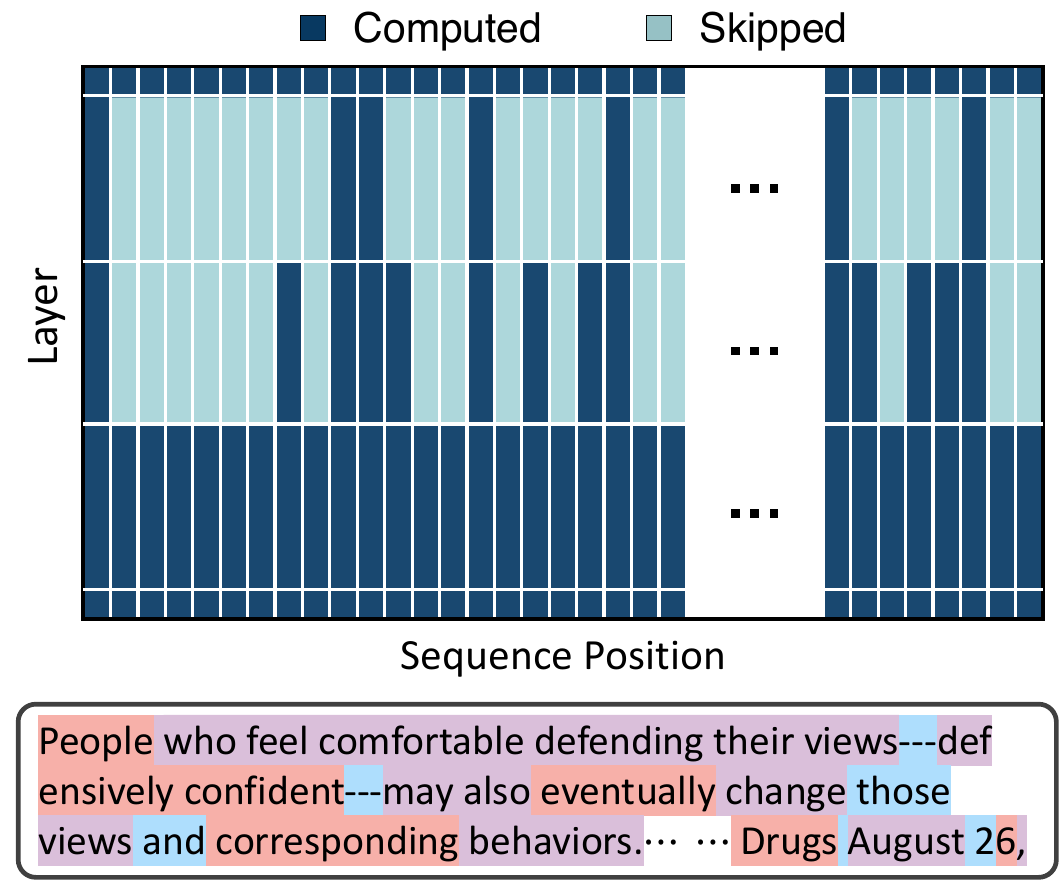}
    \caption{
    Overview of Mixture-of-Recursions (MoR). (\textit{Left}) Each recursion step consists of a fixed stack of layers and a router that determines whether each token should pass through or exit. This recursion block corresponds to the gray box in the middle.
    (\textit{Middle}) The full model structure, where the shared recursion step is applied up to $N_r$ times for each token depending on the router decision.
    (\textit{Right}) An example routing pattern showing token-wise recursion depth, where darker cells indicate active computation through the recursion block. Below shows the number of recursion steps that each subword token undergoes to predict the \textit{next} token, shown in colors: \raisebox{0pt}[0.5em][0.1em]{\colorbox{DarkOrchid!25}{1}}, \raisebox{0pt}[0.5em][0.1em]{\colorbox{SkyBlue!55}{2}}, and \raisebox{0pt}[0.5em][0.1em]{\colorbox{Salmon!60}{3}}. 
    }
    \label{fig:mor_overview}
\end{figure}

Scaling Transformer networks to hundreds of billions of parameters has unlocked impressive few-shot generalization and reasoning abilities~\citep{brown2020language, chowdhery2023palm, grattafiori2024llama, openai2023gpt4, reid2024gemini, deepseek-ai2024deepseek0v3, comanici2025gemini25pushingfrontier}.
However, the accompanying memory footprint and computational requirements make both training and deployment outside hyperscale data centers challenging~\citep{patterson2021carbon, momeni2024training}.
This has motivated researchers to seek alternative ``efficient'' designs~\citep{tay2022efficient, wan2023efficient}. 
Among the different axes of efficiency, \textit{parameter efficiency}~\citep{dehghani2018universal, bae2024relaxed, shazeer2017outrageously, fedus2022switch, lecun1989optimal}—reducing or sharing model weights—and \textit{adaptive computation}~\citep{raposo2024mixture, schuster2022confident, fedus2022switch, leviathan2023fast}—spending more compute only when it is needed—are promising, actively studied research directions.

One proven route to parameter efficiency is \textit{layer tying}, in which a shared set of weights is reused across multiple layers~\citep{dehghani2018universal, lan2019albert0, takase2021lessons, gholami2023generative, bae2024relaxed}.
For adaptive computation, a common approach is \textit{early-exiting}, which dynamically allocates compute by exiting  earlier in the network when predicting simpler tokens~\citep{elhoushi2024layerskip, schuster2022confident, Elbayad2020Depth-Adaptive, bae-etal-2023-fast}.
Despite the progress achieved along each of these individual efficiency axes, an architecture that effectively unifies both parameter efficiency and adaptive computation is still missing.
Recursive Transformers~\citep{bae2024relaxed, fan2024looped, giannou2023looped, yang2023looped, saunshi2025reasoning, geiping2025scaling, aleksandrov2025abbie}, models that repeatedly apply the same set of shared layers multiple times, offer a strong foundation due to their built-in weight sharing.
However, prior attempts at dynamic recursion have often been constrained by practical hurdles, such as requiring additional specialized training procedures or facing challenges in efficient deployment. This has led most approaches to still default to a simpler fixed-depth recursion, which applies the same amount of computation to every token and is thus incapable of delivering truly adaptive token-level compute allocation.

In this work, we introduce \textit{Mixture-of-Recursions} (MoR), a unified framework that fully leverages the potential of Recursive Transformers (see \autoref{fig:mor_overview}).
MoR trains lightweight routers end-to-end to assign token-specific recursion depths: it decides how many times a shared parameter block is applied to each token according to its required depth of ``thinking'', thereby directing computation to where it is most needed. 
This dynamic, token-level recursion inherently facilitates recursion-wise key–value (KV) caching, selectively storing and retrieving key–value pairs corresponding to each token’s assigned recursion depth. This targeted caching strategy reduces memory traffic, thereby improving throughput without relying on post-hoc modifications.
Therefore, MoR simultaneously (i) ties weights to cut parameters, (ii) routes tokens to cut redundant FLOPs, and (iii) caches key-values recursion-wise to cut memory traffic—all inside a single architecture.

Conceptually, MoR provides a \textit{pre-training} framework for latent space reasoning—performing non-verbal thinking by iteratively applying a single parameter block~\citep{hao2024training, geiping2025scaling, goyal2023think}. However, unlike approaches that deliberate on augmented continuous prompts before generation~\citep{liu2024deliberation, goyal2023think, hao2024training, shen2025codi}, MoR enables this latent thinking directly during the decoding of each token~\citep{zelikman2024quiet}. Furthermore, the routing mechanism facilitates adaptive reasoning along the model's vertical axis\footnote{While this thinking occurs along the depth axis, it is analogous to generating continuous thoughts along the horizontal sequence axis.}, moving beyond the uniform, fixed thinking depth common in prior work~\citep{geiping2025scaling, tack2025llm}. In essence, MoR enables models to efficiently adjust their thinking depth on a per-token basis, unifying parameter efficiency with adaptive computation.\looseness=-1

\vspace{-6pt}
\paragraph{Contributions.}
In summary, our key contributions in this paper are as follows.
\vspace{-3pt}
\begin{itemize}[leftmargin=*, itemsep=2pt]
    \item \textbf{Unified framework for efficient language modeling:} We present \emph{Mixture-of-Recursions} (MoR), the first architecture to unify efficiency paradigms—parameter sharing\,(\S\ref{subsec:preliminary}), token-level adaptive thinking depth\,(\S\ref{subsubsec:routing}), and memory-efficient KV caching\,(\S\ref{subsubsec:kv_cache})—within a single framework.

     \item \textbf{Dynamic recursion routing:} We introduce a router trained from scratch to assign dynamic per-token recursion depths. This aligns training with inference-time behavior and eliminates the need for costly, performance-degrading post-hoc routing stages used in conventional early-exit methods.

    \item \textbf{Extensive empirical validation:} Across models from 135M to 1.7B parameters\footnote{These are base model sizes, while MoR models have fewer unique parameters due to parameter sharing.} under equal compute budgets, MoR establishes a new Pareto frontier by improving validation loss and few-shot accuracy relative to vanilla and recursive baselines (\S\ref{subsec:main_results}, \S\ref{subsec:scaling_laws}).
    
    \item \textbf{Efficient architecture:} MoR dramatically reduces training FLOPs by selectively engaging only essential sequences in attention operations. Simultaneously, reduction in KV cache sizes leads to enhanced inference throughput itself, further boosted by continuous depth-wise batching (\S\ref{subsec:efficiency}).
    
\end{itemize}

\section{Method}

\subsection{Preliminary}\label{subsec:preliminary}

\paragraph{Recursive Transformers.}
The standard Transformer \citep{vaswani2017attention} constructs token representations through a stack of $L$ unique layers, each with a self-attention and a feed-forward network. At time step $t$, the hidden state $h$ evolves as: $\mathbf{h}_t^{\ell+1} = f\bigl(\mathbf{h}_t^{\ell}; \Phi_\ell\bigr),$ where $\ell = 0,\dots,L{-}1$ and $\Phi_\ell$ represents the parameters of the $\ell$-th layer.
Recursive Transformers~\citep{bae2024relaxed, fan2024looped, giannou2023looped, yang2023looped, saunshi2025reasoning} aim to reduce parameter count by reusing layers across depth. Instead of having $L$ distinct sets of weights, they partition the model into $N_r$ recursion \emph{blocks}, where each block uses a shared pool of parameters $\Phi'$. This design allows for more computation (by increasing the effective network depth) without increasing parameter size.

\newcolumntype{M}[1]{>{\centering\arraybackslash}m{#1}}

\begin{table*}[t]
\centering
\caption{
Parameter-sharing strategies in Recursive Transformers. This table shows \textit{Cycle} and \textit{Middle-Cycle} schemes with cyclic layer reuse, where Middle-Cycle retains unique first and last layers. 
}

\label{tab:sharing_strategy}
\renewcommand{\arraystretch}{1.1}
\resizebox{\textwidth}{!}{%
\setlength{\tabcolsep}{7pt}
\begin{tabular}{
    l
  | M{0.32\textwidth} 
  | M{0.16\textwidth} 
  | M{0.32\textwidth} 
  | >{\raggedright\arraybackslash}m{0.16\textwidth} 
}
\toprule
& \multicolumn{2}{c|}{\textbf{Cycle Strategy}} & \multicolumn{2}{c}{\textbf{Middle-Cycle Strategy}}\\
\cmidrule(l{2pt}r{2pt}){2-3}\cmidrule(l{2pt}r{2pt}){4-5}
\textbf{Layers} & Equation & Figure & Equation & \multicolumn{1}{M{0.2\textwidth}}{Figure} \\
\midrule
Last & -- & & $f\bigl(\mathbf{h}_t^{L-1}; \Phi_{L-1}\bigr)$ &\\[14pt]
Recursion & $f\!\Bigl(\mathbf{h}_t^{\ell}; \Phi'_{\ell \bmod (L/N_r)}\Bigr)$ & \multirow{-2}{*}{\includegraphics[width=0.91\linewidth]{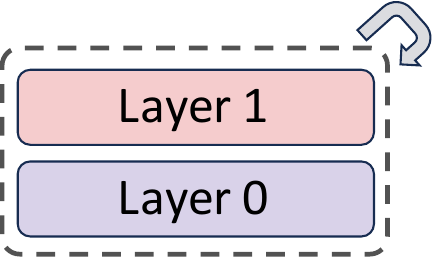}}
  & $f\!\Bigl(\mathbf{h}_t^{\ell}; \Phi'_{(\ell - 1 \bmod ((L-2)/N_r)) + 1}\Bigr)$ & \multirow{-3}{*}{\includegraphics[width=0.91\linewidth]{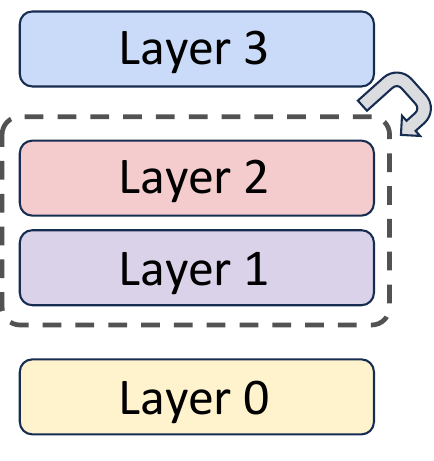}}  \\[14pt]
First   & --                  &   & $f\bigl(\mathbf{h}_t^{0}; \Phi_0\bigr)$ \\
[4pt]
\bottomrule
\end{tabular}
}
\end{table*}

\vspace{-10pt}
\paragraph{Parameter-sharing strategies.}
We examine four parameter-sharing strategies: \textit{Cycle}, \textit{Sequence}, and their variants \textit{Middle-Cycle} and \textit{Middle-Sequence}. \autoref{tab:sharing_strategy} summarizes two main designs, and the full list is provided in \autoref{tab:sharing_strategy_middle} in the Appendix. In Cycle sharing, recursion blocks are reused cyclically. For example, consider an original non-recursive model with $L{=}9$ layers and its recursive counterpart using $N_r{=}3$ recursions. Under the ``Cycle'' strategy, the layers are shared and unrolled as [$(0,1,2),(0,1,2),(0,1,2)$]. In ``Sequence'' sharing, each recursion block reuses the same layer consecutively before moving to the next, resulting in [$(0,0,0),(1,1,1),(2,2,2)$] for the same configuration. Both have the same effective number of layers when unrolled ($L{=}9$), but with a different order. Furthermore, the ``Middle'' variants preserve full-capacity parameters at the first and last layers ($\Phi_0$ and $\Phi_{L-1}$), while sharing weights among the intermediate layers.

\vspace{-10pt}
\paragraph{Enhanced training and inference efficiency in recursive models.}

Parameter sharing strategies can reduce the number of unique trainable parameters by a factor of the recursion number, effectively amortizing the memory footprint of the model. From a distributed training perspective, this becomes highly efficient when using Fully Sharded Data Parallel (FSDP)~\citep{zhao2023pytorch}. While a single \texttt{all-gather} operation would only support one iteration previously (i.e., 1 iter/gather), a recursive model reuses the same gathered parameters across all recursive steps (i.e., $N_r$ iter/gather). Furthermore, recursive architectures enable a novel inference paradigm, continuous depth-wise batching~\citep{bae2024relaxed, hooper2023speed}. This technique allows tokens at different stages to be grouped into a single batch, as they all utilize the same block of parameters. This can eliminate the bubbles—idle periods spent waiting for other samples to complete—thereby leading to significant throughput gains.

\vspace{-10pt}
\paragraph{Limitations in prior works.}

Although model parameters are tied, the distinct KV caches are typically used for each depth. This design fails to reduce the cache sizes, meaning the high retrieval latency still remains a severe inference bottleneck. Moreover, most existing recursive models simply apply a fixed recursion depth to all tokens, ignoring the varying complexity. While post-hoc methods like early-exiting methods can introduce some adaptivity, they often require separate training phases that can degrade performance~\citep{schuster2022confident, elhoushi2024layerskip, bae2024relaxed}. Ideally, the recursion depth should be learned dynamically during pretraining, allowing the model to adapt its computational path to each token's difficulty in a data-driven manner. However, such dynamic paths introduce a new challenge: exited tokens will have missing KV pairs at subsequent recursion depths. Addressing this would require a parallel decoding mechanism~\citep{bae-etal-2023-fast, elhoushi2024layerskip, kim2023speculative} to efficiently compute the actual KV pairs, but this requires separate, complex engineering and complicates the system.

\subsection{Mixture-of-Recursions}
\label{subsec:mor}

We propose \textit{Mixture-of-Recursions} (MoR)—a framework that dynamically adjusts recursion step for each token during \textit{pretraining} and \textit{inference}. The core of MoR lies in two components: a routing mechanism that assigns token-specific recursion steps to adaptively concentrate computation on more challenging tokens, and a KV caching strategy that defines how KV pairs are stored and selectively utilized for attention at each recursive step.\looseness=-1

\begin{figure}[t!]
    \centering
    \begin{subfigure}[t]{0.31\textwidth}
    \centering
    \includegraphics[width=0.7\textwidth]{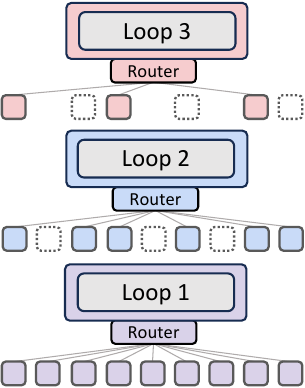}
        \centering
        \captionsetup{justification=centering}
        \subcaption{Expert-choice routing}
        \label{fig:method_overview-a}
    \end{subfigure}
    \hfill
    \begin{subfigure}[t]{0.41\textwidth}
    \centering
    \includegraphics[width=\textwidth]{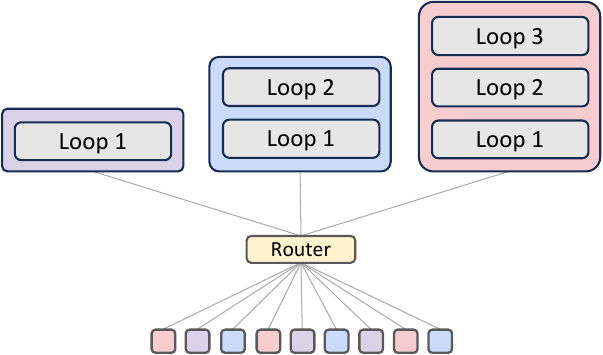}
        \centering
        \captionsetup{justification=centering}
        \subcaption{Token-choice routing}
        \label{fig:method_overview-b}
    \end{subfigure}
    \centering
    \hfill
    \begin{subfigure}[t]{0.245\textwidth}
        \centering\includegraphics[width=0.65\textwidth]{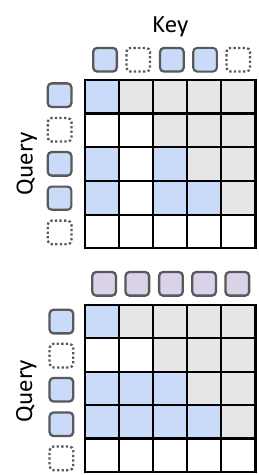}
        \captionsetup{justification=centering}
        \subcaption{Caching mechanism}
        \label{fig:method_overview-c}
    \end{subfigure}
    \caption{
        Architectural components of Mixture-of-Recursions (MoR).  
        (\textit{a}) {Expert-choice routing:} At each recursion step, a router selects top-$k$ tokens to continue, progressively narrowing the set of active tokens with depth.  
        (\textit{b}) {Token-choice routing:} Each token is assigned a fixed recursion step at the outset via a single routing decision, defining its complete compute path through the model.  
        (\textit{c}) {KV caching strategies:} Each square in the matrix represents whether a token (row) attends to another token’s cached key (column).  
        In ``recursion-wise KV caching'' (\textit{Top}), only the keys of currently selected (non-dropped) tokens at each recursion step are cached (\raisebox{0pt}[0.1em][0.1em]{\colorbox{custom_blue}{blue}}), and attention is restricted only to these entries.  
        In ``recursive KV sharing'' (\textit{Bottom}), all keys of previous tokens are cached at the first recursion step (\raisebox{0pt}[0.1em][0.1em]{\colorbox{custom_purple}{purple}}) and shared across subsequent recursion steps for attention operations.\looseness=-1
        }
    \label{fig:method_overview}
\end{figure}

\subsubsection{Routing Strategies: Expert-choice vs. Token-choice}
\label{subsubsec:routing}

\paragraph{Expert-choice routing. (\autoref{fig:method_overview-a})}
Inspired by top-$k$ gating in MoD models~\citep{raposo2024mixture}, 
in expert-choice routing, each recursion depth becomes an expert and selects their preferred top-$k$ tokens (e.g., for $N_r = 3$, we have three experts: Expert 1 applies the first recursion step, Expert 2 applies the second recursion step, and so on).
At each recursion step $r$, the corresponding router uses the hidden state $\mathcal{H}_t^r$ (input to the $r$-th recursion block) and its routing parameters $\theta_r$ to compute a scalar score $g_t^r = \mathcal{G}(\theta_r^\top \mathcal{H}_t^r$) for token $t$. Here, $\mathcal{G}$ represents an activation function like \texttt{sigmoid} or \texttt{tanh}.
Then, the top-$k$ tokens are selected to pass through the recursion block:\looseness=-1
\begin{align}
\mathcal{H}_t^{r+1} =
\begin{cases}
g_t^r f(\mathcal{H}_t^r,\, \Phi') + \mathcal{H}_t^r, & \text{if } g_t^r > P_\beta({G}^r) \\
\mathcal{H}_t^r, & \text{otherwise}
\end{cases}
\end{align}
where $P_\beta({G}^r)$ is the $\beta$-percentile threshold over all scores at recursion step $r$.

To ensure coherent progression through steps, we adopt \textit{hierarchical filtering}: only tokens selected at recursion step $r$ can be re-evaluated at $r{+}1$. This simulates early-exit behavior while learning from scratch. As deeper layers tend to encode increasingly abstract and sparse information~\citep{li2022lazy, yang2024pyramidinfer, nawrot2024dynamic}, this mechanism prioritizes computation for only the most demanding tokens.

\vspace{-10pt}
\paragraph{Token-choice routing. (\autoref{fig:method_overview-b})}  Unlike expert-choice, where token selection is made at each recursion step, token-choice commits each token to a full sequence of recursion blocks from the start.
Formally, given the hidden state $\mathcal{H}_t^1$ (in Middle-Cycle strategy, $\mathcal{H}_t^1 = h_t^1$), the router computes a non-linear function (\texttt{softmax} or \texttt{sigmoid}) over experts: ${g}_t = \mathcal{G}(\theta_r^\top \mathcal{H}_t^1)$, where $g_t^j$ denotes the routing score for expert $j \in \{1,\dots,N_r\}$. The token is assigned to expert $i = \arg\max_j g_t^j$ (top-1 gating), which corresponds to sequentially applying the recursion $i$ times. The hidden state is then updated recursively as:
\begin{align}
\mathcal{H}_t^{r+1} =
\begin{cases}
g_t^r f(\mathcal{H}_t^r,\, \Phi') + \mathcal{H}_t^1, & \text{if } r = i  \\
g_t^r f(\mathcal{H}_t^r,\, \Phi'), & \text{otherwise} \\
\end{cases}
\end{align}

To compare routing strategies under equal compute, we align the token allocation budgets of expert-choice with that of token-choice. Specifically, we calibrate token capacity (i.e., top-\textit{k}) of expert-choice to match the expected token distribution of token-choice routing with perfect load balancing. 
In perfectly balanced token-choice, each token is assigned to recursion depth $i \in \{1, \dots, N_r\}$ with equal probability $1/N_r$. Thus, recursion step $j$ processes a fraction $(N_r - j + 1)/N_r$ of the tokens. For example, when $N_r = 3$, recursion steps 1, 2, and 3 handle $\{3/3, 2/3, 1/3\}$ of tokens, respectively. Therefore, we apply this same fractional allocation in the top-$k$ selection of the expert-choice routing (i.e., $k$ is sequenced like $N_r / N_r$, $\cdots$, $1 / N_r$ over $N_r$ recursion steps).\looseness=-1

\vspace{-12pt}
\paragraph{Strengths and limitations. (\autoref{tab:method_comparison}--\textit{Left})} 
Although expert-choice routing guarantees perfect load balancing with static top-$k$ selection, it suffers from information leakage~\citep{zhou2022mixture, wang2024auxiliary, raposo2024mixture}. This violation of causality during training forces to exploit an auxiliary router or a regularization loss~\citep{zhou2022mixture, raposo2024mixture}, aiming to precisely detect top-$k$ tokens at inference without access to future token information. Meanwhile, token-choice is free from such leakage, but typically requires a balancing loss or loss-free algorithms~\citep{wang2024auxiliary, fedus2022switch, zoph2022st} due to its inherent load balancing challenges. We explore each of these components for MoR in further detail (\S\ref{subsec:ablation_routing}).

\subsubsection{KV Caching Strategies: Recursion-wise Caching vs. Recursive Sharing}
\label{subsubsec:kv_cache}

Dynamic-depth models often struggle with KV cache consistency during autoregressive decoding. When a token exits early, its corresponding keys and values in deeper layers will be missing, which can be crucial information for subsequent tokens.
Some methods attempt to reuse stale entries~\citep{schuster2022confident} or run parallel decoding~\citep{bae-etal-2023-fast}, but these solutions still introduce overhead and complexity. To this end, we design and explore two KV cache strategies tailored to MoR models: \emph{recursion-wise caching} and \emph{recursive sharing}.

\vspace{-12pt}
\paragraph{Recursion-wise KV caching. (\autoref{fig:method_overview-c}--\textit{Top})}

Inspired by \citet{raposo2024mixture}, we cache KV pairs selectively: only tokens routed to a given recursion step store their key–value entries at that level. Thereby, the KV cache size at each recursion depth is determined exactly by the capacity factor in expert-choice,  or according to actual balancing ratios in token-choice. Attention is then restricted to those locally cached tokens. This design promotes block-local computation, which improves memory efficiency and reduces IO demands.

\vspace{-12pt}
\paragraph{Recursive KV sharing. (\autoref{fig:method_overview-c}--\textit{Bottom})}

A key design choice for our MoR model is that all tokens traverse at least the first recursion block\footnote{Though this is not a strict requirement of the MoR framework itself.}. We leverage this by caching KV pairs exclusively at this initial step and reusing them across all subsequent recursions. Therefore, the query length might get shorter at each recursion depth based on the selection capacity, but the key and value lengths will consistently maintain the full sequence. This ensures that all tokens can access to past context without recomputation, despite any distribution mismatch.\looseness=-1

\begin{table*}[t!]
    \caption{
    Comparison of routing strategies and key-value caching strategies.
    (\textit{Left}) Summary of two routing strategies: expert-choice and token-choice, highlighting their pros, cons, and mitigating solutions from previous works~\citep{raposo2024mixture, wang2024auxiliary, zoph2022st}. 
    (\textit{Right}) Relative cost efficiency of caching strategies against a vanilla Transformer (normalized to 1). Here, $N_r$ denotes the number of recursions, and $k$ ($k < N_{\mathrm{ctx}}$) denotes the number of selected tokens per layer. We compare only the recursion blocks, excluding the non-shared layers. KV cache memory and IO are measured across the entire model, whereas attention FLOPs are reported per layer.
    \looseness=-1
    }
    \label{tab:method_comparison}
    \small
    \begin{subtable}[t]{0.478\textwidth}
    \renewcommand{\arraystretch}{1.1}
    \centering
    \resizebox{\linewidth}{!}{
    \setlength{\tabcolsep}{7pt}
    \begin{tabular}{l|c|c}
    \toprule
      &  \textbf{Expert-choice} & \textbf{Token-choice} \\
    \midrule
    Pros & Static\,compute\,budget & No\,leakage \\[3pt]
    Cons & Causality\,violation  & Load\,imbalance\\[1pt]
    $\llcorner$\,Sol & Aux\,Rout, Aux\,Loss & Bal\,Loss, Loss-free \\
    \bottomrule
    \end{tabular}
    }
    \end{subtable}
    \hspace{-4pt}
    \begin{subtable}[t]{0.522\textwidth}
    \renewcommand{\arraystretch}{1.1}
    \centering
    \resizebox{\linewidth}{!}{
    \setlength{\tabcolsep}{3.5pt}
    \begin{tabular}{l|c|c}
    \toprule
     & \textbf{\,Recursion-wise\,Caching\,} & \textbf{\,Recursive\,Sharing\,}\\
    \midrule
    KV\,Memory  & $(N_r + 1)/2N_r$ & $1/N_r$ \\[3pt]
     KV\,Cache\,IO  & $(N_r + 1) / 2N_r$ & 1 \\[3pt]
    Attn\,FLOPs  & $k^2/N_{\mathrm{ctx}}^2$  & $k/ N_{\mathrm{ctx}}$ \\
    \bottomrule
    \end{tabular}
    }
    \end{subtable}
    \vspace{-3pt}
\end{table*}

\vspace{-12pt}
\paragraph{Strengths and limitations. (\autoref{tab:method_comparison}--\textit{Right})}

Recursion-wise caching cuts KV memory and IO to approximately $(N_r + 1)/2N_r$ times across the entire model (when assuming capacity factors follow a sequence like $N_r / N_r$, $\cdots$, $1 / N_r$ over $N_r$ recursion steps). It also reduces per-layer attention FLOPs to a factor of $(k / N_{\text{ctx}})^2$ of those in vanilla models, resulting in substantially improved efficiency for both training and inference phases.
Meanwhile, recursive sharing can yield maximal memory savings by globally reusing context. Speedups can be further achieved by skipping KV projection and prefill at shared depths (compatible only with Cycle strategy~\citep{sun2024you}). However, attention FLOPs only decrease by $k / N_{\text{ctx}}$, and high volume of KV IO still leads to a decoding bottleneck.\looseness=-1

\section{Experiments}

We pretrain our models from scratch using a Llama-based Transformer architecture\footnote{Experiments on Llama are conducted without direction or involvement from Google advisors.}~\citep{grattafiori2024llama}, referring to the configurations of SmolLM open-source models~\citep{allal2024SmolLM}, on a deduplicated subset of the FineWeb-Edu dataset~\citep{penedo2024the} in SmolLM-Corpus~\citep{benallal2024smollmcorpus}.
We evaluate the models on validation set of FineWeb-edu and six few-shot benchmarks~\citep{eval-harness}. Detailed training and evaluation procedures, as well as throughput measurement protocols, are described in \autoref{app:experimental_setup}.

\begin{table*}[!t]
    \vspace{-2pt}
    \caption{
    Comparison of MoR, Recursive, and Vanilla Transformers under both fixed FLOPs (16.5e18) and token (20B) settings. All models are trained on FineWeb-Edu and evaluated by validation negative log-likelihood\,(NLL) and few-shot accuracy. For the isoFLOP rows, the number of training tokens\,($N_{tok}$) varies by model efficiency. For the fixed-token rows, we report the effective FLOPs consumed.
    For the model sizes, we report non-embedding parameter counts. For the KV mechanisms, we distinguish between Cache (recursion-wise caching) and Share (recursive sharing). $\text{}^\dagger$In recursive models, all tokens go through fixed recursion depths ($N_r$), instead of adaptive depths.\looseness=-1
    }
    \label{tab:main_results}
    \small
    \centering
    \addtolength{\tabcolsep}{-1pt}
    \resizebox{\textwidth}{!}{
    \begin{tabular}{l|cc|cc|ccc|c|cccccc|c}
    \toprule
      &  \multicolumn{2}{c|}{\textbf{MoR}} &  \multicolumn{2}{c|}{\textbf{Recursion}} & \multicolumn{3}{c|}{\textbf{Pretrain}} & \textbf{NLL\,$\downarrow$} & \multicolumn{7}{c}{\textbf{Few-shot Accuracy\,$\uparrow$}} \\
    \cmidrule(l{2pt}r{2pt}){2-3} \cmidrule(l{2pt}r{2pt}){4-5} \cmidrule(l{2pt}r{2pt}){6-8} \cmidrule(l{2pt}r{2pt}){9-9} \cmidrule(l{2pt}r{2pt}){10-16} 
     \textbf{Models} & Type & KV & Share & $N_r$ & Param & FLOPs & $N_{tok}$ & FineWeb & LD & HS & PQ & WG & ARC & \!MMLU\! & Avg  \\
    \midrule
    Vanilla & - & - & - & -  & 315M &  16.5 & 20B & 2.7824 & 32.0 & 37.8 & 65.6 & 50.5 & 39.6 & 28.0 & 42.3 \\
    \midrule
    \multirow{3}{*}{$\text{Recursive}^{\dagger}$} & - & -  & M-Cyc  & 2 & 167M & 16.5& 20B  & 2.8079 & 31.0 & 37.1 & 66.7 & 52.3 & \textbf{40.8} & 27.5 & 42.6 \\
     & - & - & M-Cyc & 3 & 118M & 16.5 & 20B & 2.8466 & 29.8 & 35.9 & 65.0 & 52.3 & 39.0 & 27.2 & 41.5 \\
     & - & - & M-Cyc & 4 & \,\,\,98M & 16.5& 19B  & 2.8781  & 28.2 & 35.4 & 65.5 & {52.5} & 38.0 & 26.8 & 41.0 \\
     \midrule
     \rowcolor[gray]{0.9}
    \cellcolor{white}& Expert & Cache  & M-Cyc  & 2 & 167M & 16.5 & 27B & \textbf{2.7511} & \textbf{34.4} & \textbf{39.3} & 65.7 & 51.2 & 39.6 & \textbf{28.1} & \textbf{43.1} \\
    \rowcolor[gray]{0.9}
     \cellcolor{white}& Expert & Cache & M-Cyc & 3& 118M & 16.5 & 30B  & 2.7925 & 33.1 & 37.9 & \textbf{66.9} & 52.1 & 38.3 & 27.4 & 42.6 \\
     \rowcolor[gray]{0.9}
     \cellcolor{white}& Expert & Cache & M-Cyc & 4 & \,\,\,98M & 16.5 & 30B & 2.8204 & 30.1 & 37.3 & 65.0 & 51.1 & 38.9 & 27.4 & 41.6 \\
     \cmidrule(l{2pt}r{2pt}){2-16} 
     \rowcolor[gray]{0.9}
     \cellcolor{white}& Expert & Cache  & M-Cyc  & 2 & 167M & 12.3 & 20B & {2.7749} & {33.2} & {38.3} & 65.2 & \textbf{52.6} & 40.1 & \textbf{28.1} & {42.9} \\
      \cellcolor{white}& Expert & Cache & M-Cyc & 3& 118M & 11.0 & 20B  & 2.8246 & 31.9 & 37.0 & 65.7 & 50.5 & 38.3 & 27.4 & 41.8 \\
      \cellcolor{white}& Expert & Cache & M-Cyc & 4 & \,\,\,98M & 11.0 & 20B & 2.8519 & 30.2 & 36.5 & 64.3 & 52.3 & 38.6 & 27.2 & 41.5 \\
     \cmidrule(l{2pt}r{2pt}){2-16} 
     \cellcolor{white}& Token & Cache & M-Cyc & 3& 118M & 16.5  & 30B & 2.9163 & 27.6 & 34.1 & 63.8 & 50.6 & 37.4 & 26.8 & 40.0 \\
     \cmidrule(l{2pt}r{2pt}){2-16} 
     \cellcolor{white}\multirow{-9}{*}{MoR\,(ours)}
     & Expert & Share & M-Cyc & 3 & 118M & 16.5 & 31B & 2.7983 & 31.7 & 37.2 & 65.1 & 51.0 & 39.0 & 27.1  & 41.9 \\
    \bottomrule
    \end{tabular}
    }
    \vspace{-1pt}
\end{table*}

\subsection{Main Results}
\label{subsec:main_results}

\paragraph{MoR outperforms baselines with fewer parameters under equal train compute.}
Under an equal training budget of 16.5e18 FLOPs, we compared our Mixture-of-Recursions (MoR) model against both Vanilla and Recursive Transformers. As shown in \autoref{tab:main_results}, the MoR model, using an expert-choice router and two recursions, achieves a lower validation loss and surpasses the vanilla baseline in average few-shot accuracy (43.1\% vs. 42.3\%). Remarkably, this superior performance is achieved despite using nearly 50\% fewer parameters. This is attributed to MoR's higher computational efficiency, which allows it to process more training tokens within the same FLOPs budget. Furthermore, as $N_r$ increases to 3 or 4, MoR maintains its competitive accuracy, consistently outperforming the recursive baselines while remaining within a tight margin of the full-capacity vanilla model.\looseness=-1

\vspace{-10pt}
\paragraph{MoR outperforms baselines with less compute at equal data.}

To isolate architectural differences, we analyze performance under a fixed number of training tokens (20B). Specifically, our MoR model with $N_r=2$ outperforms both vanilla and recursive baselines—achieving lower validation loss and higher accuracy—despite using 25\% fewer training FLOPs. This theoretical efficiency translates into significant practical gains: compared to the vanilla baseline, our model reduces training time by 19\% and cuts peak memory usage by 25\%. These improvements stem from our hierarchical filtering and recursion-wise attention mechanism, which shortens sequence lengths to achieve a superior compute-accuracy trade-off, even during pretraining.

\vspace{-10pt}
\paragraph{MoR performance varies with routing and caching strategies.}

We also evaluate a few design variants within MoR framework, specifically with $N_r=3$ that is lightweight and still comparable with Vanilla. In this case, using token-choice routing yields lower performance (40.0\%) compared to expert-choice routing (42.6\%), indicating that routing granularity plays a pivotal role in model performance. Additionally, applying KV cache sharing slightly reduces performance compared to independent caching, while providing improved memory efficiency. This trade-off remains favorable for practical deployment when memory usage is a key concern.

\begin{figure}[t]
    \centering
    \begin{subfigure}[t]{\textwidth}
        \centering
        \captionsetup{justification=centering}
        \includegraphics[width=0.57\textwidth]{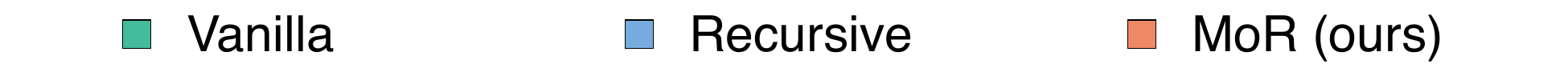}
    \end{subfigure}
    \centering
    \begin{subfigure}[t]{0.265\textwidth}
    \captionsetup{justification=centering}
        \includegraphics[width=\textwidth]{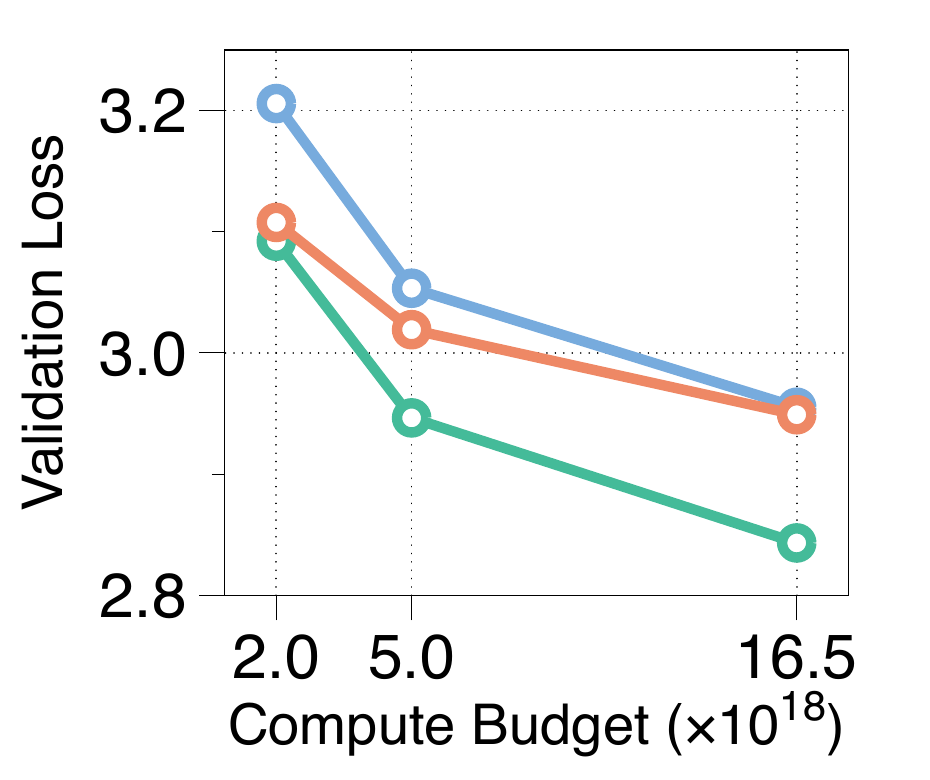}
        \subcaption{135M-based model}
    \end{subfigure}
    \hspace{-8pt}
    \centering
    \begin{subfigure}[t]{0.24\textwidth}
    \captionsetup{justification=centering}
        \includegraphics[width=\textwidth]{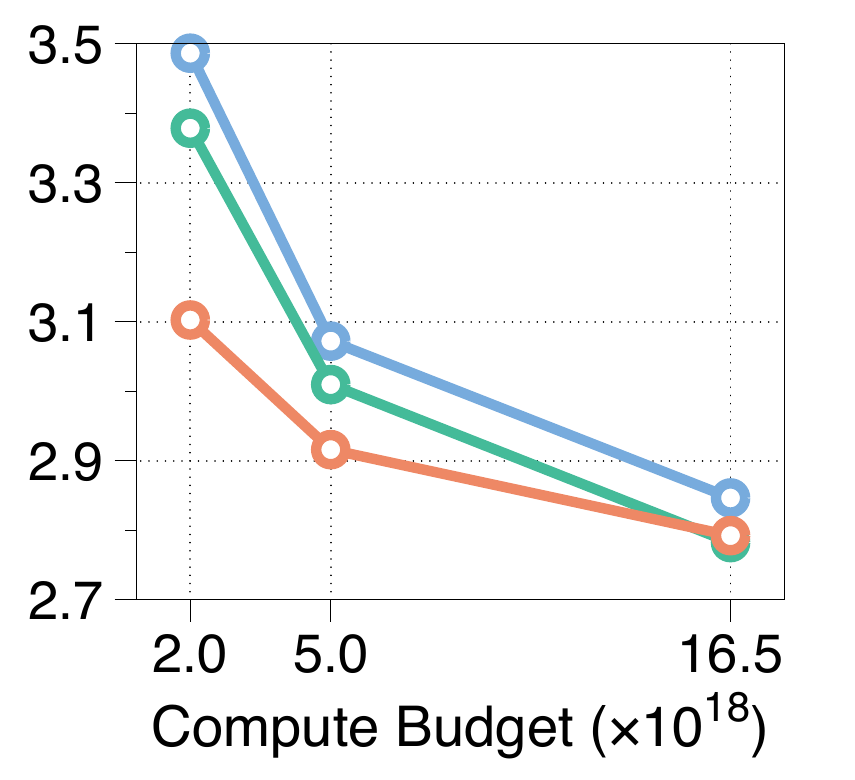}
        \subcaption{360M-based model}
    \end{subfigure}
    \hspace{-8pt}
    \centering
    \begin{subfigure}[t]{0.24\textwidth}
    \captionsetup{justification=centering}
        \includegraphics[width=\textwidth]{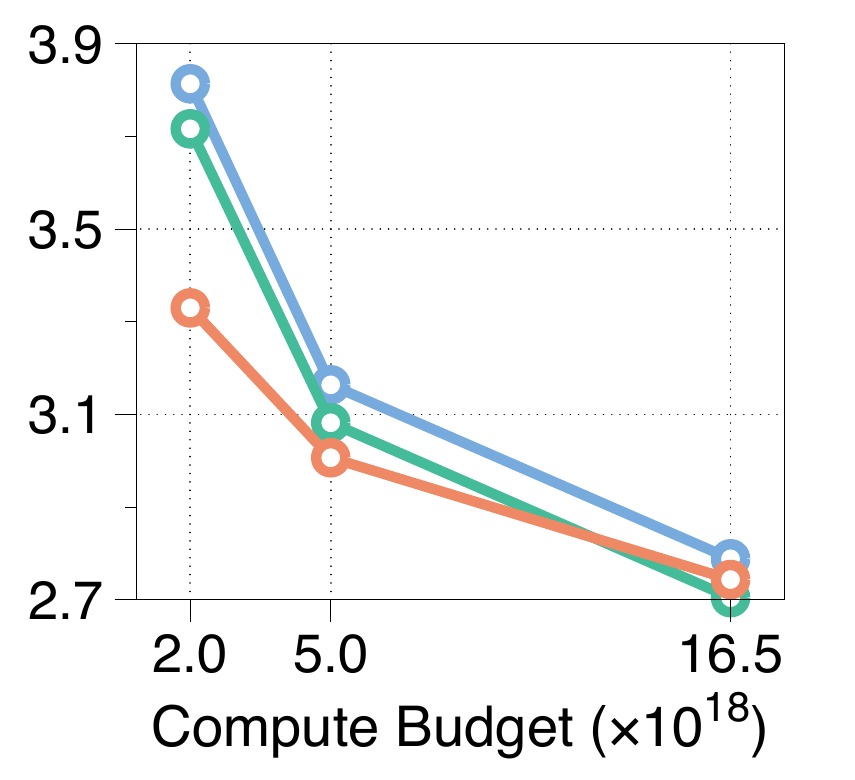}
        \subcaption{730M-based model}
    \end{subfigure}
    \hspace{-8pt}
    \centering
    \begin{subfigure}[t]{0.24\textwidth}
    \captionsetup{justification=centering}
        \includegraphics[width=\textwidth]{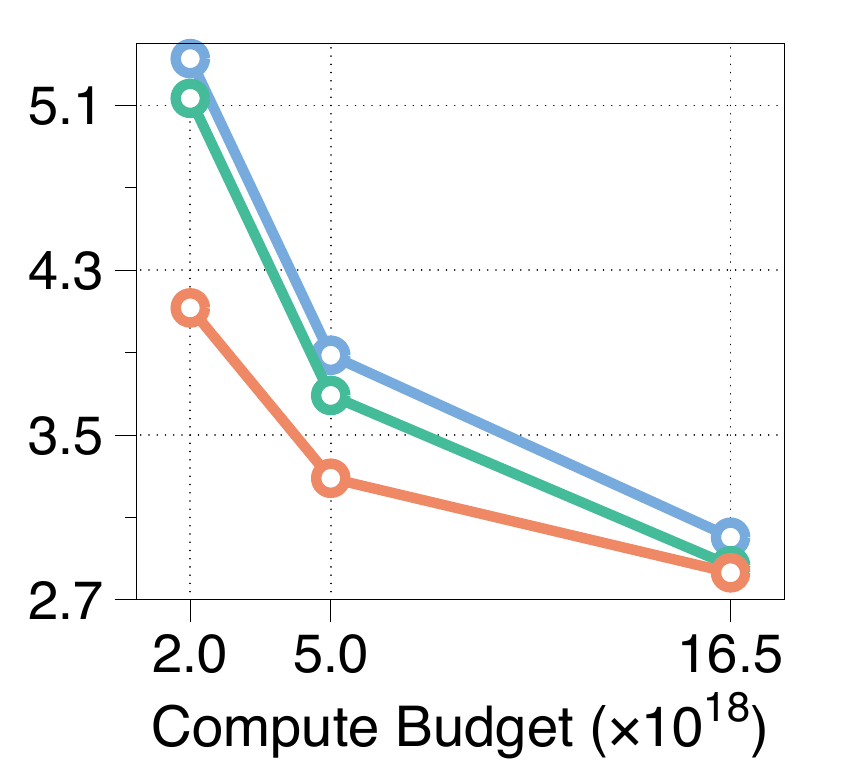}
        \subcaption{1.7B-based model}
    \end{subfigure}
    \caption{
    Validation loss across different compute budgets across four model sizes: 135M, 360M, 730M, and 1.7B parameters. For MoR models, we use expert-choice routing and recursion-wise caching. MoR consistently outperforms recursive baselines and matches or exceeds the standard Transformers at larger scales, despite using significantly fewer parameters (approximately one-third due to layer tying with $N_R=3$).\looseness=-1
    }
    \label{fig:scaling_laws_app}
\end{figure}

\subsection{IsoFLOP Analysis}
\label{subsec:scaling_laws}

A core criterion for evaluating a new model architectural design is whether performance continues to improve as model and compute scales grow~\citep{kaplan2020scaling}. Therefore, we evaluate MoR against both Vanilla and Recursive Transformers across a wide range of model sizes and computational budgets to show that it maintains competitive or superior predictive performance as the scale increases.

\vspace{-10pt}
\paragraph{Experimental Setup.}

We experiment with four scales—135M, 360M, 730M, and 1.7B parameters—fixing the number of recursions to three for both Recursive and MoR configurations, resulting in roughly one-third the number of unique parameters. Each model is pretrained under three FLOPs budgets: 2e18, 5e18, and 16.5e18.\looseness=-1

\vspace{-10pt}
\paragraph{MoR is a scalable and parameter-efficient architecture.}

As shown in \autoref{fig:scaling_laws_app}, MoR consistently outperforms recursive baselines across all model sizes and compute budgets. While it underperforms the vanilla model at the smallest model size (135M)—likely due to a recursive capacity bottleneck—this gap closes rapidly at scale. For >360M parameters, MoR not only matches but often exceeds the Vanilla Transformer, particularly under low and mid-range budgets. Overall, these results highlight that MoR is a scalable and efficient alternative to standard Transformers. It achieves strong validation performance with significantly lower parameter counts, making it a strong candidate for both pretraining and large‑scale deployment. Further details are presented in \autoref{app:scaling_laws}.\looseness=-1

\subsection{Inference Throughput Evaluation}
\label{subsec:efficiency}

As a parameter-shared architecture, MoR can leverage continuous depth-wise batching~\citep{bae2024relaxed} to dramatically boost inference throughput compared to Vanilla Transformers. This maintains high and consistent GPU utilization by immediately replacing completed sequences with incoming tokens during decoding. The early-exiting mechanism in MoR further eliminates bubbles in the computational batch.\looseness=-1

\vspace{-10pt}
\paragraph{Experimental Setup.}

We measure throughput for 360M scale-based MoR models with recursion depths of 2, 3, and 4, trained under a 16.5e18 FLOPs budget. Throughput (tokens/second) is measured based on the generation time for tokens per sample, where the number of tokens is sampled from a normal distribution with a mean of 256, starting without any input prefix. We examine two batching configurations: a fixed batch size of 32 and a (relative) maximum batch size derived by multiplying 32 by the ratio of the maximum batch sizes of vanilla and MoR models. Further details on the experimental setup are provided in \autoref{app:throughput}.\looseness=-1

\vspace{-10pt}
\paragraph{MoR boosts inference throughput with continuous depth-wise batching.}

\begin{figure}[t!]
    \centering
    \begin{subfigure}[t]{0.325\textwidth}
        \includegraphics[width=\textwidth]{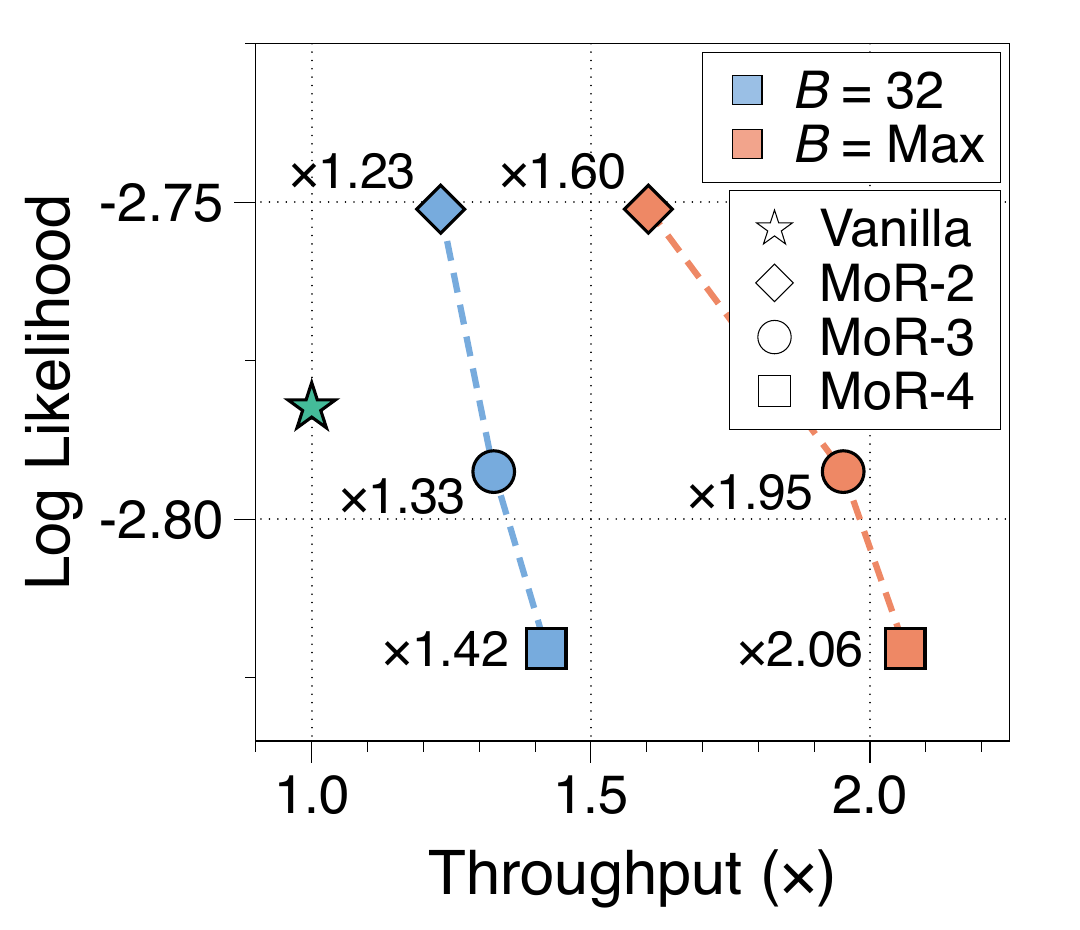}
        \captionsetup{justification=centering}
        \subcaption{Pareto frontier of throughput}
        \label{fig:pareto_frontier_sharing-a}
    \end{subfigure}
    \centering
    \begin{subfigure}[t]{0.325\textwidth}
        \includegraphics[width=\textwidth]{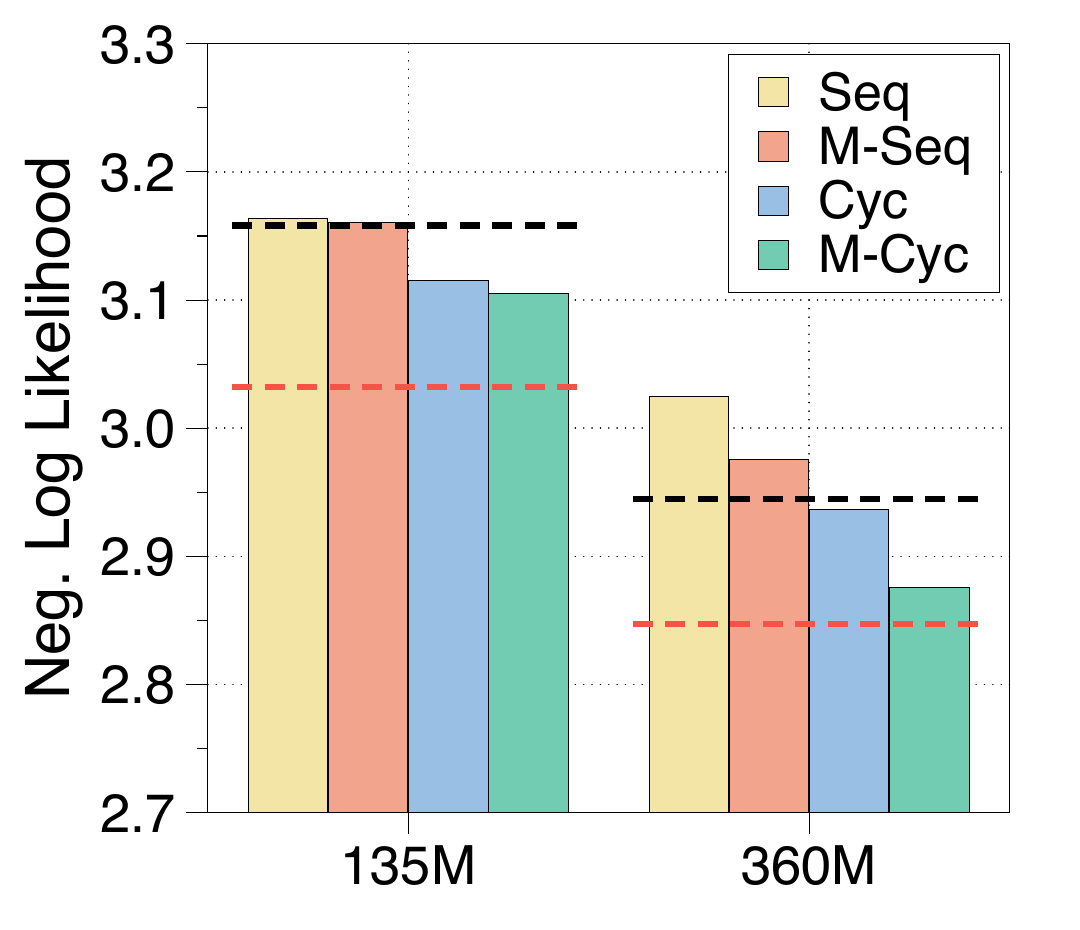}
        \captionsetup{justification=centering}
        \subcaption{Sharing strategy}
        \label{fig:pareto_frontier_sharing-b}
    \end{subfigure}
    \begin{subfigure}[t]{0.325\textwidth}
        \includegraphics[width=\textwidth]{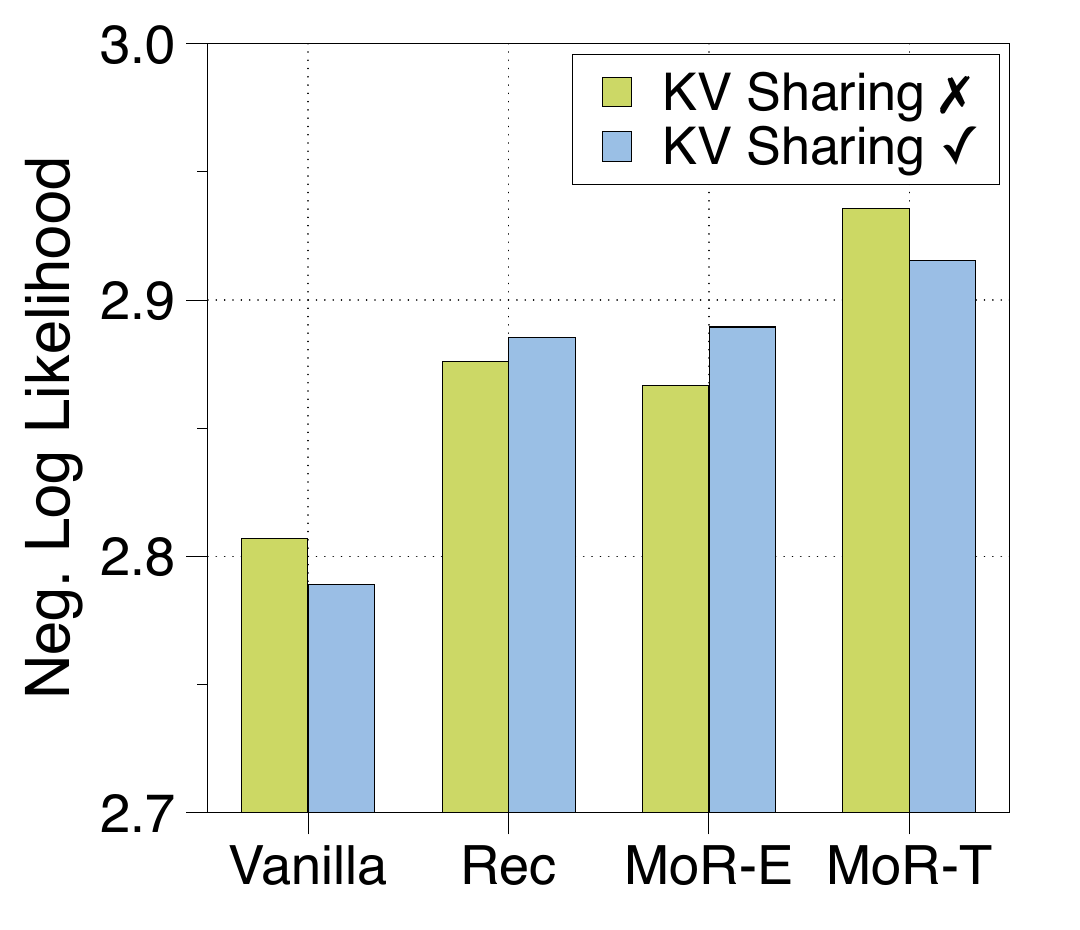}
        \captionsetup{justification=centering}
        \subcaption{KV cache sharing}
        \label{fig:pareto_frontier_sharing-c}
    \end{subfigure}
    \centering
    \caption{
    (a) Pareto frontier of inference throughput and log-likehood for MoR and Vanilla Transformer under fixed and maximum batching scenarios. Setting details are in \autoref{app:throughput}.
    (b) Negative log-likelihood (NLL) of Recursive Transformers with $N_r=3$ across four different parameter-sharing strategies. We pretrained the models on 10 billion tokens. The dashed red and black lines denote the full-size Vanilla Transformer and parameter-matched vanilla models (approximately one-third scales), respectively.
    (c) NLL performance comparison across four different architectures with KV sharing. 
    For MoR, green (disabled) and blue (enabled) refer to recursion-wise KV caching and recursive KV sharing strategies. MoR-E and MoR-T denotes expert-choice and token-choice MoR, respectively. All models are based on 360M scale and trained on 10 billion tokens.\looseness=-1
    }    
    \label{fig:pareto_frontier_sharing}
\end{figure}

In \autoref{fig:pareto_frontier_sharing-a}, across both batch settings, all MoR variants outperform the vanilla baseline, which even leverages continuous sequence-wise batching~\citep{yu2022orca, kwon2023efficient}. Increasing recursion depth leads to more tokens exiting early and a further reduction in KV cache usage. This, in turn, boosts throughput significantly (e.g., MoR-4 achieves up to a 2.06$\times$ speedup with $B=\text{Max}$). While there's a slight performance degradation, it can be a favorable trade-off given the substantial throughput gain. These results support that the integration of the depth-wise batching paradigm with early-exiting can significantly accelerate MoR's actual deployment throughput.\looseness=-1

\section{Ablation Studies}

\subsection{Parameter Sharing Strategies}
\label{subsec:sharing}

\paragraph{Middle-Cycle is the most effective parameter sharing strategy.}

As discussed in \S\ref{subsec:preliminary}, parameter sharing is a key component of Recursive Transformers and MoR. To identify the most effective sharing configuration, we empirically compare four aforementioned strategies: Cycle, Sequence, Middle-Cycle, and Middle-Sequence. We evaluate each strategy on Recursive Transformers based on 135M and 360M model sizes. As shown in \autoref{fig:pareto_frontier_sharing-b}, ``Middle-Cycle'' consistently achieves the lowest validation loss, and its superiority is further confirmed by the detailed results in \autoref{app:parameter_sharing}. Based on these findings, we adopt the ``Middle-Cycle'' configuration for all subsequent MoR and Recursive Transformers presented in this paper.\looseness=-1

\subsection{Routing Strategies}
\label{subsec:ablation_routing}

We conduct an extensive ablation study to understand the impact of various design choices for expert-choice and token-choice routing in our MoR framework. Detailed results are summarized in \autoref{app:router_ablation}.
\looseness=-1

\vspace{-10pt}
\paragraph{In expert-choice routing, auxiliary loss and linear router yield the best performance.}

For the \textit{expert-choice} routing setup (\textit{Left} of \autoref{tab:ablation_study_router}), we evaluate several design aspects: solution to mitigate causality violation (auxiliary router vs. auxiliary loss), normalization functions (\texttt{sigmoid} vs. \texttt{tanh}), router architectures (MLP, Linear, or Wide-MLP), and the impact of an auxiliary z-loss~\citep{zoph2022st}. To assess how well the router performs dynamic allocation, we measure the proportion of ``dead'' tokens—those never selected by the final recursion in a batch—on the validation dataset. Our key findings are as follows: First, using an auxiliary loss is more effective for inference-time behavior than training a separate auxiliary router. Second, a \texttt{sigmoid} normalization function and a simple linear router architecture yield the best performance. Finally, the auxiliary z-loss has a negligible impact on accuracy, though it does slightly reduce the proportion of dead tokens.\looseness=-1

\vspace{-10pt}
\paragraph{In token-choice routing, balancing loss yields stable and accurate routing.}
For \textit{token-choice} routing (\textit{Right} of \autoref{tab:ablation_study_router}), we follow common MoE practices and enable z-loss by default. We compare two balancing strategies: using a balancing loss and training in a loss-free manner using router bias. While both approaches achieve similar log-probability and few-shot accuracy, the explicit balancing loss yields a significantly lower MaxVio~\citep{wang2024auxiliary} in our MoR architectures, making it the preferable choice for stable routing. However, despite this, the model often struggles to balance loads among its heterogeneous experts, even for nearly half of the training steps. Softmax activation with an MLP router performs best, and removing z-loss—though we add back with a very small coefficient in the final design—results in higher performance and routing stability.\looseness=-1

\begin{table*}[t!]
    \vspace{-3pt}
    \caption{
    Ablation results for expert-choice (\textit{Left}) and token-choice (\textit{Right}) routers with various design choices. We use MoR models that apply three recursions to a 360M model with recursion-wise caching. Model performance is measured by NLL and average few-shot accuracy. We evaluate router metrics—dead token ratio (for expert-choice) and MaxVio (for token-choice)—on the validation set. The dead token ratio denotes the proportion of tokens that remain unselected during the last recursion step, measured on 500 samples, each with 2K sequence length. The selected design choice is highlighted in gray.\looseness=-1
    }
    \label{tab:ablation_study_router}
    \small
    \begin{subtable}[t]{0.5\textwidth}
    \label{tab:expert_choice_router}
    \centering
    \resizebox{\linewidth}{!}{
    \setlength{\tabcolsep}{2.5pt}
    \begin{tabular}{lccc|c|c|c}
    \toprule
      \multicolumn{4}{c|}{\textbf{Expert-choice Router}} &  \multicolumn{3}{c}{\textbf{Performance\,($\downarrow$ / $\downarrow$ / $\uparrow$)}} \\
    \cmidrule(l{2pt}r{2pt}){1-4} \cmidrule(l{2pt}r{2pt}){5-7}
    Sampling & Func & Arch & z-loss & Dead & NLL & Few-shot  \\
    \midrule
     Aux\,Rout & $\sigma$ & MLP & \xmark & \,0.0 & 2.8893 & 39.4 \\
     Aux\,Rout & \,\texttt{tanh}\! & MLP & \xmark  & \!66.7 & 2.8720 & 36.2 \\
     \midrule
     Aux\,Loss & $\sigma$ & MLP & \xmark  & \,0.0 & 2.8816 & 40.0 \\
     Aux\,Loss & \,\texttt{tanh}\! & MLP & \xmark  & 0.0 & 2.9933 & 38.8 \\
     \midrule
     \rowcolor[gray]{0.9}
     Aux\,Loss & $\sigma$ & Linear & \xmark  & \,0.1 & \textbf{2.8667} & \textbf{40.1} \\
     Aux\,Loss & $\sigma$ & W-MLP & \xmark  & \,0.4 & 2.8716 & 39.4 \\
     \midrule
     Aux\,Loss & $\sigma$ & Linear & \cmark  & \,0.0 & 2.8824  & 40.0 \\
    \bottomrule
    \end{tabular}
    }
    \end{subtable}
    \hfill
    \begin{subtable}[t]{0.5\textwidth}
    \label{tab:token_choice_router}
    \renewcommand{\arraystretch}{1.04}
    \centering
    \resizebox{\linewidth}{!}{
    \setlength{\tabcolsep}{2.5pt}
    \begin{tabular}{lccc|c|c|c}
    \toprule
      \multicolumn{4}{c|}{\textbf{Token-choice Router}} &  \multicolumn{3}{c}{\textbf{Performance\,($\downarrow$ / $\downarrow$ / $\uparrow$)}} \\
    \cmidrule(l{2pt}r{2pt}){1-4} \cmidrule(l{2pt}r{2pt}){5-7}
    Balancing & Func\!\!\! & Arch & z-loss & M-Vio & NLL & Few-shot  \\
    \midrule
     Loss\,(0.1) & \texttt{soft} & MLP & \cmark & 0.200 & 3.0239 & 38.5  \\
     Loss\,(0.01) & \texttt{soft} & MLP & \cmark & 0.682 & 2.9118 & 39.4 \\
     \midrule
     Loss-free & \texttt{soft} & MLP & \cmark & 0.852 & 2.9081 & 39.4 \\
     Loss-free & $\sigma$ & MLP & \cmark & 1.281  & 3.0188 & 37.6 \\
     \midrule
     Loss\,(0.1) & \texttt{soft} & Linear & \cmark & 0.492 &  2.9974 & 38.4 \\
     Loss\,(0.1) & \texttt{soft} & W-MLP & \cmark & 0.384 & 3.0293 & 38.8 \\
     \midrule
     \rowcolor[gray]{0.9}
     Loss\,(0.1) & \texttt{soft} & Linear & \xmark & 0.266 &  \textbf{2.9358} & \textbf{39.1} \\
    \bottomrule
    \end{tabular}
    }
    \end{subtable}
\end{table*}

\subsection{KV Caching Strategies}
\label{subsec:ablation_kv_cache}

\paragraph{KV sharing robustly works even in parameter-shared architectures.}

In \autoref{fig:pareto_frontier_sharing-c}, we first investigate the effect of KV sharing in Vanilla and Recursive Transformers. As consistent with prior works~\citep{brandon2024reducing, wu2024layer, sun2024you}, if we pretrain models from the scratch, KV sharing does not often compromise performance due to the greater parameter flexibility. Surprisingly, the Recursive Transformer remains relatively robust to KV sharing, despite its reduced degrees of freedom. We found evidence for this by decomposing the KV pairs at each recursion depth into their magnitude and direction. As detailed in \autoref{app:expanded_kv}, depths that share parameters exhibit highly consistent magnitude patterns and high cosine similarity, providing a clear justification for why KV sharing results in only a slight performance drop.

\vspace{-12pt}
\paragraph{KV sharing degrades expert-choice but benefits token-choice routing in MoR.}

We compare recursion-wise KV caching and recursive KV sharing mechanisms in our MoR framework. 
We observe that while recursive KV sharing offers the advantages of reduced memory footprint and overall FLOPs\footnote{Although attention FLOPs increase by $N_{\text{ctx}}/k$ than recursion-wise KV caching, reduced KV projection FLOPs lead to an overall reduction.\looseness=-1}, it leads to quite large performance degradation in expert-choice routing under a fixed token setting. This suggest that exclusively updating and attending to the tokens active in that recursion depth may be more beneficial. Conversely, MoR with token-choice routing could benefit from KV sharing, where its weaker, inaccurate routing decisions can be complemented by the additional contextual information provided by shared KV pairs.

\vspace{-3pt}
\section{Analysis}

\vspace{-1pt}
\subsection{Compute-optimal Scaling Analysis}
\label{subsec:scaling_analysis}

\begin{figure}[t!]
    \centering
    \begin{subfigure}[t]{0.32\textwidth}
        \vspace{0pt}
        \includegraphics[width=\textwidth]{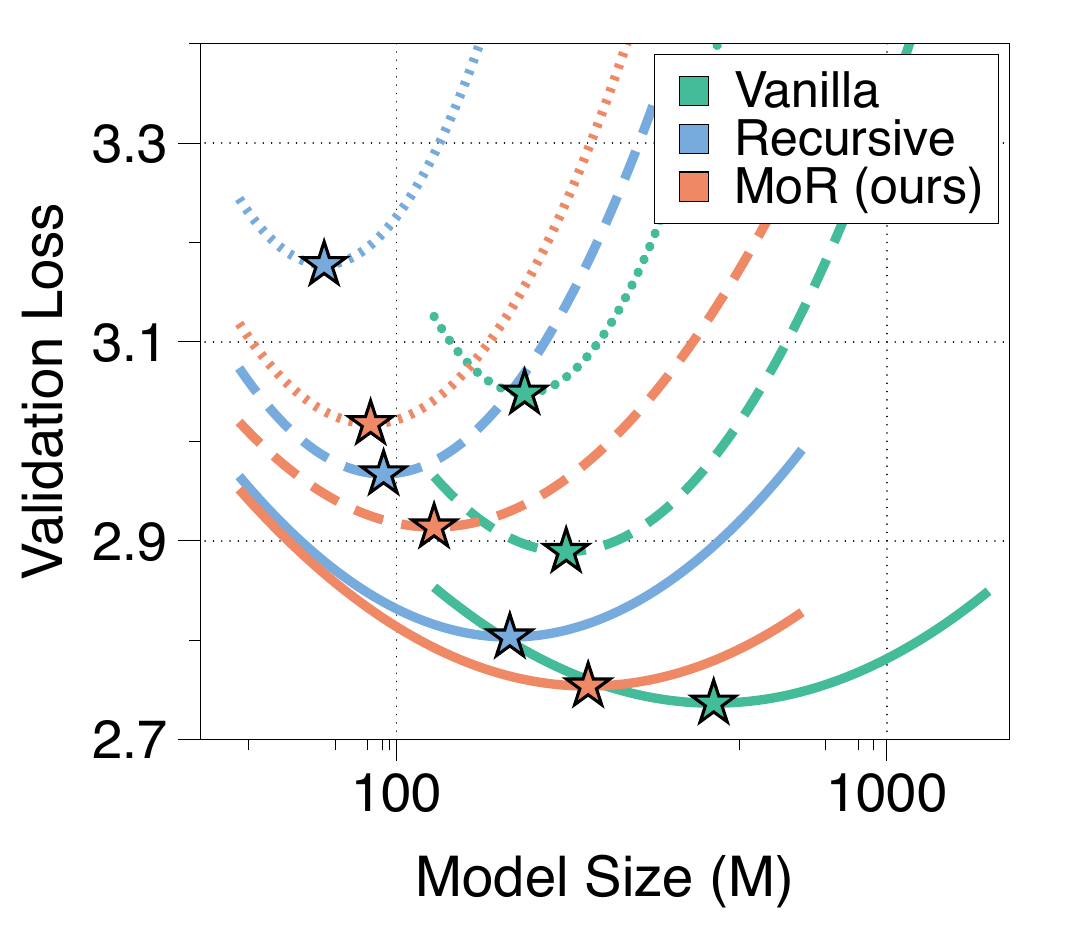}
            \captionsetup{justification=centering}
        \subcaption{Compute-optimal scaling}
        \label{fig:isoflops_qualitative-a}
    \end{subfigure}
    \hfill
    \begin{subfigure}[t]{0.385\textwidth}
        \vspace{-4pt}
        \includegraphics[width=\textwidth]{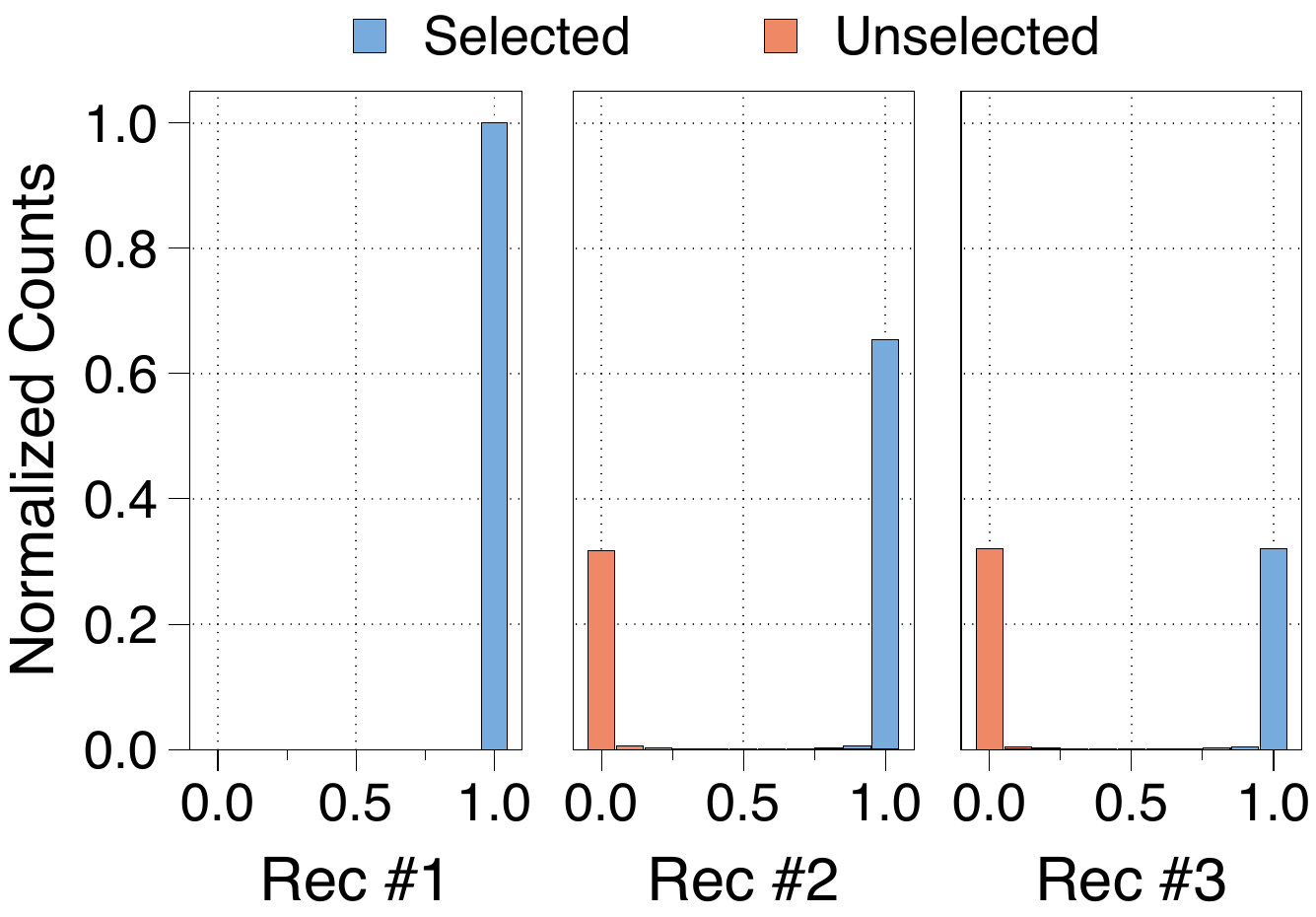}
        \vspace{-4pt}
            \captionsetup{justification=centering}
        \subcaption{Learned router scores}
        \label{fig:isoflops_qualitative-b}
    \end{subfigure}
    \hfill
    \begin{subfigure}[t]{0.275\textwidth}
        \vspace{0pt}
        \includegraphics[width=\textwidth]{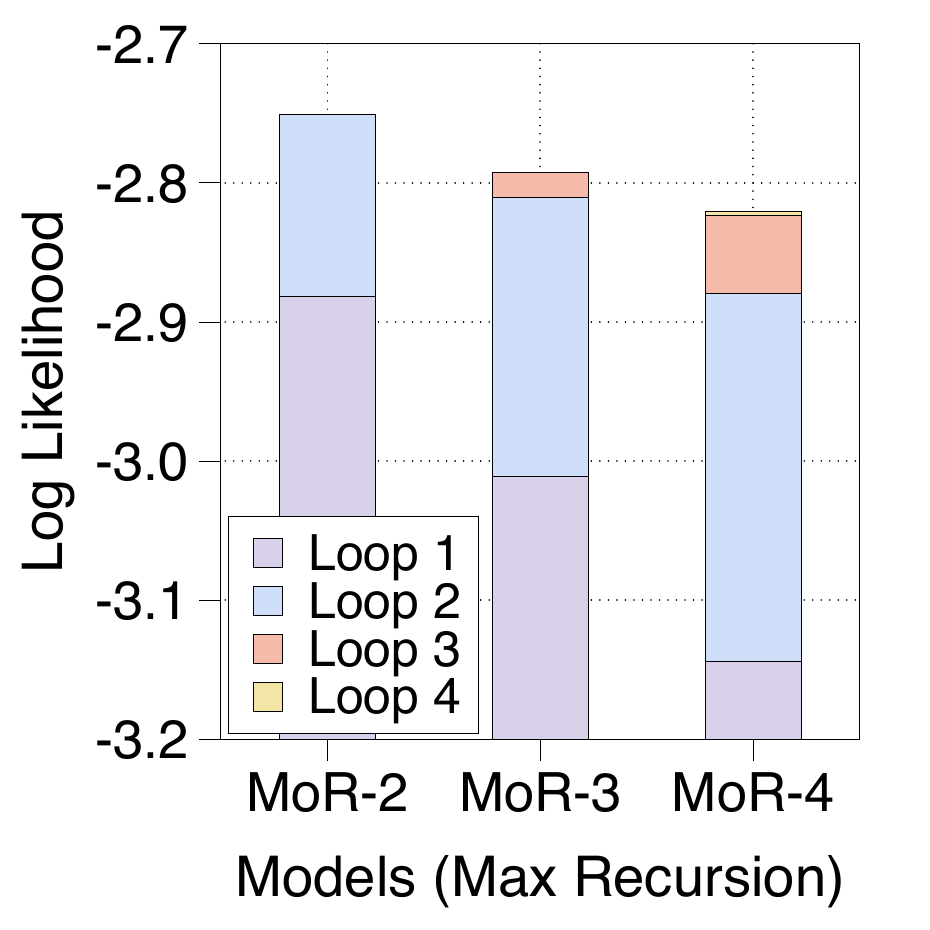}
            \captionsetup{justification=centering}
        \subcaption{Test-time scaling}
        \label{fig:isoflops_qualitative-c}
    \end{subfigure}
    \caption{
    (a) Compute-optimal scaling analysis for three model architectures. Each star indicates the optimal model size for a given compute budget. We visualize the results in \S\ref{subsec:scaling_laws} by fitting polynomial functions for each architecture and FLOPs budget, and derive the optimal points from these fits. (b) Distribution of router outputs for selected and unselected tokens at each recursion step. As an example, a 360M size-based MoR model with $N_r=3$, expert-choice router with auxiliary loss, and recursion-wise caching, is used. (c) Test-time scaling analysis illustrating the cumulative log-likelihood improvement with increasing recursion depth, measured over 500 samples. As we increase $N_r$ based on a 360M model size, the number of unique parameters in MoR decreases, resulting in a gradual decline in overall performance (i.e., a decrease in log-likelihood). All models are trained by an expert-choice router with auxiliary loss and a recursion-wise caching mechanism.\looseness=-1
    }    
    \label{fig:isoflops_qualitative}
\end{figure}

\paragraph{MoR scaling favors model size over training length under isoFLOPs.}

As illustrated in \autoref{fig:isoflops_qualitative-a}, MoR exhibits a distinct compute-optimal scaling behavior compared to baselines under isoFLOPs constraints. The flatter slope of MoR's optimal path (a line connecting stars) indicates that it benefits more significantly from increases in parameter count (i.e., less data-hungry). This is likely because the performance of the shared parameter block itself becomes important, even more than feeding in additional data. Therefore, the optimal scaling policy for MoR models favors allocating resources to increasing model capacity by using larger models trained for shorter steps.\looseness=-1

\vspace{-3pt}
\subsection{Routing Analysis}

\paragraph{The allocation of recursion depth reflects contextual predictability of the subsequent token.}

In the \textit{Right} of \autoref{fig:mor_overview}, we illustrate each token's recursion depth, which directly indicates how easily the next token can be predicted within the given context. For example, the second subword part (e.g., ``-ensively'') of a word is often straightforward to predict, thus needing fewer steps. As also shown in \autoref{tab:qualitative_highlighted_results}, while the initial generation (opening) of function words like ``{-}{-}{-}'', ``('', ``.'', and ``,'' appears easy for models, predicting their closing parts or the first token immediately following their opening might be more challenging.\looseness=-1

\paragraph{Expert-choice router with auxiliary loss perfectly separates selected from unselected tokens.}

\autoref{fig:isoflops_qualitative-b} visualizes one example of the expert-choice router output distribution at each recursion step in a MoR model with $N_r = 3$. For each recursion step, the normalized counts of routing scores are plotted, distinguishing between tokens selected by the expert (blue) and those not selected (orange). In all steps, auxiliary loss achieves a perfect separation in router outputs, with selected tokens sharply concentrated near a routing score of 1.0 and unselected tokens clustering near 0.0. Router output distributions for the two routing strategies and their associated design choices are detailed in \autoref{app:qualitative_results}.\looseness=-1

\subsection{Test-time Scaling Analysis}
\paragraph{MoR enables test-time scaling via deeper recursion.}

We visualize how log-likelihood evolves across recursion steps in MoR models with $N_r = \{2,3,4\}$ in \autoref{fig:isoflops_qualitative-c}. The overlaid bars illustrate the performance of each model when the maximum thinking (recursion) depth of tokens gradually increases. This suggests that deeper recursion not only provides additional compute but also enables each subsequent step to specialize further in refining the token representation or ``thought process'' at its particular depth, leading to better performance. Thereby, these results support the view that MoR enables test-time scaling: allocating more recursion steps at inference can improve generation quality.

\section{Related Work}
\label{sec:related_work}

\paragraph{Recursive Transformers.}
Parameter sharing provides an orthogonal path to efficiency~\citep{dehghani2018universal,lan2019albert0,xia2019tied,jaegle2021perceiver,takase2021lessons,tan2023sparse,ng2024loop,bae2024relaxed}.
The \textit{Universal Transformer} first showed that repeatedly applying a single block can match the representational power of deep, non-shared stacks~\citep{dehghani2018universal}. 
\textit{Looped Transformers} shown to be effective, can act as programmable computers~\citep{giannou2023looped}, learn iterative data-fitting algorithms~\citep{yang2023looped}, generalize to much longer inputs on algorithmic tasks~\citep{fan2024looped}, and illuminate few-shot learning by mimicking multi-step optimizers~\citep{gatmiry2024can}.
Furthermore, \citet{bae2024relaxed} mitigate the accuracy loss often associated with weight tying by adding low-rank adaptation (LoRA) adapters~\citep{hu2022lora} in each loop, yielding \emph{Relaxed Recursive Transformers}. 
Recent work further demonstrates that Recursive Transformers excel at latent reasoning via recurrent depth~\citep{geiping2025scaling}.  
While most prior studies focus on the efficiency gains from weight tying, the \emph{recursive} architecture itself offers a second level: inspired by early-exiting~\citep{schuster2022confident} and compute routing~\citep{raposo2024mixture}, one can vary the number of recursions per input (e.g., per token), allocating compute only where it is most beneficial during both training and inference.\looseness=-1

\paragraph{Adaptive computation.}
Many works have shown that \emph{dynamic} compute allocation can markedly reduce the cost of training and inference, from traditional neural networks~\citep{bengio2015conditional, DBLP:conf/eccv/HuangSLSW16, teerapittayanon2016branchynet, panda2016conditional, graves2016adaptive} to large language models~\citep{hou2020dynabert,Elbayad2020Depth-Adaptive,fedus2022switch,bae-etal-2023-fast, elhoushi2024layerskip, raposo2024mixture}.
Early exiting methods learn to halt processing for ``easy'' samples (e.g., tokens or sequences in language modeling) by skipping the remaining layers~\citep{Elbayad2020Depth-Adaptive, schuster2022confident, dehghani2018universal,mofakhami2024performance}. Alternatively, early exits can be combined with speculative decoding techniques~\citep{chen2023accelerating,leviathan2023fast} during inference by leveraging lower layers for fast drafting~\citep{bae-etal-2023-fast, elhoushi2024layerskip}.
Recently, \textit{Mixture-of-Depths} (MoD)~\citep{raposo2024mixture} reframed adaptive depth as a routing problem: a router at each layer selects a \emph{subset} of tokens to receive the full computation, while the rest bypass the layer, yielding finer-grained conditional compute.  
This new form of adaptive allocation is well suited to Transformer architectures and has already been extended to other modalities~\citep{zhang2024p,luo2024gamma}, highlighting a promising paradigm of dynamic compute at token-level granularity.
MoR applies this routing idea to \emph{recursive} Transformers: tokens are dynamically sent through repeated calls of a single, weight‑tied block instead of through distinct layers. This shift keeps parameter count constant, allows arbitrarily deep (adaptive) compute beyond the model’s physical depth.\looseness=-1

\vspace{-10pt}
\paragraph{Routing mechanism.} 
LLMs have increasingly employed routers to enable adaptive computation, primarily in sparse Mixture-of-Experts (MoE) frameworks~\citep{shazeer2017outrageously, lepikhin2020gshard, dai2022stablemoe, zoph2022st}, i.e., each token is processed by a subset of expert networks chosen by a learned router, dramatically increasing model capacity without a computational overhead. Early MoE architectures~\citep{lepikhin2020gshard, fedus2022switch, jiang2024mixtral} adopted a \textit{token-choice} routing strategy, wherein the router selects the top-k experts for each token based on its hidden state. While effective, this approach often leads to load imbalance across experts, necessitating auxiliary balancing losses. 
To address this, \textit{expert-choice} routing~\citep{zhou2022mixture, guo2025deepseek} has been proposed, wherein each expert selects the tokens to serve, ensuring perfect load balancing and improved efficiency. 
Building on this, a few works employed trainable routers to determine which layers to skip~\citep{zeng2023learning, raposo2024mixture, gadhikar2024attention}. Unlike traditional early-exit methods, these expert-choice routing mechanisms enforce a static compute budget by capping the number of tokens processed per layer (or depth).

\vspace{-10pt}
\paragraph{Key-value caching.}
Key–value (KV) caching stores the per-token key and value tensors produced at each layer during autoregressive decoding; reusing them eliminates quadratic-time recomputation and boosts throughput~\citep{shazeer2019fast, ge2023model, liu2024minicache, xiao2023efficient, pope2022efficiently, kang2024gear0, brandon2024reducing}.
Unfortunately, retaining these tensors quickly saturates GPU memory, especially for long contexts and large batches~\citep{chowdhery2023palm, brandon2024reducing}.
Prior work tackles this issue by quantizing KV activations to lower precision~\citep{hooper2024kvquant0, zhang2024kv}, discarding entries that contribute little to the final output~\citep{zhang2023h2o, liu2023scissorhands}, and sharing keys and values across attention heads~\citep{shazeer2019fast, ainslie2023gqa}.
\citet{brandon2024reducing} push this idea further, allowing adjacent layers to share the same key and value tensors and achieving additional memory savings with negligible quality loss.
Our Mixture-of-Recursions offer a complementary avenue: KV caches generated in early recursions can be reused in later ones, potentially reducing memory consumption even further.
This also provides the advantage of only needing to run the first recursion during prefill phase~\citep{sun2024you} (only with Cycle strategy), promising significant speedups for prompt settings over 1 million tokens. Two caching strategies in MoR can be optimized based on their distinct benefits to suit various deployment settings.\looseness=-1

\vspace{-10pt}
\paragraph{Latent reasoning.}
An emerging line of work enables LLMs to perform reasoning internally within hidden states rather than through explicit verbalization~\citep{goyal2023think, pfau2024let, cheng2024compressed, tack2025llm, konglatent}. Many approaches adopt a \textit{fixed} latent reasoning depth: they insert special tokens or structured prompts (e.g., a learnable ``pause” token~\citep{goyal2023think} or filler punctuation~\citep{pfau2024let}) that allow the model to execute a predetermined number of hidden reasoning passes before producing an answer. Others reuse the model’s hidden states in a closed loop for a fixed number of iterations by feeding final hidden states back as input to simulate chain-of-thought~\citep{hao2024training, shen2025codi, saunshi2025reasoning}. Another line of research enhances latent reasoning by augmenting hidden states with intermediate semantic signals~\citep{zelikman2024quiet, tack2025llm}. However, these methods lack the flexibility to allocate computation where it is most needed, leading to unnecessary overhead on easy inputs and insufficient reasoning on complex ones. 
This motivates leveraging looping mechanisms for more adaptive latent reasoning~\citep{chen2025inner, geiping2025scaling, saunshi2025reasoning, zeng2025pretraining}.

\section{Conclusion}

Mixture-of-Recursions (MoR) presents a unified Transformer architecture that simultaneously leverages parameter sharing, adaptive recursion depth, and efficient KV caching without compromising model quality. 
By dynamically assigning recursion depth to tokens via lightweight routers and selectively caching key-value states for selected tokens, MoR reduces both quadratic attention computation and redundant memory access costs.
Extensive empirical evaluations show that MoR lowers validation perplexity and improves average few-shot accuracy compared to both vanilla and previous recursive baselines, even with higher inference throughput.
These results demonstrate that MoR offers an effective path towards achieving large-model capabilities with significantly reduced computational and memory overhead.

\subsection{Limitations and Future Works}

\paragraph{Reasoning MoR models.}
Recent studies have highlighted the redundancy within reasoning chains and address it by applying token-level adaptive computation, like early-exit mechanisms~\citep{yang2025dynamic, jiang2025flashthink, dai2025s}. Our MoR framework inherently enables latent reasoning by adaptively determining the necessary recursion depth for individual tokens. Therefore, a crucial future work involves exploring how the router can dynamically learn to adjust to the necessity of chain-of-thought (CoT) chains when post-trained on actual reasoning datasets. Developing advanced routing strategies that explicitly align recursion depth with reasoning complexity may enhance reasoning accuracy, computational efficiency, and even interpretability for deliberative reasoning process.

\vspace{-12pt}
\paragraph{Further scaling model family.}

Due to current compute constraints, our experiments have been limited to models of up to 1.7 billion parameters.
The next step is to train Mixture-of-Recursions (MoR) models at a larger scale (over 3 billion parameters) on substantially larger corpora. To enhance the scalability of the MoR architecture, we can first increase the size of the non-shared block. A potential follow-up work could introduce depth-specific LoRA~\citep{bae2024relaxed} or experts (i.e., incorporating Mixture-of-Experts principles) and leverage expert parallelism~\citep{rajbhandari2022deepspeed} to efficiently compute the input at each depth in parallel, which is expected to improve model quality without huge speed bottlenecks. Furthermore, to reduce overall pre-training costs, we could also explore continued pre-training (uptraining), starting from existing pre-trained vanilla LLM checkpoints. As future work, we plan to investigate MoR performance using various initialization strategies for recursive models, as explored in prior work~\citep{bae2024relaxed}.

\vspace{-12pt}
\paragraph{Adaptive capacity control.}
Expert-choice routing offers the significant advantage of guaranteeing perfect load balancing through pre-determined capacity factors~\citep{raposo2024mixture, zhou2022mixture}. However, a limitation arises when we want to allocate different capacities during inference. Specifically, in our MoR models, we observe that when using an auxiliary loss, the router outputs for selected and unselected tokens are almost perfectly separated. This makes it challenging to adjust top-k values after training. Therefore, a more adaptive model design, which can leverage different capacities during both training and inference phases, is needed to address this limitation.

\vspace{-12pt}
\paragraph{Compatibility with sparse algorithms.}

Given MoR's token-level adaptive recursion, we can further optimize computation by integrating structured sparsity. This approach allows for the selective activation of subnetworks or parameters~\citep{liu2023deja}, dynamically pruning unnecessary computations at both the token and layer levels~\citep{raposo2024mixture, elhoushi2024layerskip}. This investigation into sparse model designs promises significant efficiency improvements. We believe many sparsity-based techniques, such as pruning~\citep{han2015learning} or quantization~\citep{jacob2018quantization}, are highly complementary to our MoR framework. This will provide deeper insights into effective sparse architectures within recursive models, offering promising directions for future research.

\vspace{-12pt}
\paragraph{Expansion to multimodal and non‑text domains.} 

MoR's recursion block is inherently modality-agnostic, allowing its adaptive depth mechanism to extend beyond text processing. This crucial property enables MoR to readily integrate into vision, speech, and unified multimodal transformer architectures. Applying token-adaptive recursion to long-context video or audio streams holds the potential for even greater memory efficiencies and substantial throughput gains, crucial for real-world applications. By dynamically adjusting the processing depth for each token or segment, MoR could unlock these significant benefits.

\clearpage
\subsection{Acknowledgements}

We thank Jacob Eisenstein for valuable feedback on an earlier version of the paper. We also acknowledge the support and helpful conversations from Seungyeon Kim, Mostafa Elhoushi, Pascal Vincent, Irina Rish, Sangdoo Yun, and Baeseong Park. Finally, we thank the Google Cloud Platform for awarding Google Cloud credits for this project, and Google's research grant project for its support.
\looseness=-1

This work was supported by Institute of Information \& communications Technology Planning \& Evaluation (IITP) grant funded by the Korea government (MSIT) ([RS-2019-II190075, Artificial Intelligence Graduate School Program (KAIST), 5\%], [No. RS-2024-00457882, AI Research Hub Project, 5\%], and [No. 2022-0-00871, Development of AI Autonomy and Knowledge Enhancement for AI Agent Collaboration, 90\%]). This research was also enabled in part by computational resources, software, and technical assistance provided by \href{https://mila.quebec/en}{Mila},  \href{https://www.ulaval.ca/}{ULaval}, \href{https://www.calculquebec.ca/}{ Calcul Québec}, and \href{https://alliancecan.ca/en}{Digital Research Alliance of Canada}.
\looseness=-1

\bibliographystyle{plainnat}
\bibliography{paper}

\clearpage
\appendix{\setlength{\cftbeforesecskip}{16pt}
\setlength{\cftbeforesubsecskip}{4pt}
\renewcommand\cftsecpagefont{\color{RoyalBlue}}
\renewcommand\cftsubsecpagefont{\color{RoyalBlue}}
\renewcommand\cftsubsubsecpagefont{\color{RoyalBlue}}
\renewcommand{\contentsname}{\large{Contents}}
{
  \hypersetup{linkcolor=}
  \tableofcontents
}

\clearpage

\section{Details of Design Choices for Mixture-of-Recursions}
\label{appx:mor_design_choices}

In this section, we provide detailed descriptions of the design choices employed in Mixture-of-Recursions, expanding upon the summary provided in the main pages.

\subsection{Parameter-sharing Strategy}
\label{app:sharing_strategy}

\autoref{tab:sharing_strategy_middle} shows formulation and visualization of four parameter-sharing strategies: Cycle, Middle-Cycle, Sequence, and Middle-Sequence. These strategies determine how a shared pool of blocks $\Phi'$ are reused across a total of $L$ unrolled layers. The optimal strategy for parameter sharing in recursive models remains an open question.

In the \textbf{Cycle} strategy, a fixed set of parameters is reused cyclically across all recursion steps. By forcing the model to re-engage with the input through the same shared block, it encourages a deeper, iterative refinement process, akin to ``rethinking'' the problem from the ground up at every stage.
However, because the same transformations are applied repeatedly regardless of input variation, it may limit the model’s capacity to learn diverse or highly specialized features.

On the other hand, the \textbf{Sequence} strategy assigns distinct parameters to each recursion block in sequential order. A potential drawback is that simply applying similar transformations twice in a row may lead to redundant features with diminishing returns. Nevertheless, the use of a fixed, sequential order of layers may provide a stable and predictable structure.

Building upon these strategies, the \textbf{Middle} sharing variant further refines parameter reuse by preserving unique parameters at the first and last layers while sharing weights only among the intermediate layers. This approach aims to balance the trade-off between parameter efficiency and representational flexibility, maintaining distinct entry and exit transformations while benefiting from reduced parameter redundancy in the middle layers. In line with recent findings~\citep{kim2023solar, geiping2025scaling}, Middle sharing can capture important input and output nuances more effectively than pure Cycle or Sequence sharing, without significantly increasing model size.

\newcolumntype{M}[1]{>{\centering\arraybackslash}m{#1}}

\begin{table*}[h]
\centering
\caption{
Parameter-sharing strategies in Recursive Transformers. This table shows \textit{Cycle}, \textit{Middle-Cycle}, \textit{Sequence}, and \textit{Middle-Sequence} schemes with layer reuse, where Middle-* retains unique first and last layers.
}
\label{tab:sharing_strategy_middle}
\renewcommand{\arraystretch}{1.}
\resizebox{\textwidth}{!}{%
\setlength{\tabcolsep}{7pt}
\begin{tabular}{
    l
  | M{0.32\textwidth} 
  | M{0.16\textwidth} 
  | M{0.32\textwidth} 
  | M{0.16\textwidth}  
}
\toprule
& \multicolumn{2}{c|}{\textbf{Cycle Strategy}} & \multicolumn{2}{c}{\textbf{Middle-Cycle Strategy}} \\
\cmidrule(l{2pt}r{2pt}){2-3}\cmidrule(l{2pt}r{2pt}){4-5}
\textbf{Layers} & Equation & Figure & Equation & Figure \\
\midrule
Last 
& -- 
& 
& $f\bigl(\mathbf{h}_t^{L-1}; \Phi_{L-1}\bigr)$ 
& \\
[10pt]
Recursion 
& $f\!\Bigl(\mathbf{h}_t^{\ell}; \Phi'_{\ell \bmod (L/N_r)}\Bigr)$ 
& \multirow{-2}{*}{\includegraphics[width=0.91\linewidth]{figs/sharing_strategy/sharing_strategy_cycle.pdf}}
& \!\!\!$f\!\Bigl(\mathbf{h}_t^{\ell}; \Phi'_{(\ell - 1 \bmod ((L-2)/N_r)) + 1}\Bigr)$ 
& \multirow{-3}{*}{\includegraphics[width=0.91\linewidth]{figs/sharing_strategy/sharing_strategy_middle_cycle.pdf}} \\
[18pt]
First 
& -- 
& 
& $f\bigl(\mathbf{h}_t^{0}; \Phi_0\bigr)$ 
& \\ [5pt]
\midrule
& \multicolumn{2}{c|}{\textbf{Sequence Strategy}} & \multicolumn{2}{c}{\textbf{Middle-Sequence Strategy}} \\
\cmidrule(l{2pt}r{2pt}){2-3}\cmidrule(l{2pt}r{2pt}){4-5}
\textbf{Layers} & Equation & Figure & Equation & Figure \\
\midrule
Last 
& -- 
& 
& $f\bigl(\mathbf{h}_t^{L-1}; \Phi_{L-1}\bigr)$ 
& \\
[10pt]
Recursion 
& $f\!\Bigl(\mathbf{h}_t^{\ell}; \Phi'_{ \lfloor{\ell / N_r \rfloor}}\Bigr)$ 
& \multirow{-2}{*}{\includegraphics[width=0.91\linewidth]{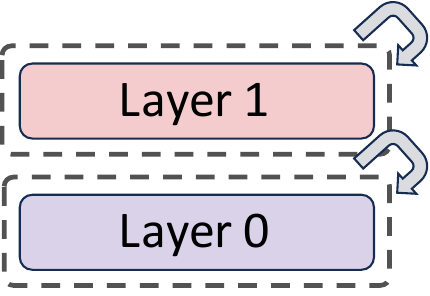}} 
& $f\!\Bigl(\mathbf{h}_t^{\ell}; \Phi'_{\lfloor(\ell - 1) / N_r\rfloor + 1)}\Bigr)$ 
& \multirow{-3}{*}{\includegraphics[width=0.91\linewidth]{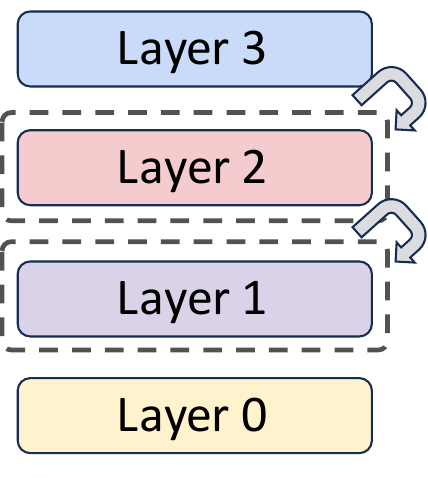}} \\
[20pt]
First 
& -- 
& 
& $f\bigl(\mathbf{h}_t^{0}; \Phi_0\bigr)$ 
& \\[7pt]
\bottomrule
\end{tabular}
}
\end{table*}

\clearpage
\subsection{Routing Strategy}
In this section, we provide an in-depth explanation of the two routing strategies employed in Mixture-of-Recursions: \textit{Expert-choice} and \textit{Token-choice} routers. Each approach has distinct advantages and inherent limitations, which we first outline before discussing the mitigation techniques we utilized.

\vspace{-10pt}
\paragraph{Expert-choice routing.}
The expert-choice router offers several advantages, including a fixed compute budget that simplifies resource management. However, it suffers from a key issue: the top-$k$ selection operation, which requires information of tokens that appear later in the sequence, violates causality in autoregressive inference. This non-causal dependency (i.e., information leakage) can cause unexpected behavior during inference, potentially reducing model reliability. 

To address these challenges, we explore two approaches: the auxiliary router and the auxiliary loss \citep{raposo2024mixture}. The \textbf{auxiliary router} is an additional lightweight network trained jointly but used only during inference; it predicts whether a token will be among the top-$k$ selection. This additional router is trained with a binary cross-entropy loss, where the top-k selections from the main router are defined as the targets. Importantly, its training is isolated from the main objective through gradient blocking, so it does not affect the primary model training. Meanwhile, the \textbf{auxiliary loss} applies the binary cross-entropy loss to the main router itself, enabling it to simultaneously learn to push top-$k$ tokens towards one and others towards zero during training. This ensures the router can reliably predict which tokens will be selected as top-$k$ during inference.

\vspace{-10pt}
\paragraph{Token-choice routing.}
In contrast, the token-choice router assigns recursion depths on a per-token basis without enforcing a fixed compute budget, thus avoiding leakage of information across tokens and preserving autoregressive properties. However, this introduces load imbalance across experts, which results in uneven token distribution across experts (or recursion depths), potentially causing inefficient compute allocation and unbalanced training.

To mitigate load imbalance, we employ two solutions from existing literature. \textbf{Balancing Loss}~\citep{lepikhin2020gshard, fedus2022switch} regularizes for a more uniform distribution of tokens across experts. For a sequence of length $T$, a balancing loss for MoR is calculated as follows: 
\begin{equation*}
\begin{aligned}
\mathcal{L}_{\text{Balance}} &= \alpha \sum_{i=1}^{N_r} f_i P_i, \\
f_i &= \frac{N_r}{T} \sum_{t=1}^T \mathbb{I}(\text{Token } t \text{ selects Expert } i), \\
P_i &= \frac{1}{T} \sum_{t=1}^T g_t^i,
\end{aligned}
\end{equation*}
where $N_r$ is the total number of experts (which is also the number of recursion), $g_t^i$ is the routing score of expert $i$ for token $t$, $f_i$ represents the fraction of tokens routed to expert $i$, $P_i$ denotes the average routing scores of expert $i$, and $\lambda$ is a hyperparameter controlling the strength of the auxiliary loss.\looseness=-1

\textbf{Loss-free}~\citep{wang2024auxiliary} utilizes router biasing without explicit regularization loss. Specifically, this method adjusts per-expert bias terms \(b_i\) to balance token assignments across experts. During each training batch, routing scores are computed, and the number of tokens assigned to each expert ($c_i$) is counted. The load violation error is calculated as $e_i = \bar{c}_i - c_i$ where \(\bar{c}_i\) is the average token count for expert \(i\). Biases are then updated via $b_i \leftarrow b_i + u \times \operatorname{sign}(e_i)$, where \(u\) is a bias update rate.
The biased routing scores for selecting top-$k$ expert are calculated as
\begin{equation*}
g_{i}^t =
\begin{cases}
g_{i}^t, & \text{if } g_{i}^t + b_i \in \texttt{topk}\left(\{g_{j}^t + b_j \mid 1 \leq j \leq N\}, k\right) \\
0, & \text{otherwise}
\end{cases}
\end{equation*}
Note that the expert bias term is only utilized to adjust the routing strategy by influencing the top-$k$ selection.

\clearpage
\subsection{KV Caching Strategy}
This work investigates two principal strategies for key-value (KV) caching to optimize memory usage during Recursive Transformer computations: \textit{recursion-wise caching} and \textit{recursive KV sharing}.

\vspace{-10pt}
\paragraph{Recursion-wise caching.} This keeps separate KV caches for each recursion step, ensuring tokens attend only to the KV pairs generated in their current recursion block. This prevents distribution mismatches between recursion steps, helping to maintain model accuracy while reducing memory and computational costs.

\vspace{-10pt}
\paragraph{Recursive KV sharing.} In contrast, recursive sharing reuses KV pairs computed in the first recursion step for all subsequent steps. Although this approach further lowers memory usage and eliminates the need to compute deeper recursion during the prefill phase, it introduces potential mismatches as later recursion steps receive KV representations originally intended for earlier steps. Such mismatch can negatively impact model performance when token routing is precise.
Therefore, recursion-wise caching is generally preferred in settings with selective token routing to avoid performance degradation, while recursive KV sharing may be considered when memory efficiency is prioritized and prefill time is main bottleneck in the system.

\section{Experimental Setup}
\label{app:experimental_setup}

\paragraph{Training settings.}
We utilized a Llama-based Transformer architecture~\citep{grattafiori2024llama}, referring to the configurations of the open-source SmolLM models~\citep{allal2024SmolLM}. All models were pretrained on a deduplicated subset of the FineWeb-Edu dataset~\citep{penedo2024the} in SmolLM-Corpus~\citep{benallal2024smollmcorpus}, which comprises 220 billion tokens sourced from educational materials. Pretraining was conducted using four H100 or A100 GPUs. 
In our main and isoFLOPs analysis experiments, we utilized a Trapezoid learning rate scheduler, which consists of warmup (about 5\%), stable, and cooldown (20\%) phases. This approach allows us to efficiently continue pretraining for scaling laws from intermediate checkpoints, eliminating the need to train all models independently. In contrast, for all other experiments, we used a simple cosine annealing scheduler.

\vspace{-10pt}
\paragraph{Evaluation settings.}

To assess model performance, we evaluated few-shot accuracy on six benchmarks using the Language Model Evaluation Harness: LAMBADA (LD), HellaSwag (HS), PIQA (PQ), WinoGrande (WG), ARC (Easy and Challenge), and MMLU. For all few-shot datasets, excluding LAMBADA, WinoGrande, and MMLU, we normalized accuracy by the byte length of the target string. We adhered to the standard number of shots for each dataset, and used the continuation task specifically for MMLU for simplicity. 
All evaluation performance measurements were conducted using a single H100 or A100 GPU.

\vspace{-10pt}
\paragraph{Model architecture details.}

\autoref{tab:model_arch} summarizes the architectural specifications of the four Vanilla Transformer models used as the base for our recursive models. Each model variant differs in scale, ranging from 135M to 1.7B total parameters (including both non-embedding and embedding components). For consistency and comparability, all models are trained using a vocabulary size of 49K and a maximum input sequence length of 2K tokens.

\begin{table*}[h!]
    \caption{
    Key parameters of four model size variants. A model's size is defined by the total number of its non-embedding and embedding parameters. Three small models utilize Grouped-Query Attention~\citep{DBLP:conf/emnlp/AinslieLJZLS23}, reducing the number of key-value heads. We refer to the base configurations of the open-sourced SmolLM models~\citep{allal2024SmolLM}.
    }
    \label{tab:model_arch}
    \small
    \centering
    \resizebox{\textwidth}{!}{
    \setlength{\tabcolsep}{10pt}
    \begin{tabular}{l|cccc|cccc|cc}
    \toprule
      &  \multicolumn{4}{c|}{\textbf{Base Configuration}} & \multicolumn{4}{c|}{\textbf{Attention \& Feed-Forward}} & \multicolumn{2}{c}{\textbf{Input}} \\
    \cmidrule(l{2pt}r{2pt}){2-5} \cmidrule(l{2pt}r{2pt}){6-9} \cmidrule(l{2pt}r{2pt}){10-11}
     \textbf{Models} & N-emb & Emb & $N_L$ & $d_{model}$ & $N_{head}$ & $N_{KV}$ & $d_{head}$ & $d_{inter}$ & Vocab & $L_{ctx}$  \\
    \midrule
    Vanilla 135M & 106M & 28M & 30 & 576 & \,\,\,9 & 3 & 64 & 1536 & 49K & 2K \\[1pt]
    Vanilla 360M & 315M & 47M & 32 & 960 & 15 & 5 & 64 & 2560 & 49K & 2K \\[1pt]
    Vanilla 730M & 654M & 75M & 26 & \!\!\!1536 & 24 & 8 & 64 & 4096 & 49K & 2K \\[1pt]
    Vanilla \,\,\,1.7B & 1.61B & \!\!\!101M & 24 & \!\!\!2048 & 32 & \!\!\!32 & 64 & 8192 & 49K & 2K \\[1pt]
    \bottomrule
    \end{tabular}
    }
\end{table*}

\clearpage
\section{Expanded Results of Main Experiments}
\label{app:main_results}

\subsection{Performance Validation at 1.7B Scale}

We validate our models by scaling up the size to 1.7B, all under the same FLOPs budget. As shown in \autoref{tab:main_results_1_7b}, the MoR model with expert-choice routing shows strong performance, but the vanilla model performs slightly better. While this could be due to the differences in the optimal number of tokens associated with each model's optimal scaling policy, it might also signal that the current architecture design of MoR is not suitable for scaling. Nevertheless, we may be able to further improve performance by increasing the number of non-shared blocks (despite reduced efficiency), applying depth-specific LoRA or MoE to each recursion~\citep{bae2024relaxed}, or uptraining with initialization from a well-pretrained checkpoint. Further validation with larger sizes or training on more tokens is required.

\begin{table*}[!h]
    \caption{
    Comparison of MoR, Recursive, and Vanilla Transformers at 1.7B scale under both fixed FLOPs (68.5e18) settings. All models are trained on FineWeb-Edu and evaluated by few-shot accuracy. For the isoFLOP rows, the number of training tokens\,($N_{tok}$) varies by model efficiency.
    For the model sizes, we report non-embedding parameter counts. $\text{}^\dagger$In recursive models, all tokens go through fixed recursion depths ($N_r$), instead of adaptive depths.
    \looseness=-1
    }
    \label{tab:main_results_1_7b}
    \small
    \centering
    \addtolength{\tabcolsep}{0pt}
    \resizebox{\textwidth}{!}{
    \begin{tabular}{l|cc|cc|ccc|cccccc|c}
    \toprule
      &  \multicolumn{2}{c|}{\textbf{MoR}} &  \multicolumn{2}{c|}{\textbf{Recursion}} & \multicolumn{3}{c|}{\textbf{Pretrain}} & \multicolumn{7}{c}{\textbf{Few-shot Accuracy\,$\uparrow$}} \\
    \cmidrule(l{2pt}r{2pt}){2-3} \cmidrule(l{2pt}r{2pt}){4-5} \cmidrule(l{2pt}r{2pt}){6-8} \cmidrule(l{2pt}r{2pt}){9-15} 
     \textbf{Models} & Type & KV & Share & $N_r$ & Param & FLOPs & $N_{tok}$  & LD & HS & PQ & WG & ARC & \!MMLU\! & Avg  \\
    \midrule
    Vanilla & - & - & - & -  & 1.61B &  68.5 & 20B & 40.8 & \textbf{49.4} & \textbf{70.6} & 54.8 & \textbf{47.4} & 30.2 & \textbf{48.9} \\
    \midrule
    \multirow{2}{*}{$\text{Recursive}^{\dagger}$} & - & -  & M-Cyc  & 2 & 0.87B & 68.5 & 18B  & 37.3 & 46.5 & 68.9 & 52.6 & 44.2 & 29.6 & 46.5 \\
     & - & - & M-Cyc & 3 & 0.67B & 68.5 & 20B & 36.4 & 45.3 & 69.5 & 52.7 &  43.9 & 29.1 & 46.2 \\
     \midrule
    & Expert & Cache  & M-Cyc  & 2 & 0.87B & 68.5 & 26B & \textbf{41.1} & 47.5 & 70.0 & \textbf{55.6} & 46.0 & \textbf{30.3} & 48.4 \\
     & Expert & Cache & M-Cyc & 3& 0.67B & 68.5 & 27B  & 37.2 & 46.6& 69.1 & 53.7 & 44.1 & 29.7 & 46.7 \\
     \cmidrule(l{2pt}r{2pt}){2-15}
     \multirow{-3.5}{*}{MoR\,(ours)} & Token & Cache & M-Cyc & 3& 0.67B & 68.5  & 30B & 35.6 & 43.2& 68.1 & 53.0 & 43.4 & 29.0 & 45.4 \\
    \bottomrule
    \end{tabular}
    }
\end{table*}

\subsection{Increasing Recursion Depth under Fixed Parameters}

\autoref{tab:recursion_number_under_same_param} illustrates the performance change when the number of recursions ($N_r$) is increased while the number of unique parameters is fixed at 118M. As the number of recursions rises from 1 (Vanilla) to 2 and 3 (MoR), the negative log-likelihood decreases on both the train and validation sets, and the average few-shot accuracy improves. This demonstrates that the MoR model can effectively scale performance by merely increasing the recursion steps, even with a fixed model size.

\begin{table*}[h!]
    \caption{
    Comparison of MoR with different numbers of recursions under the same number of parameters. Vanilla equals MoR with recursion 1. All models are trained on FineWeb-Edu with 10B tokens, and we apply the Middle-Cycle parameter sharing for MoR models. We report negative log-likelihood (NLL) on the train and validation sets and few-shot accuracy across six benchmarks. 
    }
    \label{tab:recursion_number_under_same_param}
    \small
    \centering
    \resizebox{\linewidth}{!}{
    \setlength{\tabcolsep}{4pt}
    \begin{tabular}{l|ccc|cc|cc|cccccc|c}
    \toprule
      & \multicolumn{3}{c|}{\textbf{Pretrain}} & \multicolumn{2}{c|}{\textbf{Recursion}} & \multicolumn{2}{c|}{\textbf{NLL\,$\downarrow$}} & \multicolumn{7}{c}{\textbf{Few-shot Accuracy\,$\uparrow$}} \\
    \cmidrule(l{2pt}r{2pt}){2-4} \cmidrule(l{2pt}r{2pt}){5-6} \cmidrule(l{2pt}r{2pt}){7-8} \cmidrule(l{2pt}r{2pt}){9-15}
    \textbf{Models} & N-Emb & $N_L$ & $N_{tok}$ & $N_r$ & Share & Train & Valid & LD & HS & PQ & WG & ARC & \!\!MMLU & Avg  \\ 
    \midrule
Vanilla & 118M & 12 & 10B &  1 & - & 2.9182 &  2.9678 & 25.81 & 33.08 & 61.81 & 51.46 & 38.10 & 26.64 & 39.48  \\
MoR & 118M  & 1+10+1 & 10B & 2 & M-Cyc &  2.8767 &  2.9205 & 27.09 & 33.84 & 62.13 & 53.04 & 36.77 & 26.88 & 39.96  \\
MoR & 118M  & 1+10+1 & 10B & 3 & M-Cyc & 2.8667 &  2.9111 & 27.38 & 34.60 & 63.17 & 51.46 & 37.15 & 26.83 & 40.10\\
    \bottomrule
    \end{tabular}
    }
\end{table*}

\clearpage
\section{Expanded Results of IsoFLOP Analysis}
\label{app:scaling_laws}

In the main paper (\S\ref{subsec:scaling_laws}), we compared Vanilla, Recursive and our Mixture‑of‑Recursions (MoR) models under matched \emph{training compute}. 
Four base model capacities were studied—135M, 360M, 730M and 1.7B parameters. 
For recursive and MoR models, we fix the recursion count to $N_r\!=\!3$, so the number of \emph{unique} parameters is roughly one‑third of the vanilla counterpart.  
Each architecture is trained once for the \emph{largest} compute budget (16.5EB)\footnote{1EB $\,{=}\,10^{18}$ floating‑point operations.} and the resulting checkpoint is re‑used to obtain the 5EB and 2EB variants, as detailed below.

\vspace{-10pt}
\paragraph{FLOPs approximated calculation of Transformers.}

We follow the approximation for calculating FLOPs as detailed in \citet{kaplan2020scaling}. Our analysis solely focuses on forward pass FLOPs, since the FLOPs involved in the backward pass are typically just double those of the forward pass. For most operations within Transformers, which primarily consist of linear projections, the forward pass FLOPs are calculated as two times the number of parameters, excluding the attention mechanism.

Regarding attention, we specifically account for the operations from the dot product between queries and keys and the scaling of values with softmax values. We only calculate FLOPs that contribute to the actual loss, excluding redundant computations in the upper triangular portion due to causality masking. Furthermore, we omit any additional computational costs associated with FlashAttention~\citep{dao2022flashattention}, normalization, and non-linearity operations from our overall FLOPs calculation.

As a result, Vanilla and Recursive Transformers have the same FLOPs. For MoR, the FLOPs calculation varies based on the routing and KV caching strategy. Especially, we calculated FLOPs based on the sequence length at each recursion depth, which is determined by the capacity factor and caching mechanism. In the case of token-choice routing, since the actual token allocation changes at every step, we approximated the FLOPs by assuming perfect balancing.
Furthermore, we add extra layers to a few MoR models to ensure their effective depth is divisible by the recursion number. For example, in a 135M model with 30 layers, setting a base depth of 10 and applying recursion three times (as in the Middle-Cycle strategy) results in a total of 32 layers. These additional layers introduce extra FLOPs, so we reduce the number of training steps accordingly to maintain our predefined FLOP budget.

\vspace{-10pt}
\paragraph{Trapezoid learning‑rate schedule with checkpoint reuse.}
To avoid retraining every model from scratch for each FLOPs budget, we employ the \textit{trapezoid} schedule~\citep{xing2018walk}. The rule of this scheduler is as follows:\looseness=-1
\begin{equation*}
\eta(t)=
\begin{cases}
\frac{t}{w}\,\eta_{\max}, & 0 \le t < w             \quad\text{(warm‑up)},\\[4pt]
\eta_{\max},                            & w \le t < p             \quad\text{(plateau)},\\[2pt]
\eta_{\max}\!\Bigl(1-\tfrac{t-p}{d}\Bigr), & p\le t < p+d \quad\text{(cool‑down)},
\end{cases}
\end{equation*}

where $w$ denotes the warm‑up interval, $p-w$ is the constant‑LR plateau, and $d$ represents the cool-down segment. This stable phase allows us to efficiently manage experiments by saving intermediate checkpoints and then running additional cool-down steps from those points, according to each budget. For the warmup, we allocate \(5\,\%\) of the total training steps of the smallest budget (2EB), and we set the cool‑down steps to \(20\,\%\) of the total training steps for each \emph{corresponding} budget.

\clearpage
\paragraph{Results at a glance.}
Table~\ref{tab:scaling_laws} reports NLL on FineWeb-Edu validation set and few‑shot accuracy on six benchmarks. Our findings reveal a clear trend: more compute consistently leads to better models, evidenced by lower NLL and improved accuracy with higher FLOPs. However, weight-sharing alone in recursive models degraded performance compared to Vanilla, a clear trade-off for the reduced parameters. Crucially, token-routed MoR models overcome this shortcoming, catching up to and then surpassing Vanilla models from 360M parameters upward, all while utilizing only one-third of the parameters. This performance advantage persists at 730M and 1.7B parameter scales.\looseness=-1

\begin{table*}[h!]
    \caption{Detailed results of isoFLOP analysis across three compute budgets. We evaluate negative log‑likelihood (NLL) on the FineWeb‑Edu validation set and few‑shot accuracy on six downstream tasks for four base model sizes (135M, 360M, 730M, 1.7B). Each model was initially trained up to 16.5EB and sliced back to 5EB and 2EB via mid‑training checkpoints using a trapezoid learning‑rate schedule. All models used three recursion steps. For MoR models, we use expert-choice routing and recursion-wise caching mechanisms. We highlight the best-performing model in each setting in gray.
    }

    \label{tab:scaling_laws}
    \small
    \centering
    \resizebox{\textwidth}{!}{
    \setlength{\tabcolsep}{4pt}
    \begin{tabular}{lc|cccc|cc|c|cccccc|c}
    \toprule
      & & \multicolumn{4}{c|}{\textbf{Pretrain}} &  \multicolumn{2}{c|}{\textbf{Recursion}} & \textbf{NLL\,$\downarrow$} & \multicolumn{7}{c}{\textbf{Few-shot Accuracy\,$\uparrow$}} \\
    \cmidrule(l{2pt}r{2pt}){3-6} \cmidrule(l{2pt}r{2pt}){7-8} \cmidrule(l{2pt}r{2pt}){9-9} \cmidrule(l{2pt}r{2pt}){10-16}
     \textbf{Models} & \textbf{Base} & N-Emb  & $N_L$ & FLOPs & $N_{tok}$ & Share & Loop & FineWeb & LD & HS & PQ & WG & ARC & \!MMLU\! & Avg  \\
    \midrule
    \rowcolor[gray]{0.9}
    Vanilla & 135M & 106M & 30 & \,\,\,2.0e+18  & 6.5B  & -  & - & 3.0922 & 22.80 & 30.93 & 62.35 & 51.14 & 36.28 & 26.29 & 38.30  \\
    Recursive & 135M & \,\,\,42M & 1+10+1 & \,\,\,2.0e+18  & 6.1B & M-Cyc  & 3 & 3.2058 & 19.79 & 29.32 & 60.17 & 50.59 & 34.83 & 25.40 & 36.68  \\
    MoR & 135M & \,\,\,42M & 1+10+1 & \,\,\,2.0e+18  & 9.2B & M-Cyc  & 3 & 3.1077 & 21.13 & 31.00 & 59.79 & 49.09 & 34.87 & 25.63 & 36.92  \\
    \midrule
    \rowcolor[gray]{0.9}
    Vanilla & 135M & 106M & 30 &  \,\,\,5.0e+18   & \!\!\!16.1B & -  & - & 2.9464 & 26.88 & 33.69 & 63.98 & 51.46 & 37.08 & 27.07 & 40.03  \\
    Recursive & 135M & \,\,\,42M & 1+10+1  &  \,\,\,5.0e+18   & \!\!\!15.1B & M-Cyc  & 3 & 3.0534 & 24.51 & 31.57 & 62.40 & 50.83 & 35.78 & 25.94 & 38.51  \\
    MoR & 135M & \,\,\,42M & 1+10+1  &  \,\,\,5.0e+18   & \!\!\!23.1B & M-Cyc  & 3 & 3.0192 & 22.01 & 32.53 & 61.75 & 49.88 & 35.39 & 26.19 & 37.96  \\
    \midrule
    \rowcolor[gray]{0.9}
    Vanilla & 135M & 106M & 30 & 16.5e+18  & \!\!\!53.3B & -  & - & 2.8432 & 30.16 & 36.51 & 64.80 & 53.43 & 40.17 & 27.82 & 42.15  \\
    Recursive & 135M & \,\,\,42M & 1+10+1 & 16.5e+18  & \!\!\!50.0B & M-Cyc  & 3 & 2.9552 & 25.98 & 33.36 & 63.98 & 51.78 & 36.96 & 26.68 & 39.79  \\
    MoR & 135M & \,\,\,42M & 1+10+1 & 16.5e+18  & \!\!\!76.2B & M-Cyc  & 3 & 2.9490 & 22.61 & 33.99 & 61.92 & 47.83 & 35.95 & 26.36 & 38.11  \\
    \midrule
    \midrule
    Vanilla & 360M & 315M & 32 & \,\,\,2.0e+18   & 2.4B & -  & - & 3.3785 & 17.27 & 27.90 & 59.36 & 51.38 & 32.10 & 25.49 & 35.58  \\
    Recursive & 360M & 118M & 1+10+1 & \,\,\,2.0e+18   & 2.4B & M-Cyc  & 3 & 3.4864 & 10.34 & 26.66 & 58.00 & 51.54 & 30.94 & 24.94 & 33.74  \\
    \rowcolor[gray]{0.9}
    MoR & 360M & 118M & 1+10+1 & \,\,\,2.0e+18   & 3.6B & M-Cyc  & 3 & 3.1026 & 24.14 & 30.53 & 61.86 & 50.99 & 34.74 & 25.50 & 37.96  \\
    \midrule
    Vanilla & 360M & 315M & 32 &  \,\,\,5.0e+18   & 6.0B & -  & - & 3.0097 & 25.17 & 32.10 & 63.22 & 48.62 & 36.01 & 26.69 & 38.63  \\
    Recursive & 360M & 118M & 1+10+1 &  \,\,\,5.0e+18   & 6.0B & M-Cyc  & 3 & 3.0722 & 23.29 & 31.19 & 62.62 & 51.30 & 35.85 & 25.99 & 38.37  \\
    \rowcolor[gray]{0.9}
    MoR & 360M & 118M & 1+10+1 &  \,\,\,5.0e+18   & 9.0B & M-Cyc  & 3 & 2.9161 & 28.33 & 34.53 & 63.22 & 51.07 & 36.70 & 26.98 & 40.14  \\
    \midrule
    Vanilla & 360M & 315M & 32 & 16.5e+18  & \!\!\!19.8B & -  & - & 2.7824 & 31.94 & 37.92 & 66.10 & 51.30 & 39.70 & 27.95 & 42.49  \\
    Recursive & 360M & 118M & 1+10+1 & 16.5e+18  & \!\!\!19.8B & M-Cyc  & 3 & 2.8466 & 29.75 & 35.92 & 64.91 & 51.46 & 39.12 & 27.18 & 41.39  \\
    \rowcolor[gray]{0.9}
    MoR & 360M & 118M & 1+10+1 & 16.5e+18  & \!\!\!29.7B & M-Cyc  & 3 & 2.7924 & 33.15 & 37.94 & 66.97 & 52.09 & 38.46 & 27.49 & 42.68  \\
    \midrule
    \midrule
    Vanilla & 730M & 654M & 26 & \,\,\,2.0e+18   & 1.2B & -  & - & 3.7164 & 07.74 & 26.58 & 57.62 & 51.14 & 29.74 & 24.46 & 32.88  \\
    Recursive & 730M & 252M & 1+8+1 & \,\,\,2.0e+18   & 1.2B & M-Cyc  & 3 & 3.8136 & 05.53 & 26.25 & 55.77 & 50.59 & 29.88 & 24.63 & 32.11  \\
    \rowcolor[gray]{0.9}
    MoR & 730M & 252M & 1+8+1 & \,\,\,2.0e+18   & 1.8B & M-Cyc  & 3 & 3.3300 & 17.93 & 28.74 & 59.30 & 51.46 & 33.14 & 25.37 & 35.99  \\
    \midrule
    Vanilla & 730M & 654M & 26 &  \,\,\,5.0e+18   & 3.1B & -  & - & 3.0821 & 22.05 & 31.99 & 62.68 & 50.67 & 35.88 & 26.12 & 38.23  \\
    Recursive & 730M & 252M & 1+8+1 &  \,\,\,5.0e+18   & 3.1B & M-Cyc  & 3 & 3.1640 & 18.51 & 30.72 & 62.13 & 47.83 & 35.97 & 25.84 & 36.83  \\
    \rowcolor[gray]{0.9}
    MoR & 730M & 252M & 1+8+1 &  \,\,\,5.0e+18   & 4.5B & M-Cyc  & 3 & 3.0067 & 26.18 & 32.76 & 62.46 & 50.91 & 36.93 & 26.37 & 39.27  \\
    \midrule
    Vanilla & 730M & 654M & 26 & 16.5e+18  & \!\!\!10.1B & -  & - & 2.7048 & 34.50 & 40.29 & 66.81 & 49.49 & 40.82 & 28.66 & 43.43  \\
    Recursive & 730M & 252M & 1+8+1 & 16.5e+18  & \!\!\!10.1B & M-Cyc  & 3 & 2.7886 & 30.76 & 37.84 & 65.51 & 52.41 & 39.26 & 27.51 & 42.21  \\
    \rowcolor[gray]{0.9}
    MoR & 730M & 252M & 1+8+1 & 16.5e+18  & \!\!\!14.9B & M-Cyc  & 3 & 2.7438 & 32.93 & 39.55 & 66.32 & 54.38 & 40.00 & 28.09 & 43.55  \\
    \midrule
    \midrule
    Vanilla & \,\,\,1.7B & 1.61B & 24 & \,\,\,2.0e+18   & 0.6B & -  & - & 5.1349 & 00.00 & 24.96 & 51.03 & 51.38 & 25.75 & 23.07 & 29.37  \\
    Recursive & \,\,\,1.7B & 0.67B & 1+8+1 & \,\,\,2.0e+18   & 0.5B & M-Cyc  & 3 & 5.3277 & 00.00 & 25.27 & 51.36 & 48.62 & 26.52 & 22.98 & 29.13  \\
    \rowcolor[gray]{0.9}
    MoR & \,\,\,1.7B & 0.67B & 1+8+1 & \,\,\,2.0e+18   & 0.8B & M-Cyc  & 3 & 4.1175 & 01.44 & 25.80 & 53.97 & 49.64 & 27.56 & 24.08 & 30.42  \\
    \midrule
    Vanilla & \,\,\,1.7B & 1.61B & 24 &  \,\,\,5.0e+18   & 1.5B & -  & - & 3.6926 & 08.33 & 26.84 & 57.29 & 51.30 & 29.72 & 24.51 & 33.00  \\
    Recursive & \,\,\,1.7B & 0.67B & 1+8+1 &  \,\,\,5.0e+18   & 1.3B & M-Cyc  & 3 & 3.8876 & 03.14 & 26.57 & 54.73 & 49.17 & 29.01 & 24.49 & 31.19  \\
    \rowcolor[gray]{0.9}
    MoR & \,\,\,1.7B & 0.67B & 1+8+1 &  \,\,\,5.0e+18   & 2.0B & M-Cyc  & 3 & 3.2905 & 17.62 & 28.32 & 59.03 & 49.80 & 32.14 & 25.28 & 35.37  \\
    \midrule
    Vanilla & \,\,\,1.7B & 1.61B & 24 & 16.5e+18  & 4.8B & -  & - & 2.8658 & 26.94 & 35.61 & 64.74 & 50.59 & 38.55 & 26.81 & 40.54  \\
    Recursive & \,\,\,1.7B & 0.67B & 1+8+1 & 16.5e+18  & 4.5B & M-Cyc  & 3 & 3.0042 & 23.25 & 32.09 & 62.95 & 50.75 & 37.64 & 26.53 & 38.87  \\
    \rowcolor[gray]{0.9}
    MoR & \,\,\,1.7B & 0.67B & 1+8+1 & 16.5e+18  & 6.5B & M-Cyc  & 3 & 2.8316 & 28.10 & 36.18 & 64.64 & 50.99 & 38.68 & 27.25 & 40.97  \\
    \bottomrule
    \end{tabular}
    }
\end{table*}

\clearpage
\section{Details of Experimental Settings for Throughput Measurement}
\label{app:throughput}

We implement a continuous depth-wise batching inference system~\citep{bae2024relaxed, hooper2023speed} to evaluate decoding throughput of MoR models. Queries are enqueued and scheduled dynamically during decoding using 1K samples from the FineWeb-Edu validation set. In particular, for MoR, when some queries exit early, the vacant slots in the batch are immediately filled with new queries waiting in the queue, maintaining a fully utilized batch at all times.\looseness=-1

We compare the throughput of Vanilla and MoR models (at a 360M parameter scale) for generating a certain length of tokens per sample, where the number is sampled from a normal distribution with a mean of 256, starting without any input prefix. The speeds of the MoR models are normalized against the speed of the Vanilla Transformer. For pretraining MoR-4 models, we add two additional layers (34 layers in total) before applying recursion. This ensures the total effective depth is divisible by the recursion number (specifically for the Middle-Cycle strategy). Consequently, the speed comparison for MoR-4 is made against a modified vanilla model that includes these two extra layers, resulting in a total of 34 layers (32 original layers + 2 added layers).\looseness=-1

We use two batching settings: (1) a \textit{fixed batch size of 32} and (2) a \textit{relative maximum batch size}, derived by multiplying 32 by the ratio of the maximum batch sizes of vanilla and MoR models. Specifically, based on the H100 GPU's VRAM size, we calculated the maximum batch sizes by considering model parameters and their KV cache memory. For simplicity, we omit the memory size from the hidden states at the current position. Under these adaptive conditions, MoR-2 supports a batch size of 42, MoR-3 supports 48, and MoR-4 supports up to 51. By employing recursion-wise KV caching, MoR allows a substantial increase in batch size stemming from its reduced parameter and KV cache memory footprint.

For implementation, we use a queue to enable continuous depth-wise batching and employ FlashAttention 2~\citep{dao2023flashattention} to support variable-length KV caches within a batch. We adopt a \textit{static}-sized cache where each position is updated over time, since this is compatible with \texttt{torch.compile}~\citep{paszke2019pytorch} to further optimize inference speeds. Furthermore, mimicking real-world deployment scenarios~\citep{kwon2023efficient, zhong2024distserve}, we decouple the transformer block phase from the rest of the computation by pre-processing the input embeddings or the first non-shared layer before passing into the transformer blocks. Then, we measured the actual time taken during the forward pass. Note that we include a warmup stage by running the model for 100 iterations before actual measurement, in order to obtain stable timing results. For further optimization, tokens that exited early were accumulated up to the maximum batch size before being processed by the last non-shared layer, classifier, and embedding layers (including the non-shared first layer in the case of MoR models). After this, we queue them for sequential batching by following a FIFO (First-In, First-Out) strategy. For implementation convenience, we exclude the time spent on caching and updating for KV pairs, as these aspects can be significantly optimized through various engineering techniques~\citep{kwon2023efficient}. We leave a more precise speed comparison, which accounts for these considerations, as future work.

\clearpage
\section{Expanded Results of Parameter Sharing Strategy}
\label{app:parameter_sharing}

This section complements the ablation in \S\ref{subsec:sharing} by providing the full quantitative panorama behind \autoref{fig:pareto_frontier_sharing-b}. We revisit the four weight‑tying schemes—\textit{Cycle}, \textit{Sequence}, \textit{Middle‑Cycle}, and \textit{Middle‑Sequence}—on two base model scales (135M and 360M non‑embedding parameters) and two different recursion depths ($N_{r}=2$ and~3). All models were trained from scratch for 10B tokens under identical optimization hyperparameters. Validation NLL on FineWeb‑Edu and averaged few‑shot accuracy over six benchmarks are summarized in \autoref{tab:revist_sharing_strategy}.

\vspace{-10pt}
\paragraph{Middle‑Cycle is consistently the safest choice.}
For the 360M models, Middle-Cycle achieves the lowest NLL at both depths ($N_{r}=2$ and $N_{r}=3$) and also shows the largest improvement in average accuracy compared to vanilla reduced models. For the 135M models, while Cycle is slightly ahead at two recursion setting (3.0071 vs. 3.0330), Middle-Cycle overtakes when recursion depth rises (3.1048 vs. 3.1154) and shows a steadier accuracy profile. Meanwhile, pure Sequence sharing records the worst NLL in all four settings, and its accuracy gap widens with recursion depth. The Middle strategy slightly improves the performance of the Sequence, but it still performs worse than the Cycle-based methodology. We visualized the results in \autoref{fig:revisit_sharing_app}.\looseness=-1

\begin{table*}[h!]
  \caption{
    Comparison of parameter-sharing strategies (Cycle, Sequence, Middle-Cycle, Middle-Sequence) across two model scales (135M and 360M) and two recursion depths ($N_R = 2$ and $N_R = 3$). All models are pretrained from scratch on 10B tokens. We report validation negative log-likelihood (NLL) on FineWeb-Edu and few-shot accuracy across six tasks. Middle-Cycle consistently outperforms other strategies in both NLL and average task accuracy, especially at higher recursion depth. We highlight the optimal strategy for each setting in gray.\looseness=-1
    }
  \label{tab:revist_sharing_strategy}
  \small
  \centering
  \resizebox{\textwidth}{!}{
  \setlength{\tabcolsep}{4pt}
  \begin{tabular}{l|ccc|cc|c|cccccc|c}
  \toprule
   & \multicolumn{3}{c|}{\textbf{Pretrain}} & \multicolumn{2}{c|}{\textbf{Recursion}} & \textbf{NLL\,$\downarrow$} & \multicolumn{7}{c}{\textbf{Few-shot Accuracy\,$\uparrow$}} \\
  \cmidrule(l{2pt}r{2pt}){2-4} \cmidrule(l{2pt}r{2pt}){5-6} \cmidrule(l{2pt}r{2pt}){7-7} \cmidrule(l{2pt}r{2pt}){8-14} 
    \textbf{Base Model} & N-Emb & $N_L$ & $N_{tok}$ & Share & Loop & FineWeb & LD & HS & PQ & WG & ARC & \!MMLU\! & Avg \\
  \midrule
  Vanilla\,135M & 106M & 30 & 10B & - & - & 3.0323 & 24.14 & 31.12 & 61.15 & 52.01 & 34.74 & 25.95 & 38.19 \\
  Vanilla\,135M & \,\,\,53M & 15 & 10B & - & - & 3.0818 & 23.64 & 30.10 & 60.94 & 50.99 & 35.38 & 25.93 & 37.83 \\
  Vanilla\,135M & \,\,\,35M & 10 & 10B & - & - & 3.1582 & 21.46 & 29.30 & 60.01 & 52.01 & 34.40 & 25.53 & 37.12 \\
  \midrule
  \rowcolor[gray]{0.9}
  Vanilla\,135M & \,\,\,53M & 15 & 10B & Cyc & 2 & 3.0071 & 25.52 & 31.25 & 61.10 & 50.99 & 36.08 & 26.11 & 38.51 \\
  Vanilla\,135M & \,\,\,53M & 15 & 10B & Seq & 2 & 3.1093 & 22.39 & 29.60 & 61.10 & 50.12 & 34.46 & 25.72 & 37.23 \\
   Vanilla\,135M & \,\,\,57M & 1+14+1 & 10B & M-Cyc & 2 & 3.0330 & 23.40 & 31.20 & 61.59 & 50.59 & 35.44 & 25.54 & 37.96 \\
    Vanilla\,135M & \,\,\,57M & 1+14+1 & 10B & M-Seq & 2 & 3.0991 & 21.70 & 30.06 & 60.45 & 49.41 & 35.20 & 25.74 & 37.09 \\
    \midrule
    Vanilla\,135M & \,\,\,35M & 10 & 10B & Cyc & 3 & 3.1154 & 21.42 & 30.14 & 60.61 & 49.72 & 34.15 & 25.57 & 36.94 \\
    Vanilla\,135M & \,\,\,35M & 10 & 10B & Seq & 3 & 3.1637 & 19.99 & 29.39 & 59.25 & 51.62 & 33.79 & 25.32 & 36.56 \\
    \rowcolor[gray]{0.9}
    Vanilla\,135M & \,\,\,39M & 1+9+1 & 10B & M-Cyc & 3 & 3.1048 & 22.41 & 30.35 & 61.04 & 49.01 & 34.80 & 25.91 & 37.26 \\
    Vanilla\,135M & \,\,\,39M & 1+9+1 & 10B & M-Seq & 3 & 3.1602 & 20.69 & 29.35 & 61.43 & 51.30 & 34.40 & 25.51 & 37.11 \\
    \midrule
    \midrule
    Vanilla\,360M & 315M & 32 & 10B & - & - & 2.8471 & 27.27 & 34.78 & 64.20 & 52.80 & 38.29 & 26.72 & 40.68 \\
    Vanilla\,360M & 157M & 16 & 10B & - & - & 2.8908 & 27.01 & 33.49 & 64.42 & 52.09 & 37.40 & 26.54 & 40.16 \\
    Vanilla\,360M & \,\,\,98M & 10 & 10B & - & - & 2.9449 & 26.41 & 32.93 & 63.38 & 50.36 & 37.15 & 26.48 & 39.45 \\
    \midrule
    Vanilla\,360M & 157M & 16 & 10B & Cyc & 2 & 2.8487 & 28.47 & 34.79 & 63.06 & 49.96 & 37.38 & 26.81 & 40.08 \\
    Vanilla\,360M & 157M & 16 & 10B & Seq & 2 & 2.9467 & 26.33 & 32.49 & 62.89 & 52.41 & 36.37 & 26.24 & 39.46 \\
    \rowcolor[gray]{0.9}
    Vanilla\,360M & 167M & 1+15+1 & 10B & M-Cyc & 2 & 2.8295 & 28.59 & 34.98 & 64.53 & 50.51 & 39.68 & 27.20 & 40.91 \\
    Vanilla\,360M & 167M & 1+15+1 & 10B & M-Seq & 2 & 2.9303 & 26.14 & 32.71 & 62.79 & 51.38 & 36.31 & 25.73 & 39.18 \\
    \midrule
    Vanilla\,360M & \,\,\,98M & 10 & 10B & Cyc & 3 & 2.9363 & 25.87 & 32.98 & 62.89 & 50.28 & 36.35 & 26.54 & 39.15 \\
    Vanilla\,360M & \,\,\,98M & 10 & 10B & Seq & 3 & 3.0245 & 24.55 & 31.48 & 63.11 & 49.25 & 35.65 & 25.73 & 38.30 \\
    \rowcolor[gray]{0.9}
    Vanilla\,360M & 118M & 1+10+1 & 10B & M-Cyc & 3 & 2.8760 & 28.51 & 34.89 & 64.31 & 50.51 & 39.51 & 27.20 & 40.82 \\
    Vanilla\,360M & 118M & 1+10+1 & 10B & M-Seq & 3 & 2.9753 & 24.18 & 31.89 & 62.08 & 49.72 & 36.47 & 26.27 & 38.44 \\
    \bottomrule
  \end{tabular}
  }
  
\end{table*}

\clearpage

\begin{figure}[h]
    \centering
    \begin{subfigure}[t]{0.35\textwidth}
    \captionsetup{justification=centering}
        \includegraphics[width=\textwidth]{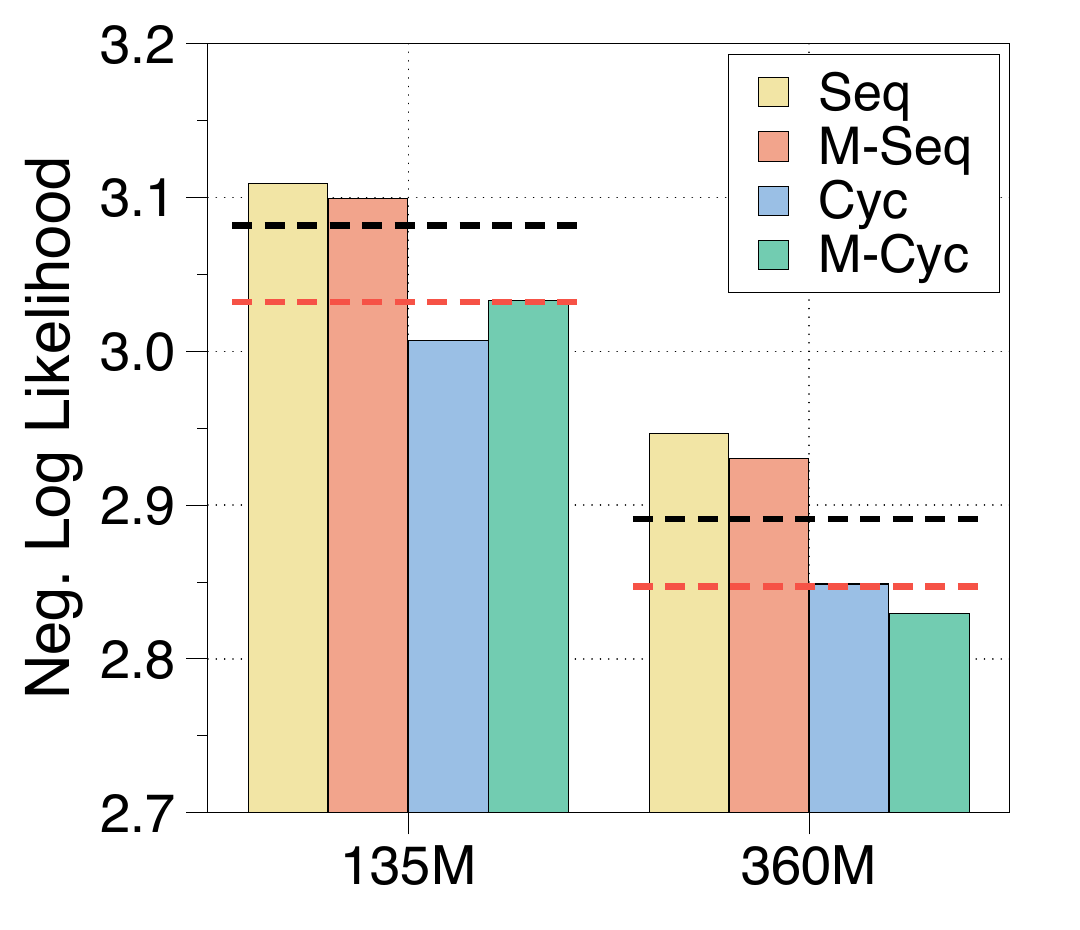}
        \subcaption{Recursion of two\,($N_R$ = 2)}
        \label{fig:revisit_sharing_rec2_app}
    \end{subfigure}
    \hspace{10pt}
    \centering
    \begin{subfigure}[t]{0.35\textwidth}
    \captionsetup{justification=centering}
        \includegraphics[width=\textwidth]{figs/revisit_sharing/revisit_sharing_rec3.pdf}
        \subcaption{Recursion of three\,($N_R$ = 3)}
        \label{fig:revisit_sharing_rec3_app}
    \end{subfigure}
    \caption{
    Validation negative log‑likelihood (lower is better) on \emph{FineWeb‑Edu} for four parameter‑sharing strategies.
    Bars are grouped by model capacity (135M vs.\ 360M parameters). 
    Middle‑Cycle consistently attains the lowest NLL, with its margin widening as either model size or depth increases.
    The horizontal dashed lines mark the \emph{untied} (non-sharing) baselines: the lower red line represents the full capacity model, while the upper black line represents a parameter-matched reduced model with a footprint equal to the unique trainable parameter sizes of the recursive model.
    }
    \label{fig:revisit_sharing_app}
\end{figure}

For fairer comparison, we also compare the Middle-Cycle and Cycle sharing strategies, fixing the number of unique parameters for both models. For 730M model comparison, both strategies use 3 recursion steps. Especially, Middle-Cycle uses 8 layers for the recursion block and 2 unshared layers (1 + 8$\times$3 + 1), resulting in 10 unique layers and 26 effective layers. We compare this to the Cycle strategy with 10 layers per recursion block (10$\times$3), which also has 10 unique layers but a total of 30 effective layers. Despite having fewer effective layers, as shown in \autoref{tab:m_cycle_cycle_compare}, the Middle-Cycle model achieves better performance, with a lower negative log-likelihood (2.7552 vs. 2.7573) and higher average few-shot accuracy (42.32 vs. 41.90). While these results are promising, we leave further validation across more scales as future work. 

\begin{table*}[h!]
    \caption{
    Comparison of Middle-Cycle and Cycle sharing strategies under the same number of unique parameters. All models use 730M scales (non-embedding parameters) as the base model with 3 recursion steps. The models are trained from scratch on 10B tokens using the Fineweb-Edu dataset.
    }
    \label{tab:m_cycle_cycle_compare}
    \small
    \centering
    \resizebox{\linewidth}{!}{
    \setlength{\tabcolsep}{4pt}
    \begin{tabular}{l|cc|ccc|c|cccccc|c}
    \toprule
      & \multicolumn{2}{c|}{\textbf{Recursion}} & \multicolumn{3}{c|}{\textbf{Pretrain}} & \textbf{NLL\,$\downarrow$} & \multicolumn{7}{c}{\textbf{Few-shot Accuracy\,$\uparrow$}} \\
    \cmidrule(l{2pt}r{2pt}){2-3} \cmidrule(l{2pt}r{2pt}){4-6} \cmidrule(l{2pt}r{2pt}){7-7} \cmidrule(l{2pt}r{2pt}){8-14}
    \textbf{Base Model} & Share & Loop & N-Emb\!\! & $N_L$ & $N_{tok}$ & Fineweb & LD & HS & PQ & WG & ARC & \!\!MMLU & Avg  \\ 
    \midrule
Vanilla 730M & Cycle & 3 & 252M & 10 & 10B & 2.7573 &  28.72 & 37.31 & 65.56 & 52.01 & 39.94 & 27.85 & 41.90 \\
Vanilla 730M & M-Cycle & 3 & 252M & 1+8+1 & 10B & 2.7552 & 29.75 & 37.73 & 66.00 & 52.49 & 40.49 & 27.50 & 42.32 \\
    \bottomrule
    \end{tabular}
    }
\end{table*}

\clearpage
\paragraph{Behavior under continued pre‑training (up‑training).}
\autoref{tab:revist_sharing_strategy_uptrain} extends the study by ``up‑training'' models---continuing from open-sourced SmolLM~\citep{allal2024SmolLM} checkpoints for an additional 5B tokens.
Both Middle strategies demonstrate superior performance across all settings, and notably, they significantly outperform the reduced baseline models that are initialized in the same manner but without recursion.
The other strategies reach a performance plateau earlier, suggesting that they have limited room for further improvement in capacity.

\begin{table*}[h!]
    \caption{
    Uptraining results across four parameter sharing strategies. Models are trained on 5B tokens from FineWeb-Edu and evaluated by train NLL and few-shot accuracy across six benchmarks. ARC denotes average of ARC-Easy and ARC-Challenge tasks, MMLU denotes the MMLU-Cont task. We highlight the optimal strategy for each setting in gray.\looseness=-1
    }
    \label{tab:revist_sharing_strategy_uptrain}
    \small
    \centering
    \resizebox{\textwidth}{!}{
    \setlength{\tabcolsep}{3pt}
    \begin{tabular}{l|ccc|ccc|c|cccccc|c}
    \toprule
      &  \multicolumn{3}{c|}{\textbf{Pretrain}} &  \multicolumn{3}{c|}{\textbf{Recursion}} & \textbf{NLL\,$\downarrow$} & \multicolumn{7}{c}{\textbf{Few-shot Accuracy\,$\uparrow$}} \\
    \cmidrule(l{2pt}r{2pt}){2-4} \cmidrule(l{2pt}r{2pt}){5-7} \cmidrule(l{2pt}r{2pt}){8-8} \cmidrule(l{2pt}r{2pt}){9-15} 
     \textbf{Base Model} & N-Emb & $N_L$ & $N_{tok}$ & Share & Init & Loop & FineWeb & LD & HS & PQ & WG & ARC & \!MMLU\! & Avg  \\
    \midrule
    Vanilla\,360M & 315M & 32 & 5B & -  & - & -  & 2.4825 & 41.67 & 50.63 & 70.35 & 55.09 & 46.99 & 30.82 & 49.26  \\
    Vanilla\,360M & 157M & 16 & 5B & - & Step  & 1   & 2.7168 & 31.85 & 37.59 & 64.74 & 53.20 & 41.06 & 27.34 & 42.63  \\
    Vanilla\,360M & 157M & 16 & 5B & Cyc  & Avg & 1  & 2.8603 & 22.14 & 30.36 & 60.07 & 48.22 & 34.99 & 25.56 & 36.89  \\
    Vanilla\,360M & 157M & 16 & 5B & Seq  & Avg & 1  & 2.7919 & 25.40 & 32.35 & 62.30 & 50.12 & 35.88 & 26.19 & 38.71  \\
    Vanilla\,360M & \,\,\,98M & 10 & 5B & -  & Step & 1  & 2.8915 & 26.63 & 35.03 & 64.42 & 52.09 & 38.75 & 26.86 & 40.63    \\
    Vanilla\,360M & \,\,\,98M & 10 & 5B & Cyc  & Avg & 1  & 3.0512 & 23.19 & 31.27 & 62.30 & 51.22 & 36.71 & 26.29 & 38.50   \\
    Vanilla\,360M & \,\,\,98M & 10 & 5B & Seq  & Avg & 1  & 2.9915 & 25.67 & 32.21 & 62.30 & 51.38 & 36.68 & 26.56 & 39.14  \\
    \midrule
     \rowcolor[gray]{0.9}
    Vanilla\,360M & 157M & 16 & 5B & Cyc  & Step & 2  & 2.7165 & 31.30 & 37.68 & 64.91 & 52.17 & 39.29 & 27.53 & 42.15  \\
    Vanilla\,360M & 157M & 16 & 5B & Cyc  & Avg & 2  & 2.8263 & 23.21 & 30.52 & 60.55 & 50.28 & 36.01 & 25.50 & 37.68  \\
    Vanilla\,360M & 157M & 16 & 5B & Cyc  & Lower & 2  & 2.8024 & 27.67 & 34.71 & 63.49 & 49.88 & 38.12 & 26.87 & 40.13  \\
    Vanilla\,360M & 157M & 16 & 5B & Cyc  & Upper & 2  & 2.7915 & 18.26 & 34.88 & 63.06 & 51.85 & 39.27 & 26.88 & 39.03  \\
    Vanilla\,360M & 157M & 16 & 5B & Cyc  & Rand & 2  & 2.7575 & 25.29 & 34.78 & 61.64 & 52.01 & 38.09 & 26.62 & 39.74   \\
    \midrule
     \rowcolor[gray]{0.9}
    Vanilla\,360M & 157M & 16 & 5B & Seq  & Step & 2  & 2.6862 & 34.14 & 42.49 & 67.90 & 53.35 & 43.24 & 28.80 & 44.99  \\
    Vanilla\,360M & 157M & 16 & 5B & Seq  & Avg & 2  & 2.7508 & 29.01 & 34.13 & 63.60 & 52.09 & 36.31 & 26.58 & 40.29  \\
    Vanilla\,360M & 157M & 16 & 5B & Seq  & Lower & 2  & 2.8300 & 27.50 & 33.28 & 63.38 & 51.38 & 37.44 & 26.32 & 39.88  \\
    Vanilla\,360M & 157M & 16 & 5B & Seq  & Upper & 2  & 2.7498 & 30.49 & 40.07 & 65.61 & 52.25 & 40.28 & 28.17 & 42.81  \\
    Vanilla\,360M & 157M & 16 & 5B & Seq  & Rand & 2  & 2.7153 & 32.00 & 41.31 & 66.10 & 53.35 & 42.13 & 28.52 & 43.90   \\
    \midrule
    Vanilla\,360M & 167M & 1+15+1 & 5B & M-Cyc  & Step & 2  & 2.6800 & 35.47 & 42.39 & 67.19 & 50.99 & 42.54 & 28.79 & 44.56  \\
    Vanilla\,360M & 167M & 1+15+1 & 5B & M-Cyc  & Avg & 2  & 2.7314 & 33.81 & 40.42 & 66.87 & 51.78 & 41.68 & 28.17 & 43.79  \\
    Vanilla\,360M & 167M & 1+15+1 & 5B & M-Cyc  & Lower & 2  & 2.7449 & 30.76 & 39.50 & 66.16 & 50.99 & 41.12 & 28.07 & 42.77  \\
     \rowcolor[gray]{0.9}
    Vanilla\,360M & 167M & 1+15+1 & 5B & M-Cyc  & Upper & 2  & 2.6605 & 34.41 & 43.74 & 67.46 & 53.20 & 43.75 & 28.93 & 45.25  \\
    Vanilla\,360M & 167M & 1+15+1 & 5B & M-Cyc  & Rand & 2  & 2.6730 & 35.65 & 43.04 & 67.74 & 52.17 & 42.60 & 28.62 & 44.97  \\
    \midrule
     \rowcolor[gray]{0.9}
    Vanilla\,360M & 167M & 1+15+1 & 5B & M-Seq  & Step & 2  & 2.6627 & 35.09 & 43.34 & 67.57 & 51.22 & 43.66 & 28.91 & 44.97  \\
    Vanilla\,360M & 167M & 1+15+1 & 5B & M-Seq  & Avg & 2  & 2.7143 & 33.92 & 40.93 & 66.49 & 51.70 & 40.72 & 28.24 & 43.66   \\
    Vanilla\,360M & 167M & 1+15+1 & 5B & M-Seq & Lower & 2  & 2.7696 & 30.74 & 38.36 & 65.94 & 51.78 & 41.27 & 27.73 & 42.64  \\
    Vanilla\,360M & 167M & 1+15+1 & 5B & M-Seq  & Upper & 2  & 2.6931 & 32.66 & 42.35 & 67.14 & 52.80 & 42.83 & 28.49 & 44.38  \\
    Vanilla\,360M & 167M & 1+15+1 & 5B & M-Seq  & Rand & 2  & 2.6908 & 35.07 & 42.03 & 66.32 & 53.75 & 43.11 & 28.42 & 44.78  \\
    \midrule
    Vanilla\,360M & \,\,\,98M & 10 & 5B & Cyc  & Step & 3  & 2.8901 & 27.46 & 35.26 & 63.82 & 51.54 & 39.35 & 27.44 & 40.81  \\
    Vanilla\,360M & \,\,\,98M & 10 & 5B & Seq & Step & 3  & 2.8258 & 30.43 & 37.37 & 63.76 & 52.33 & 40.55 & 27.65 & 42.02  \\
     \rowcolor[gray]{0.9}
    Vanilla\,360M & 118M & 1+10+1 & 5B & M-Cyc  & Step & 3  & 2.7735 & 31.38 & 39.31 & 65.51 & 50.51 & 40.70 & 27.65 & 42.51  \\
     \rowcolor[gray]{0.9}
    Vanilla\,360M & 118M & 1+10+1 & 5B & M-Seq  & Step & 3  & 2.7678 & 31.67 & 39.23 & 65.89 & 52.09 & 40.65 & 27.90 & 42.91  \\
    Vanilla\,360M & 118M & 1+10+1 & 5B & M-Cyc  & Upper & 3  & 2.8100 & 28.86 & 37.61 & 65.51 & 52.57 & 41.44 & 28.17 & 42.36  \\
    \bottomrule
    \end{tabular}
    }
\end{table*}

\clearpage
\section{Expanded Results of Design Choices for Router}
\label{app:router_ablation}

\subsection{Details of Design Configurations}\label{appx:router_details}

We investigate various router design choices to optimize performance and stability. Specifically, we tune the coefficient values controlling the strength of auxiliary or balancing loss terms (\textit{Coeff}), and adjust the scaling factor applied after the router function ($\alpha$) to modulate routing weights. Moreover, we test different activation functions (\textit{Func}), such as sigmoid or softmax, are evaluated, with architectural variations (\textit{Arch}) of the router network, including linear layer, 2-layer MLP with GELU activation, Wide-MLP that expands the hidden layer size by a factor of four.\looseness=-1

We also incorporate several techniques to stabilize training. To improve training stability, we utilize the \textit{router z-loss}~\citep{zoph2022st}, which penalizes large logits produced by the gating network. Large logits can cause numerical instability and hinder effective training of the router. The z-loss is computed as follows:
\begin{equation*}
L_z(x) = \frac{1}{B} \sum_{i=1}^B \left( \log \sum_{j=1}^{N_r} e^{x_j^{(i)}} \right)^2,
\end{equation*}
where $B$ is the number of tokens in the batch, $N_r$ is the number of experts, and $x \in \mathbb{R}^{B \times N_r}$ denotes the logits input to the router. This regularization encourages the gating network to produce smaller logits, promoting more stable and reliable routing decisions.

\subsection{Router Performance Evaluation Metrics}

\paragraph{Expert-choice routing.}
We evaluate dead token ratio and sampling accuracy to assess the router’s selection behavior. The \textit{dead token ratio} measures the proportion of tokens at specific positions within the batch that are consistently unselected during the final recursion step, indicating a positional bias where certain token positions are systematically neglected by the router. The \textit{sampling accuracy} how well the router used during inference predicts whether a token belongs to the top-$k$ tokens identified during training, reflecting the router’s ability to consistently select the most relevant tokens.
Ideally, high sampling accuracy with a low dead token ratio indicates a router that both identifies important tokens accurately and maintains diversity in token selection.

\vspace{-10pt}
\paragraph{Token-choice routing.}
We evaluate the router’s ability to balance token assignments across experts using MaxVio (maximum violation) and entropy metrics. \textit{MaxVio}~\citep{wang2024auxiliary} measures the load imbalance across experts:\looseness=-1
\begin{equation*}
\mathrm{MaxVio} = \frac{\max_i \mathrm{Load}_i - \overline{\mathrm{Load}}_i}{\overline{\mathrm{Load}}_i},
\label{eq:maxvio}
\end{equation*}
where $\mathrm{Load}_i$ denotes the actual number of tokens assigned to the $i$-th expert, and $\overline{\mathrm{Load}}_i$ represents the expected load per expert assuming perfect balance.

To measure the diversity of token assignments across experts, we also compute the \textit{Entropy} of the average selection probabilities for each expert:
\begin{equation*}
H = - \sum_{i=1}^{N_r} \overline{p}_i \log \overline{p}_i,
\label{eq:entropy}
\end{equation*}
where $\overline{p}_i$ is the average probability of selecting the $i$-th expert over all tokens in the evaluation batch, and $N_r$ is the total number of experts. A higher entropy indicates a more uniform distribution of tokens among experts, reflecting balanced and diverse routing decisions.\looseness=-1

\subsection{Extended Evaluation Results of Router Designs}

The results presented in \autoref{tab:ablation_study_expert_choice_app} indicate that although both the auxiliary router and auxiliary loss methods enhance sampling accuracy, they are also associated with high dead token ratios. In particular, some auxiliary router variants exhibit dead token ratios as high as 66.7\%, suggesting that the router always selects tokens from the same positions across inputs, reflecting a positional bias. Notably, employing a linear router architecture in conjunction with auxiliary loss effectively reduces the dead token ratio without compromising sampling accuracy.\looseness=-1

Results from \autoref{tab:ablation_study_token_choice_app} reveal that applying an explicit balancing loss significantly reduces MaxVio and increases entropy, leading to improved load balance without sacrificing overall model performance. Loss-free approaches, while simpler, tend to show higher MaxVio and lower entropy, indicating less balanced token routing. Architectures such as MLP and Linear routers perform comparably under balancing loss, with z-loss often contributing to improved routing stability and model accuracy. Nevertheless, it still struggles to achieve balance during quite long initial stage. The heterogeneity among the experts, stemming from the use of computation blocks with varying recursion depths as experts, likely complicates load balancing.\looseness=-1

\begin{table*}[h!]
    \caption{
    Ablation results on using expert-choice router with different routing configurations. We use the recursion-wise KV caching strategy by default. Coeff denotes coefficient values for auxiliary loss term, and $\alpha$ denotes scaling term after router function. Dead token ratio are measured within evaluation batch size of 500. Warmup refers to gradually decreasing the capacity from 1.0 to desired value over warmup steps. The last highlighted row represents a chosen final strategy, and intermediate best-performing designs (based on performance and routing metrics) are highlighted to illustrate how it was derived.\looseness=-1
    }
    \label{tab:ablation_study_expert_choice_app}
    \small
    \centering
    \resizebox{\linewidth}{!}{
    \setlength{\tabcolsep}{2pt}
    \begin{tabular}{lcccccc|cc|c|ccccccc}
    \toprule
      \multicolumn{7}{c|}{\textbf{Expert-choice Configurations}} &  \multicolumn{2}{c|}{\textbf{Router\,Metrics}} & \textbf{NLL}\,$\downarrow$ & \multicolumn{7}{c}{\textbf{Few-shot Accuracy\,$\uparrow$}} \\
    \cmidrule(l{2pt}r{2pt}){1-7} \cmidrule(l{2pt}r{2pt}){8-9} \cmidrule(l{2pt}r{2pt}){10-10} \cmidrule(l{2pt}r{2pt}){11-17}
    Sampling & Coeff & \!Func\! & $\alpha$ & Arch & Warmup & z-loss & Dead\,$\downarrow$ & Samp-Acc\,$\uparrow$ & FineWeb & LD & HS & PQ & WG & ARC & \!MMLU\! & Avg  \\ 
    \midrule
     - & - & \texttt{rand} & - & - & \xmark & \xmark & \,\,\,0.0 & - & 2.9335 & 26.0 & 33.1 & 61.6 & 52.3 & 35.8 & 26.2 & 39.1  \\
     \midrule
     Aux\,Router & - & - & - & MLP & \xmark & \xmark & 66.7 & 50.0 & \texttt{NaN} & 0.0 & 25.04 & 49.5 & 49.6 & 23.9 & 23.0 & 28.5  \\
     Aux\,Router & - & $\sigma$ & 0.1 & MLP & \xmark & \xmark & \,\,\,0.0 & 89.2 & 2.8893 & 26.1 & 33.8 & 62.0 & 51.5 & 36.6 & 26.4 & 39.4 \\
     \rowcolor[gray]{0.9}
     Aux\,Router & - & $\sigma$ & 1.0 & MLP & \xmark & \xmark & 66.7 & 50.0 & 2.8867 & 26.4 & 33.6 & 63.0 & 52.4 & 37.0 & 24.1 & 39.8 \\
     Aux\,Router & - & \texttt{tanh} & 0.1 & MLP & \xmark & \xmark & 66.7 & 98.6 & 2.8720 & 13.9 & 31.8 & 60.7 & 49.3 & 35.8 & 25.8 & 36.2  \\
     Aux\,Router & - & \texttt{tanh} & 1.0 & MLP & \xmark & \xmark & 66.7 & 97.0 & 3.0624 & 18.26 & 29.7 & 60.1 & 50.9 & 34.6 & 25.5 & 36.5 \\
     \midrule
     Aux\,Loss & 0.01 & - & - & MLP & \xmark & \xmark & 66.7 & 50.0 & \texttt{NaN} & 0.0 & 25.04 & 49.5 & 49.6 & 23.9 & 23.0 & 28.5 \\
     \rowcolor[gray]{0.9}
     Aux\,Loss & 0.01 & $\sigma$ & 0.1 & MLP & \xmark & \xmark & 0.0 & 99.6 & 2.8967 & 24.8 & 33.6 & 63.3 & 50.3 & 36.6 & 26.6 & 39.2\\
     Aux\,Loss & 0.01 & $\sigma$ & 1.0 & MLP & \xmark & \xmark & 65.9 & \!\!\!100.0 & 2.9189 & 12.0 & 31.6 & 59.4 & 51.5 & 33.2 & 25.3 & 35.5\\
     Aux\,Loss & 0.01 & \texttt{tanh} & 0.1 & MLP & \xmark & \xmark & 32.8 & 99.7 & 2.9426 & 23.5 & 32.4 & 62.4 & 49.8 & 35.6 & 26.0 & 38.3 \\
     Aux\,Loss & 0.01 & \texttt{tanh} & 1.0 & MLP & \xmark & \xmark & 0.0 & 98.8 & 3.2743 & 16.4 & 28.14 & 58.8 & 52.2 & 31.6 & 24.8 & 35.3 \\
     \midrule
     Aux\,Loss & 0.1 & $\sigma$ & 0.1 & MLP & \xmark & \xmark & 0.0 & 99.8 & 3.0416 & 21.5 & 31.0 & 61.8 & 50.3 & 35.0 & 26.0 & 37.6 \\
     \rowcolor[gray]{0.9}
     Aux\,Loss & \!\!0.001\!\! & $\sigma$ & 0.1 & MLP & \xmark & \xmark & 0.0 & 99.1 & 2.8816 & 27.6 & 34.3 & 63.0 & 51.6 & 36.7 & 26.5 & 40.0 \\
     Aux\,Loss & \!\!0.001\!\! & \texttt{tanh} & 0.1 & MLP & \xmark & \xmark & 0.0  & 56.4 & 2.9933 & 25.0 & 32.3 & 61.5 & 51.5 & 36.6 & 26.0 & 38.8 \\
     \midrule
     \rowcolor[gray]{0.9}
     Aux\,Loss & \!\!0.001\!\! & $\sigma$ & 0.1 & Linear & \xmark & \xmark & 0.1 & 99.2 & 2.8667 & 27.4 & 34.6 & 63.2 & 51.5 & 37.2 & 26.5 & 40.1  \\
     Aux\,Loss & \!\!0.001\!\! & $\sigma$ & 0.1 & W-MLP & \xmark & \xmark & 0.4 & 99.2 & 2.8716 & 27.8 & 33.9 & 62.4 & 49.9 & 36.3 & 26.3 & 39.4\\
     \midrule
     Aux\,Loss & \!\!0.001\!\! & $\sigma$ & 0.1 & Linear & \cmark & \xmark & 4.9 & 99.1 & 2.8744 & 26.0 & 33.9 & 62.0 & 51.2 & 36.1 & 26.1 & 39.2 \\
     Aux\,Loss & \!\!0.001\!\! & $\sigma$ & 0.1 & Linear & \xmark & \cmark & 0.0 & 99.3 & 2.8824 & 26.9 & 34.0 & 63.8 & 52.3 & 36.8 & 26.4 & 40.0\\
    \bottomrule
    \end{tabular}
    }
\end{table*}

\begin{table*}[h!]
    \caption{
    Ablation results on token-choice router under different routing configurations. We use the recursion-wise KV caching strategy by default. Coeff denotes coefficient for balancing loss term, or updating coefficient ($u$) for loss-free algorithm. The last highlighted row represents a chosen final strategy, but we added z-loss back in with a small coefficient of 1e-3 since it often stabilizes load balancing. The intermediate best-performing designs (based on performance and routing metrics) are highlighted to illustrate how it was derived.\looseness=-1
    }
    \label{tab:ablation_study_token_choice_app}
    \small
    \centering
    \resizebox{\linewidth}{!}{
    \setlength{\tabcolsep}{3.5pt}
    \begin{tabular}{lccccc|cc|c|ccccccc}
    \toprule
      \multicolumn{6}{c|}{\textbf{Token-choice Configurations}} &  \multicolumn{2}{c|}{\textbf{Router\,Metrics}} & \textbf{NLL}\,$\downarrow$ & \multicolumn{7}{c}{\textbf{Few-shot Accuracy\,$\uparrow$}} \\
    \cmidrule(l{2pt}r{2pt}){1-6} \cmidrule(l{2pt}r{2pt}){7-8} \cmidrule(l{2pt}r{2pt}){9-9} \cmidrule(l{2pt}r{2pt}){10-16}
    Balancing & Coeff & \!Func\! & $\alpha$ & Arch & Z-loss & MaxVio\,$\downarrow$ & Entropy\,$\uparrow$ & FineWeb & LD & HS & PQ & WG & ARC & \!MMLU\! & Avg  \\
    \midrule
     - & - & \texttt{rand} & - & - &  \cmark & 0.007 & 1.099 & 3.0268 & 24.8 & 32.0 & 61.4 & 52.2 & 35.5 & 26.1 & 38.7 \\
     \midrule
     Loss & 0.1 & \texttt{soft} & 1.0 & MLP &  \cmark & 0.200 & 1.076 & 3.0239 & 24.2 & 31.9 & 61.4 & 51.5 & 35.7 & 26.2 & 38.5 \\
     \rowcolor[gray]{0.9}
     Loss & 0.01 & \texttt{soft} & 1.0 & MLP &  \cmark & 0.682 & 0.921 & 2.9118 & 28.0 & 33.3 & 62.8 & 49.7 & 36.4 & 26.2 & 39.4 \\
     \midrule
     Loss-free & 0.01 & \texttt{soft} & 1.0 & MLP &  \cmark & 1.788 & 0.297 & 2.9078 & 25.5 & 32.5 & 61.3 & 52.3  & 36.1 & 26.0 & 38.9 \\
     Loss-free & 0.01 & $\sigma$ & 0.1 & MLP &  \cmark & 0.956 & 0.646 & 3.1144 &21.8 & 29.8 & 60.3 &51.6 & 34.0 & 25.7 & 37.2 \\
     Loss-free & 0.01 & $\sigma$ & 1.0 & MLP &  \cmark & 0.918 & 0.749 & 3.0188 & 23.4 & 31.3 & 59.9 & 50.0 & 35.2 & 25.8 & 37.6 \\
     \rowcolor[gray]{0.9}
     Loss-free & 0.001 & \texttt{soft} & 1.0 & MLP &  \cmark & 0.852 & 0.915 & 2.9081 & 25.8 & 33.6 & 62.8 &50.6 & 37.5 & 26.7 & 39.5  \\
     Loss-free & 0.001 & $\sigma$ & 0.1 & MLP &  \cmark & 1.281 & 0.551 & 2.9165 &23.9 & 33.1 & 61.2 & 51.6 & 37.3 & 26.2 &38.9 \\
     Loss-free & 0.001 & $\sigma$ & 1.0 & MLP &  \cmark & 0.542 & 0.941 & 3.0188 &24.9 & 32.0 & 61.9 & 51.4 & 35.5 & 25.9 & 38.6 \\
     \midrule
     \rowcolor[gray]{0.9}
     Loss & 0.1 & \texttt{soft} & 1.0 & Linear &  \cmark & 0.492 & 0.960 & 2.9974 &23.7 & 31.3 &  62.2 & 50.3 & 36.7 & 26.0 & 38.4 \\
     Loss & 0.1 & \texttt{soft} & 1.0 & W-MLP &  \cmark & 0.384  & 1.037 & 3.0293 & 25.3 & 31.5 & 62.2 &51.2 & 36.4 & 26.3 & 38.8 \\
     \midrule
     \rowcolor[gray]{0.9}
     Loss & 0.1 & \texttt{soft} & 1.0 & Linear &  \xmark & 0.266 & 1.056 & 2.9358 & 25.7 & 32.6 & 61.9 & 51.7 & 36.4 & 26.5 & 39.1 \\
    \bottomrule
    \end{tabular}
    }
\end{table*}

\section{Expanded Results of KV Cache Sharing Mechanism}
\label{app:expanded_kv}

\subsection{Key Value Representation Trends in Recursive Transformers}

Sharing KV caches across model depths has emerged as a promising approach to improve inference throughput in Vanilla Transformers~\citep{brandon2024reducing}. This technique can reduce the memory footprint required for KV caches, enabling larger inference batch sizes. Significant speedups can be also achieved by skipping the KV projection and even prefill operations at shared depths, especially with Cycle strategy~\citep{sun2024you}. Due to the high degree of freedom in Vanilla models—where trainable parameters can be well optimized for shared caches—these models exhibit only marginal performance drops when KV caches are shared between adjacent layers.
In contrast, Recursive Transformers have far fewer parameters available for being optimized to tied KV states. Nevertheless, we hypothesize that similar patterns may emerge between shared blocks. To investigate this, we decomposed the KV states from pretrained Recursive Transformers into magnitude and directional components.\looseness=-1

As shown in \autoref{fig:kv_sharing_mag_app}, the sharing of key and value projection layers across recursion depths leads to clear recursive patterns in the \textit{magnitude} values. Although the magnitudes of hidden states tend to increase, the projection layers appear to be trained to produce similar signal sizes at corresponding depths within each recursion.\looseness=-1

\begin{figure}[h]
    \centering
    \begin{subfigure}[t]{0.3334\textwidth}
    \captionsetup{justification=centering}
        \includegraphics[width=\textwidth]{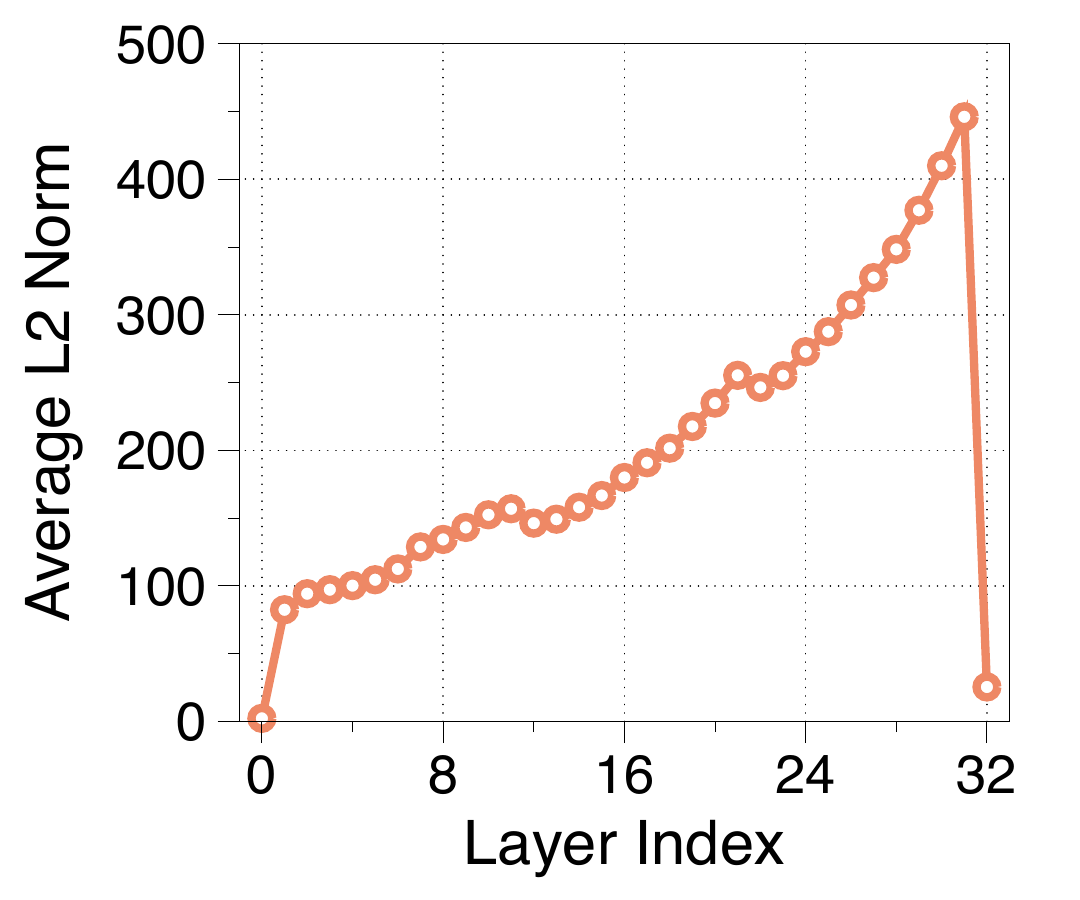}
        \subcaption{Hidden states}
    \end{subfigure}
    \centering
    \begin{subfigure}[t]{0.303\textwidth}
    \captionsetup{justification=centering}
        \includegraphics[width=\textwidth]{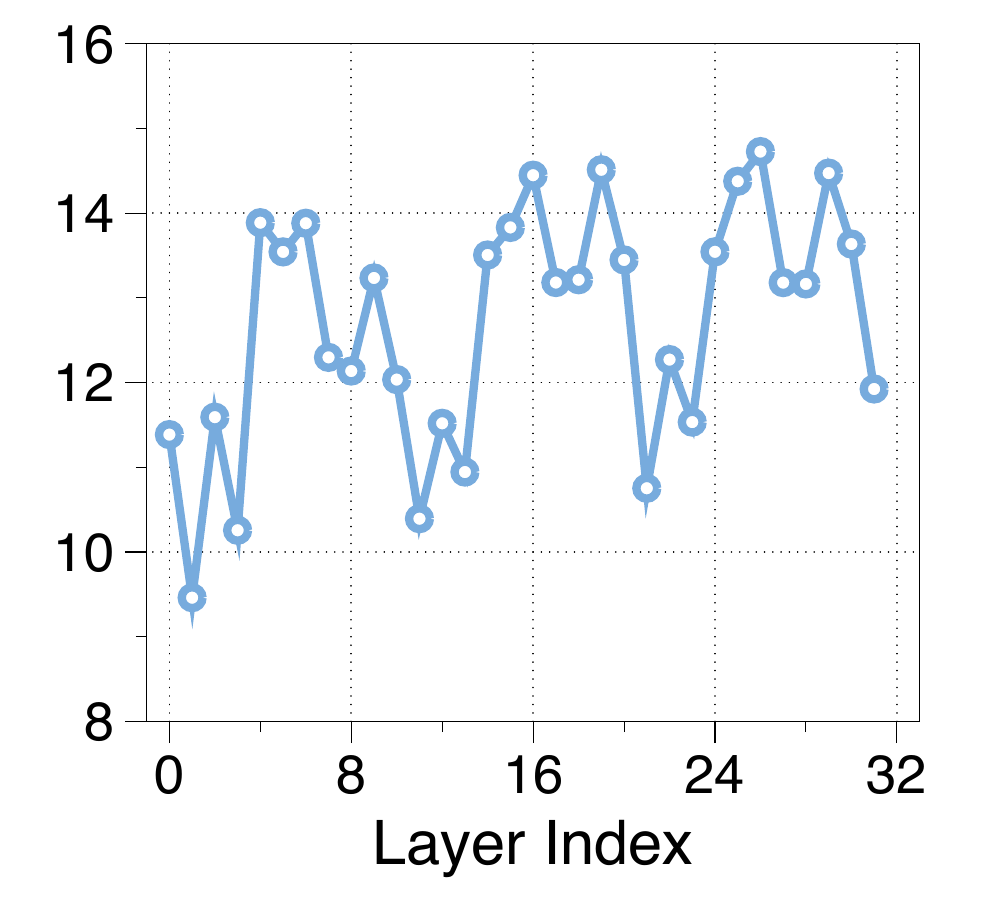}
        \subcaption{Key states}
    \end{subfigure}
    \centering
    \begin{subfigure}[t]{0.303\textwidth}
    \captionsetup{justification=centering}
        \includegraphics[width=\textwidth]{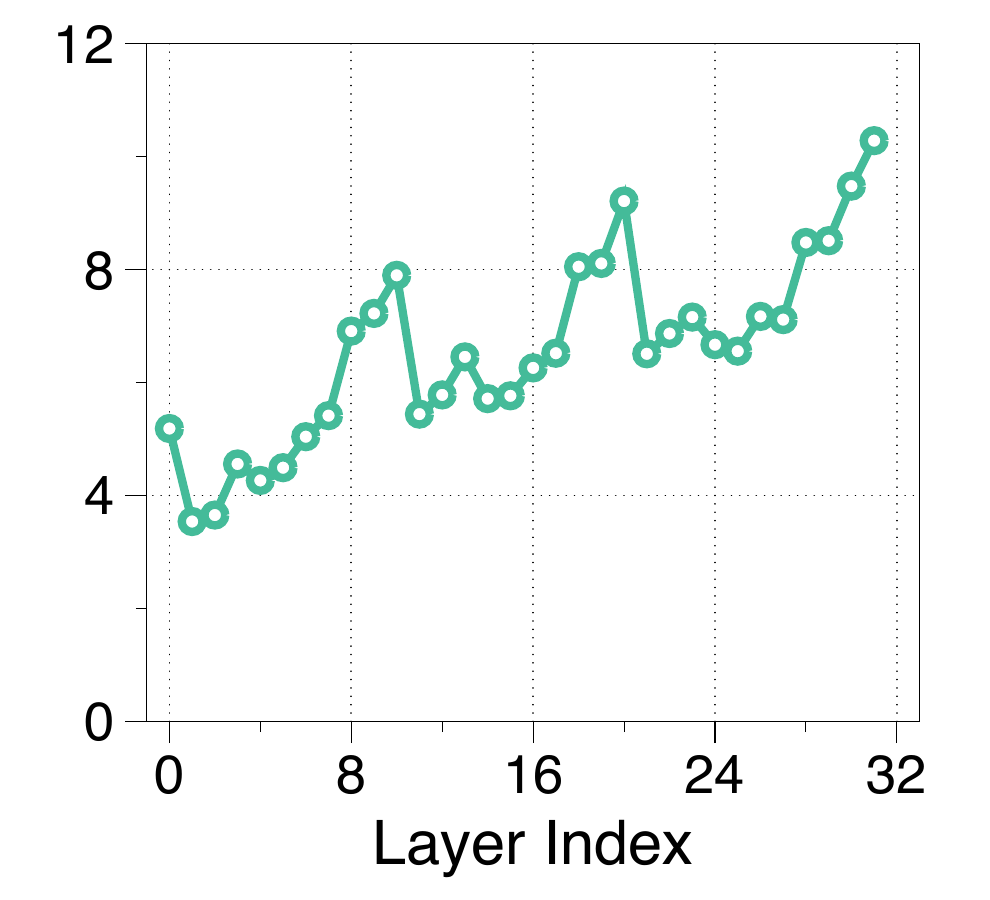}
        \subcaption{Value states}
    \end{subfigure}
    \caption{ Average L2 norm magnitude of (a) hidden states, (b) key states, and (c) value states across layers in a Middle-Cycled Recursive Transformer 360M with 3 recursion steps. 
    Note that the last hidden states correspond to the final hidden states after the last layer normalization. 
    }
    \label{fig:kv_sharing_mag_app}
\end{figure}

When we measured the cosine similarity in \autoref{fig:kv_sharing_dir_app}, distinct diagonal patterns emerge, suggesting that shared projection layers generate highly similar key and value representations. While sharing value states across recursions appears to be more challenging than sharing key states, these findings suggest that the performance drop from KV cache sharing can be marginal even in Recursive Transformers.\looseness=-1

\begin{figure}[h!]
    \centering
    \begin{subfigure}[t]{0.33\textwidth}
    \captionsetup{justification=centering}
        \includegraphics[width=\textwidth]{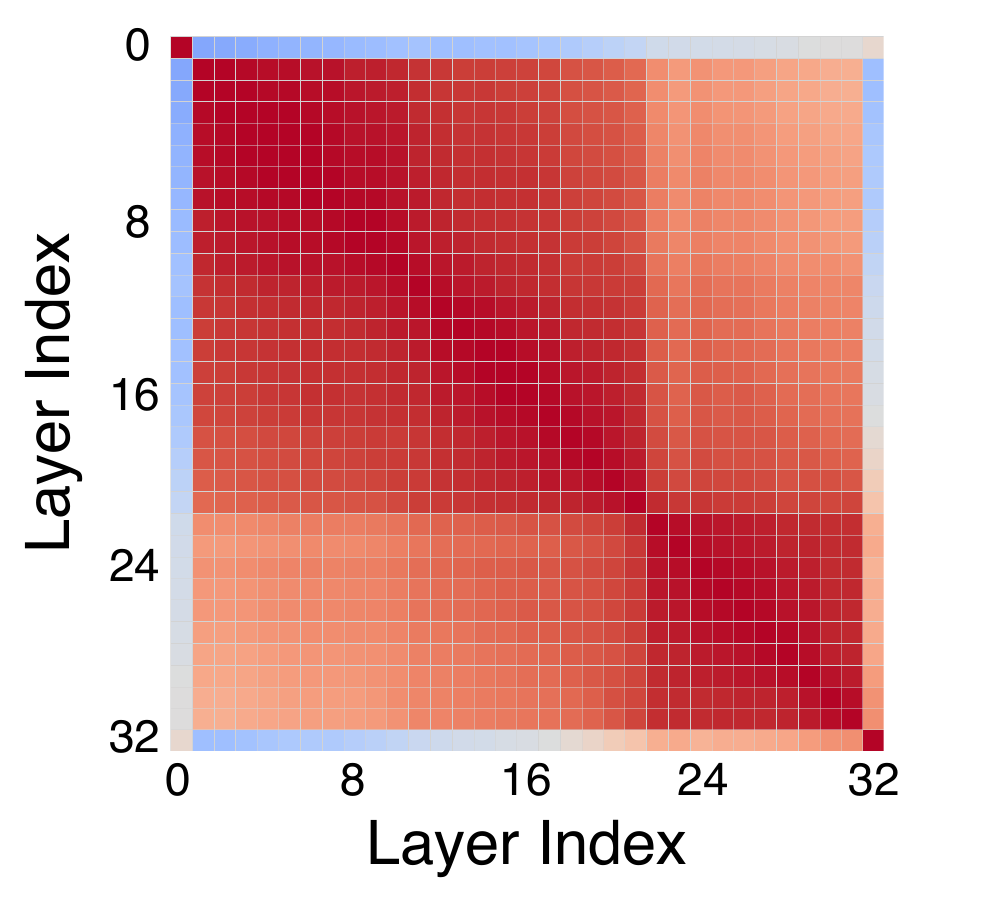}
        \subcaption{Hidden states}
    \end{subfigure}
    \centering
    \begin{subfigure}[t]{0.30\textwidth}
    \captionsetup{justification=centering}
        \includegraphics[width=\textwidth]{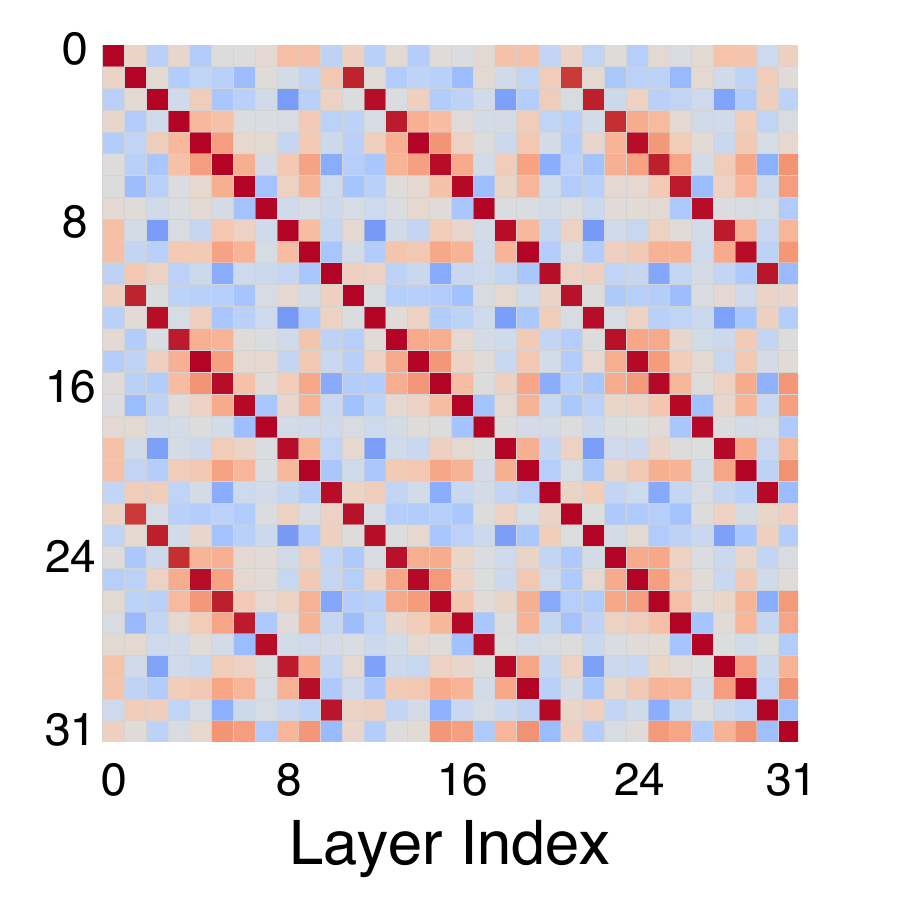}
        \subcaption{Key states}
    \end{subfigure}
    \centering
    \begin{subfigure}[t]{0.33\textwidth}
    \captionsetup{justification=centering}
        \includegraphics[width=\textwidth]{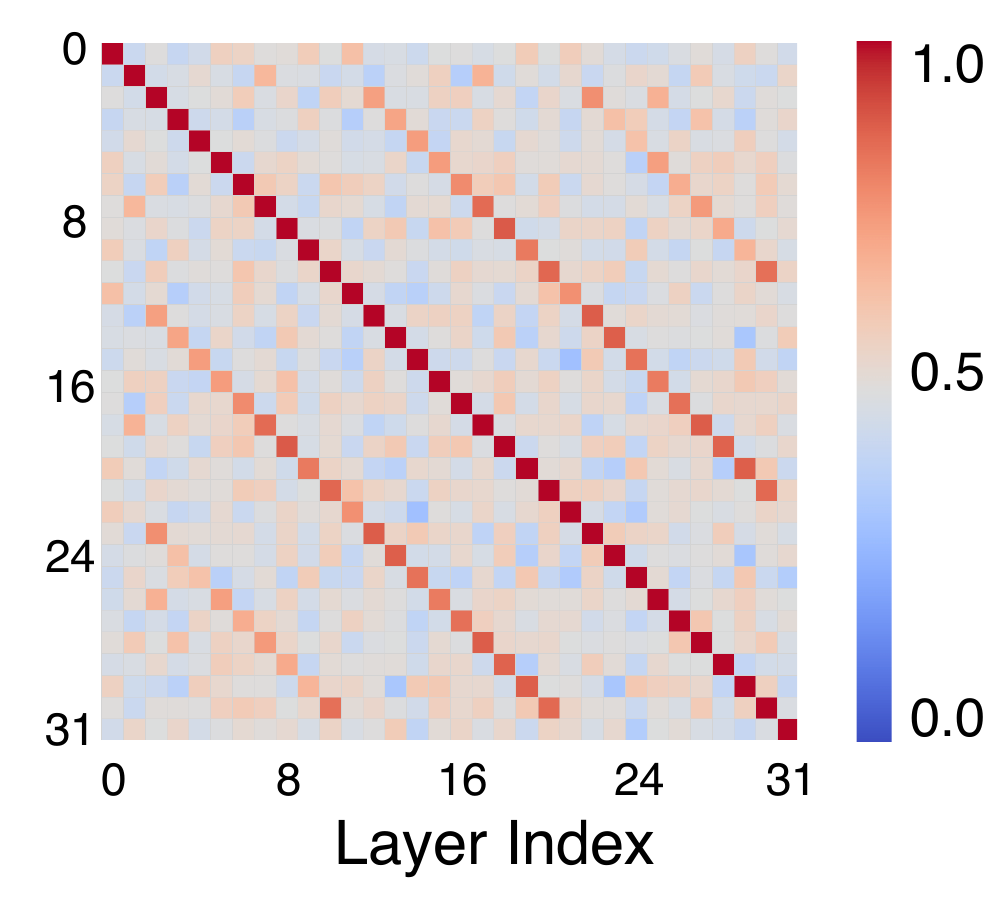}
        \subcaption{Value states}
    \end{subfigure}
    \caption{ Cosine similarity matrices showing the layer-wise similarity of (a) hidden states, (b) key states, and (c) value states in Recursive Transformer with Middle-Cycle strategy and recursion depth 3. Results are from a 360M parameter model with 32 layers. The hidden states matrix includes the final hidden states after the last layer normalization.
    }
    \label{fig:kv_sharing_dir_app}
\end{figure}

\subsection{Performance Comparison of KV Sharing Strategy}

\paragraph{Experimental results of key-value cache sharing.}
In \autoref{tab:key_value_sharing_results}, we present the performance results when applying KV cache sharing to Vanilla, Recursive, and MoR models. Especially, we tested various strategies for KV caches, including Cycle or Sequence strategies that share the same concepts as parameter sharing (see \S\ref{app:sharing_strategy} for details). Interestingly, KV cache sharing even improves the performance of vanilla models, where sharing acts as a regularization technique.
In case of recursive models, we align the sharing strategy for parameters and KV caches. Despite some variations in the results after applying KV sharing, the Middle-Cycle strategy (the best parameter sharing strategy) showed a slight perplexity drop, albeit not substantial.

When moving to MoR models, they still introduced a small amount of degradation in our best settings (expert-choice router). However, considering the reduced parameter sizes and cache sizes, we believe this minor drop is acceptable. Furthermore, we explored an alternative sharing strategy (indicated by $\dagger$) that utilized shared caches for inactive (unselected) positions while updating active positions through actual computation. This method is analogous to a recursive caching scheme but initializes inactive positions with key-value pairs from the first recursive iteration. Although it did not provide additional benefits, it is still worth exploring combinations of KV sharing and actual updates.\looseness=-1

\begin{table*}[h]
    \caption{
    Comparison of KV cache sharing strategies across Vanilla, Recursive, and MoR Transformers. Models are pretrained on 10B tokens of FineWeb-Edu, and evaluated using negative log-likelihood (NLL) on train set and few-shot accuracy across benchmarks. KV sharing denotes use of recursive KV sharing mechanism. If MoR is mentioned without further specification of KV sharing strategy, it implies the use of the recursion-wise caching strategy. $\text{}^\dagger$It indicates training with hybrid KV sharing that leverages shared caches for inactive positions while updating active ones through actual computation.\looseness=-1
    }
    \label{tab:key_value_sharing_results}
    \small
    \centering
    \resizebox{\textwidth}{!}{
    \setlength{\tabcolsep}{3.5pt}
    \begin{tabular}{l|cc|cc|c|cc|c|cccccc|c}
    \toprule
      &  \multicolumn{2}{c|}{\textbf{Pretrain}} &  \multicolumn{2}{c|}{\textbf{Recursion}} & \textbf{MoR} & \multicolumn{2}{c|}{\textbf{KV Sharing}} & \textbf{NLL\,$\downarrow$} & \multicolumn{7}{c}{\textbf{Few-shot Accuracy\,$\uparrow$}} \\
    \cmidrule(l{2pt}r{2pt}){2-3} \cmidrule(l{2pt}r{2pt}){4-5} \cmidrule(l{2pt}r{2pt}){6-6} \cmidrule(l{2pt}r{2pt}){7-8} \cmidrule(l{2pt}r{2pt}){9-9} \cmidrule(l{2pt}r{2pt}){10-16} 
     \textbf{Models} & N-Emb & $N_L$ & Share & Loop & Type & Share & Loop & FineWeb & LD & HS & PQ & WG & ARC & \!MMLU\! & Avg  \\
    \midrule
    Vanilla & 315M & 32 & - & - & - & - & - & 2.8471 & 27.3 & 34.8 & 64.2 & 52.8 & 38.3 & 26.7 & 40.7\\
    Vanilla & 315M & 32 & - & - & - & Seq & 2 & 2.7848 & 30.0 & 36.5 & 64.6 & 50.7 & 39.4 & 26.9 & 41.3 \\
    \rowcolor[gray]{0.9}
    Vanilla & 315M & 32 & - & - & - & Cyc & 2 & 2.7650 & 30.0 &	36.7 & 65.2 & 51.1  & 39.6 & 27.5 & 41.7\\
    \midrule
    Vanilla & 295M & 30 & - & - & - & - & - & 2.8069 & 29.1 & 35.6 & 65.1 & 50.4 & 38.5 & 27.3 & 41.0 \\
    \rowcolor[gray]{0.9}
    Vanilla & 295M & 30 & - & - & - & Seq & 3 & 2.7879 & 28.3 & 36.4 & 64.3 & 52.7 & 39.4 & 27.3 & 41.4 \\
    Vanilla & 295M & 30 & - & - & - & Cyc & 3 & 2.7890 & 28.9 & 36.5 & 64.6 & 51.4  & 39.0 & 27.6 & 41.3 \\
    \midrule
    Recursive & 157M & 16 & Seq & 2 & - & - & - & 2.9467 & 26.3 & 32.5 & 62.9 & 52.4 & 36.4 & 26.2 & 39.5\\
    Recursive &157M  & 16 & Seq & 2 & - & Seq & 2 & 2.8904 & 26.4 & 33.4 & 64.0 & 51.0 & 37.0 & 26.9 & 39.8\\
    Recursive & 157M & 16 & Cyc & 2 & - & - & - & 2.8487 & 28.5 & 34.8 & 63.1 & 50.0 & 37.4 & 28.8 & 40.1\\
    Recursive & 157M & 16 & Cyc & 2 & - & Cyc & 2 & 2.8577 & 26.2 & 34.5 & 64.2 & 51.4 & 37.3 & 26.9 & 40.1\\
    \rowcolor[gray]{0.9}
    Recursive & 167M & 1+15+1 & M-Cyc & 2 & - & - & - & 2.8295 & 28.6 & 35.0 & 64.5 & 50.5 & 39.7 & 27.2 & 40.9\\
    Recursive & 167M & 1+15+1 & M-Cyc & 2 & - & M-Cyc & 2 & 2.8451 & 27.3 &	34.7 & 63.7 & 50.5 & 37.8 & 27.0 & 40.2 \\
    \midrule
    Recursive & \,\,\,98M & 10 & Seq & 3 & - & - & - & 3.0245 & 24.6 & 31.5 & 63.1 & 49.3 & 35.7 & 25.7 & 38.3\\
    Recursive & \,\,\,98M & 10 & Seq & 3 & - & Seq & 3 & 2.9554 & 24.2 & 32.3 & 62.5 & 52.7 & 36.6 & 26.2 & 39.1\\
    Recursive & \,\,\,98M & 10 & Cyc & 3 & - & - & - & 2.9363 & 25.9 & 33.0 & 62.9 & 50.3 & 36.4 & 26.5 & 39.2\\
    Recursive & \,\,\,98M & 10 & Cyc & 3 & - & Cyc & 3 & 2.9155 & 24.1 & 32.9 & 62.4 & 51.2 & 37.4 & 26.7 & 39.1 \\
    \rowcolor[gray]{0.9}
    Recursive & 118M & 1+10+1 & M-Cyc & 3 & - & - & - & 2.8760 & 28.5 & 34.9 & 64.3 & 50.5 & 39.5 & 27.2 & 40.8\\
    Recursive & 118M & 1+10+1 & M-Cyc & 3 & - & M-Cyc & 3 & 2.8854 & 27.3 &	33.8 & 63.3 & 52.3 & 37.5 & 26.8 &	40.2\\
    \midrule
    \rowcolor[gray]{0.9}
    MoR & 118M & 1+10+1 & M-Cyc & 3 & Expert & - & - & 2.8667 & 27.4 & 34.6 & 63.2 & 51.5 & 37.2 & 26.5 & 40.1\\
    MoR & 118M & 1+10+1 & M-Cyc & 3 & Expert & M-Cyc & 3 & 2.8895 & 34.0 & 	61.6 & 50.2 & 26.0 & 36.5 & 27.0 & 39.2 \\
    MoR & 118M & 1+10+1 & M-Cyc & 3 & Expert & \,\,$\text{M-Cyc}^{\dagger}$ & 3 & 2.8653 & 24.8 & 34.3 & 62.0 & 50.1 & 36.7 & 26.7 & 39.1\\
    \midrule
    \rowcolor[gray]{0.9}
    MoR & 118M & 1+10+1 & M-Cyc & 3 & Token & - & - & 2.9358 & 25.7 & 32.6 & 61.9 & 51.7 & 36.4 & 26.5 & 39.1\\
    MoR & 118M & 1+10+1 & M-Cyc & 3 & Token & M-Cyc & 3 & 2.9155& 25.7 & 32.6 &	61.8 & 49.4 & 36.2 & 26.0 & 38.6\\
    \bottomrule
    \end{tabular}
    }
\end{table*}

\clearpage

\paragraph{Relaxation for key-value sharing constraints.}
We also investigated relaxing the constraints on KV sharing in \autoref{tab:key_value_sharing_relaxation}, similar to the relaxation approach in \citet{bae2024relaxed} for parameter sharing constraints. Specifically, we first re-examined four relaxation techniques for standard Recursive Transformers with very small ranks~\citep{hu2022lora, liu2024dora} or prefix lengths~\citep{liu2021p}. We also experimented with the position encoding~\citep{chen2025inner}, where trainable embeddings are element-wise multiplied with the output of each recursion block. 

Our results show that these techniques do not provide substantial performance improvements when pretraining relaxed models from scratch, consistent with prior studies, as they introduce only a limited number of additional parameters. Although we hypothesize that incorporating prefix-based approaches (such as adding trainable prefixes to attention) into KV sharing might lead to greater benefits, our experiments did not reveal substantial differences in this regard. Further exploration of more sophisticated techniques for efficiently relaxing KV cache sharing constraints remains an open direction for future research.

\begin{table*}[h]
    \caption{
    Experimental results of relaxing parameter sharing and KV cache sharing constraints in Recursive Transformers. All models are trained on FineWeb-Edu with 10B tokens, and we apply the Middle-Cycle parameter sharing for 360M models with 3 recursion depths. We evaluate them based on training NLL and few-shot accuracy across six benchmarks. Relaxation types include encoding trainable embeddings on recursion outputs via element-wise multiplication (Enc), applying LoRA and DoRA to query and value weight matrices, and adaptation prompt tuning (Adapt-P). 
    }
    \label{tab:key_value_sharing_relaxation}
    \small
    \centering
    \resizebox{\textwidth}{!}{
    \setlength{\tabcolsep}{3.5pt}
    \begin{tabular}{l|cc|ccc|cc|c|cccccc|c}
    \toprule
      &  \multicolumn{2}{c|}{\textbf{Pretrain}} &  \multicolumn{3}{c|}{\textbf{Relaxation}} &  \multicolumn{2}{c|}{\textbf{KV Sharing}} & \textbf{NLL\,$\downarrow$} & \multicolumn{7}{c}{\textbf{Few-shot Accuracy\,$\uparrow$}} \\
    \cmidrule(l{2pt}r{2pt}){2-3} \cmidrule(l{2pt}r{2pt}){4-6} \cmidrule(l{2pt}r{2pt}){7-8} \cmidrule(l{2pt}r{2pt}){9-9} \cmidrule(l{2pt}r{2pt}){10-16} 
     \textbf{Models} & N-Emb & $N_L$ & Type & Rank & Len & Share & Loop & FineWeb & LD & HS & PQ & WG & ARC & \!MMLU\! & Avg  \\
    \midrule
    Recursive & 118M & 1+10+1 & - & - & - & - & - & 2.8854 & 27.3 &	33.8 & 63.3 & 52.3 & 37.5 & 26.8 &	40.2\\
    Recursive & 118M & 1+10+1 & Enc & - & - & - & - & 2.8604 & 27.3 & 34.6 & 63.9 & 53.4 & 38.6 & 26.7 & 40.2\\
    Recursive & 124M & 1+10+1 & LoRA & 64 & - & - & - & 2.8599 & 27.3 & 34.6 & 64.3 & 50.9 & 38.0	& 26.9 & 39.7 \\
    Recursive & 124M & 1+10+1 & DoRA & 64 & - & - & - & 2.8945 & 26.4 &	33.6 & 64.4 & 50.6 & 37.4 &	26.5 & 39.2 \\
    Recursive & 126M & 1+10+1 & Adapt-P & - & 256 & - & - & 2.8626 & 27.1 & 34.7 & 64.0 & 51.9	& 37.6 & 26.8 & 39.7\\
    \midrule
    Recursive & 118M & 1+10+1 & - & - & - & M-Cyc & 3 & 2.8854 & 27.3 &	33.8 & 63.3 & 52.3 & 37.5 & 26.8 & 40.2\\
    Recursive & 126M & 1+10+1 & Adapt-P & - & 256 & M-Cyc & 3 & 2.9030 & 24.5 & 33.1 & 63.0 & 52.2 & 26.7 & 37.6 &	39.5\\
    \bottomrule
    \end{tabular}
    }
\end{table*}

\section{Expanded Qualitative Results}
\label{app:qualitative_results}

\subsection{Analysis on Adaptive Computation Paths}

\autoref{tab:qualitative_highlighted_results} illustrates a qualitative analysis of the recursion depth assigned to each subword token. This visualization provides a detailed insight into how tokens within each sample exhibit varying levels of recursive processing, showcasing the adaptive computation mechanism within the MoR framework. Notably, some tokens exit early (purple), while others require deeper processing (blue and red), reflecting the model’s ability to focus more compute on challenging parts of the input.

\begin{table*}[ht]
\centering
\footnotesize
\caption{Visualization of the recursion depth for each subword token, with colors representing the number of recursion steps: \raisebox{0pt}[0.5em][0.1em]{\colorbox{DarkOrchid!25}{1}}, \raisebox{0pt}[0.5em][0.1em]{\colorbox{SkyBlue!55}{2}}, and \raisebox{0pt}[0.5em][0.1em]{\colorbox{Salmon!60}{3}}. This depth specifically indicates how many times recursion is applied at that token's position to predict the subsequent token. Each row corresponds to a single sample, offering a clear illustration of the token-level recursion distribution in practice. We use an MoR model with $N_r = 3$, auxiliary loss, and recursion-wise KV caching. This model is built on a 360M parameter base and trained on 30B tokens.
}
\begin{tabular}{@{}l p{13cm}@{}}
\toprule
\textbf{Sample} & \textbf{Text} \\
\bottomrule
\end{tabular}
\label{tab:qualitative_highlighted_results}
\end{table*}

\clearpage

\subsection{Analysis on Router Weights}

To gain insight into the router output distributions, we visualized the results in \autoref{fig:weight_values_app}. Our analysis reveals that various routing mechanisms are optimized to balance expert loads according to the desired capacity. Notably, expert-choice routers achieved nearly perfect load balancing with the auxiliary loss, resulting in almost binary values (1 or 0) for selected and unselected tokens, respectively. 
For the auxiliary router, it was able to distinguish between selected and unselected tokens to some extent, but still showed overlapping. Since this strategy allows for different capacity factors during training and inference\footnote{The auxiliary router learns to capture intra-token differences within selected or unselected groups, rather than being biased towards extreme points.}, further research into methodologies that can more distinctly separate inter-cluster variations seems necessary.

Other token-choice strategies also exhibited good balancing properties with reasonable router values, which are used to refine the outputs of the corresponding recursion computation blocks. However, most cases failed to converge to optimal load balancing (i.e., these were edge cases where they achieved their own optimal load balancing), highlighting the challenges of achieving consistent performance in heterogeneous expert settings.

\begin{figure}[h]
    \centering
    \begin{subfigure}[t]{0.4\textwidth}
    \captionsetup{justification=centering}
        \includegraphics[width=\textwidth]{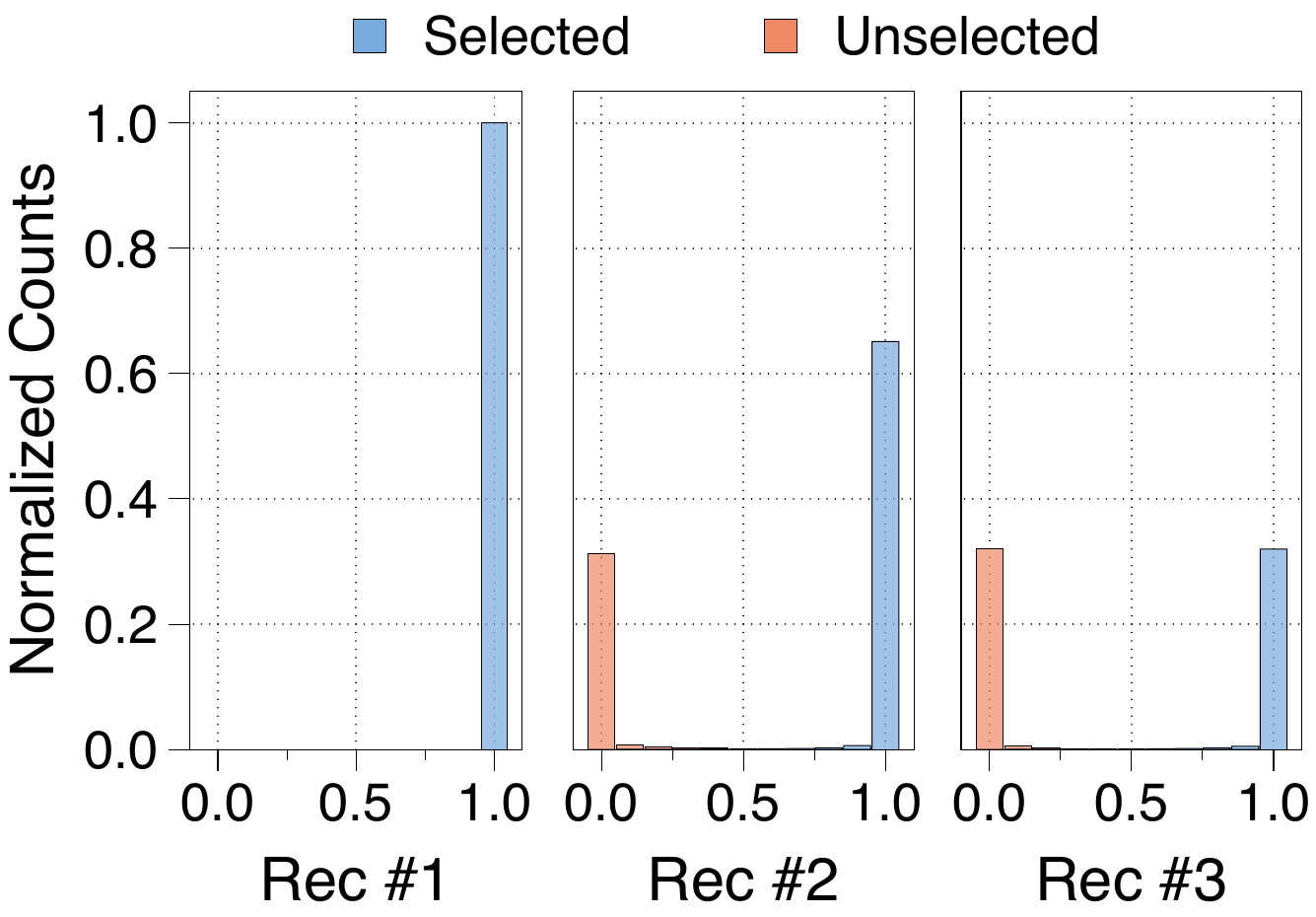}
        \subcaption{Expert-choice (Auxiliary loss)}
    \end{subfigure}
    \centering
    \hspace{25pt}
    \begin{subfigure}[t]{0.4\textwidth}
    \captionsetup{justification=centering}
        \includegraphics[width=\textwidth]{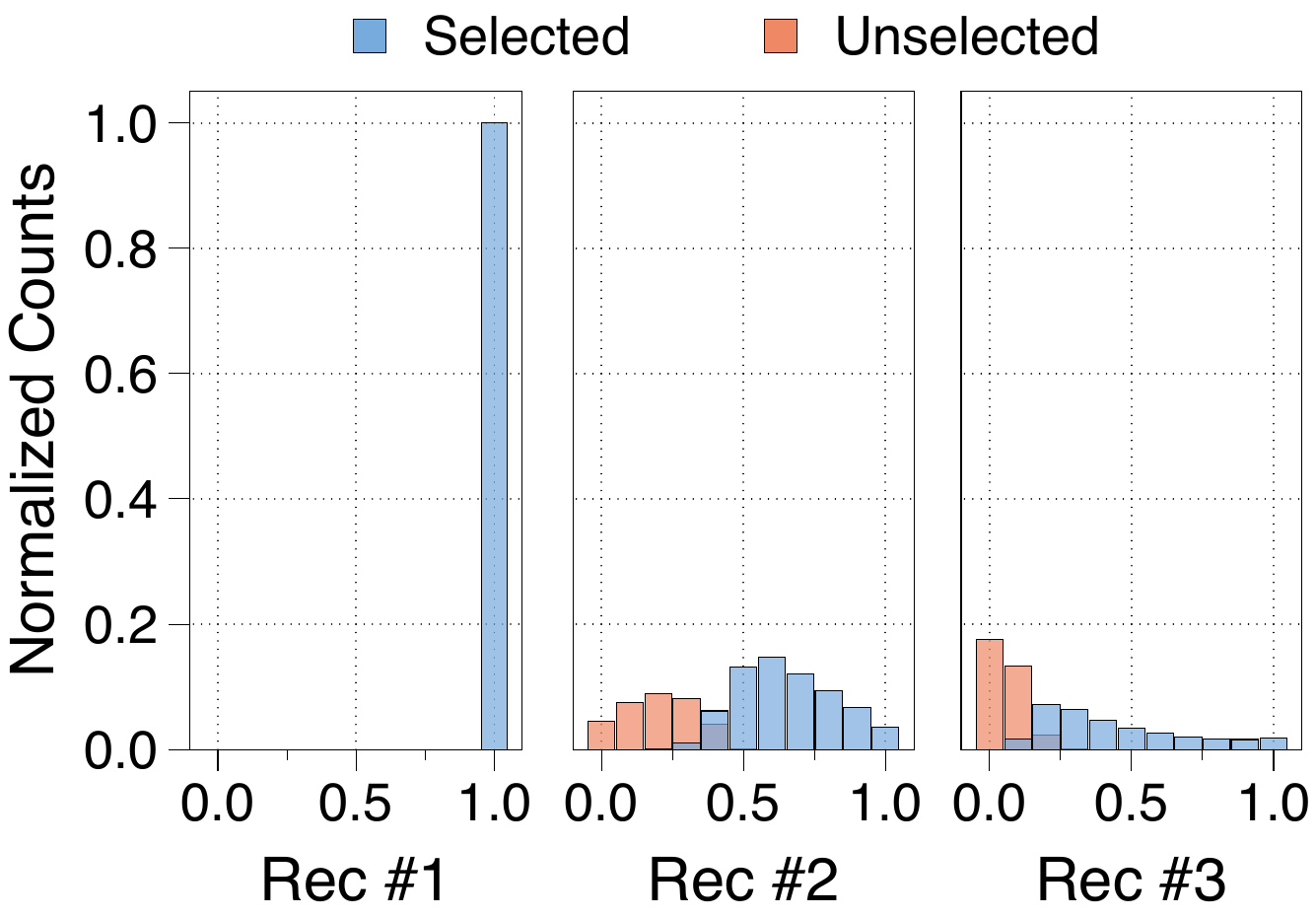}
        \subcaption{Expert-choice (Auxiliary router)}
    \end{subfigure}
    \centering
    \begin{subfigure}[t]{0.4\textwidth}
    \captionsetup{justification=centering}
        \includegraphics[width=\textwidth]{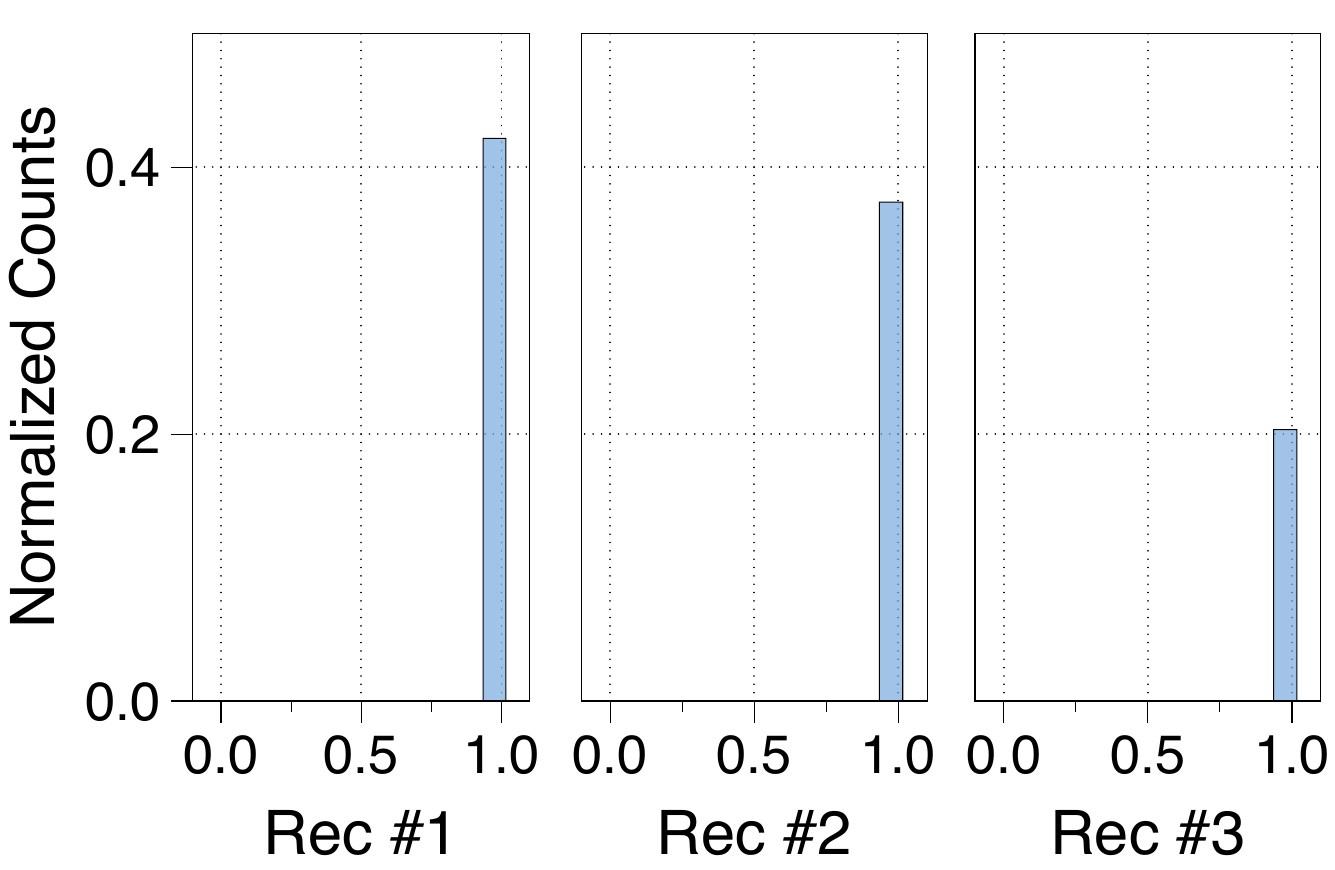}
        \subcaption{Token-choice (Balancing loss)}
    \end{subfigure}
    \centering
    \hspace{25pt}
    \begin{subfigure}[t]{0.4\textwidth}
    \captionsetup{justification=centering}
        \includegraphics[width=\textwidth]{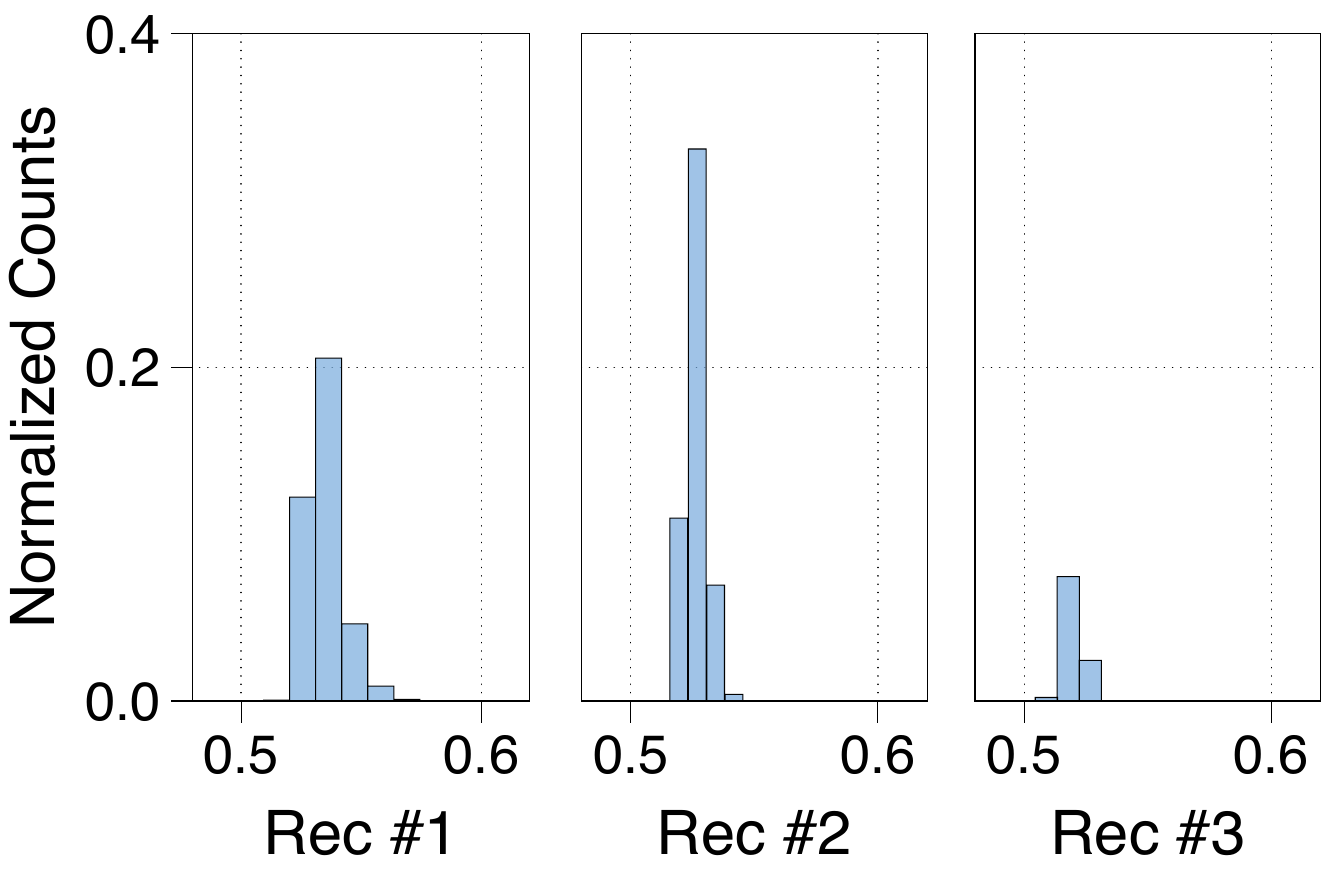}
        \subcaption{Token-choice (Loss-free)}
    \end{subfigure}
    \caption{
    Distribution of router weights for selected and unselected tokens at each recursion step in expert-choice and token-choice MoR~($N_r$=3). All models use the recursion-wise KV caching strategy. The subplots show results for (a) expert-choice routing with auxiliary loss, (b) auxiliary router, (c) token-choice routing with balancing loss, and (d) loss-free algorithm. Each subplot uses the best hyperparameter settings identified in \autoref{tab:ablation_study_router}.
    }    
    \label{fig:weight_values_app}
\end{figure}

}

\end{document}